\documentclass[acmtog,timestamp,screen]{acmart}

\usepackage[utf8]{inputenc}
\usepackage[T1]{fontenc}
\usepackage{tabularx}
\usepackage[capitalise]{cleveref}
\usepackage{float}
\Crefname{figure}{Figure}{Figures}

\setcitestyle{square}

\definecolor{skyblue}{rgb}{0.2,0.6,0.9}
\definecolor{placeholder}{rgb}{0.6,0.8,0.95}

\DeclareMathOperator*{\argmin}{argmin}

\acmJournal{TOG}
\acmYear{2019}

\setcopyright{acmcopyright}

\received{September 2018}
\received[accepted]{January 2019}
\received[final version]{January 2019}

\begin{document}

\title[LiveCap: Real-time Human Performance Capture from Monocular Video]{LiveCap:\\ Real-time Human Performance Capture from Monocular Video}

\author{Marc Habermann} 
\orcid{0000-0003-3899-7515}
\affiliation{
	\institution{Max Planck Institute for Informatics}
	\streetaddress{Campus E1 4, Stuhlsatzenhausweg}
	\city{Saarbrücken} 
	\state{Germany} 
	\postcode{66123}
}

\author{Weipeng Xu}
\orcid{0000-0001-9548-5108}
\affiliation{
  \institution{Max Planck Institute for Informatics}
  \streetaddress{Campus E1 4, Stuhlsatzenhausweg}
  \city{Saarbrücken} 
  \state{Germany} 
  \postcode{66123}
}

\author{Michael Zollh{\"o}fer}
\orcid{0000−0003−1219−0625}
\affiliation{
	\institution{Stanford University}
	\streetaddress{353 Serra Mall}
	\city{Stanford} 
	\state{United States of America} 
	\postcode{94305}
}

\author{Gerard Pons-Moll}
\orcid{0000−0001−5115−7794}
\affiliation{
	\institution{Max Planck Institute for Informatics}
	\streetaddress{Campus E1 4, Stuhlsatzenhausweg}
	\city{Saarbrücken} 
	\state{Germany} 
	\postcode{66123}
}

\author{Christian Theobalt}
\orcid{0000−0001−6104−6625}
\affiliation{
	\institution{Max Planck Institute for Informatics}
	\streetaddress{Campus E1 4, Stuhlsatzenhausweg}
	\city{Saarbrücken} 
	\state{Germany} 
	\postcode{66123}
}

\email{mhaberma@mpi-inf.mpg.de}
\renewcommand\shortauthors{Marc Habermann, Weipeng Xu, Michael Zollh{\"o}fer, Gerard Pons-Moll, and Christian Theobalt}

\begin{teaserfigure}
	\includegraphics[width=\textwidth]{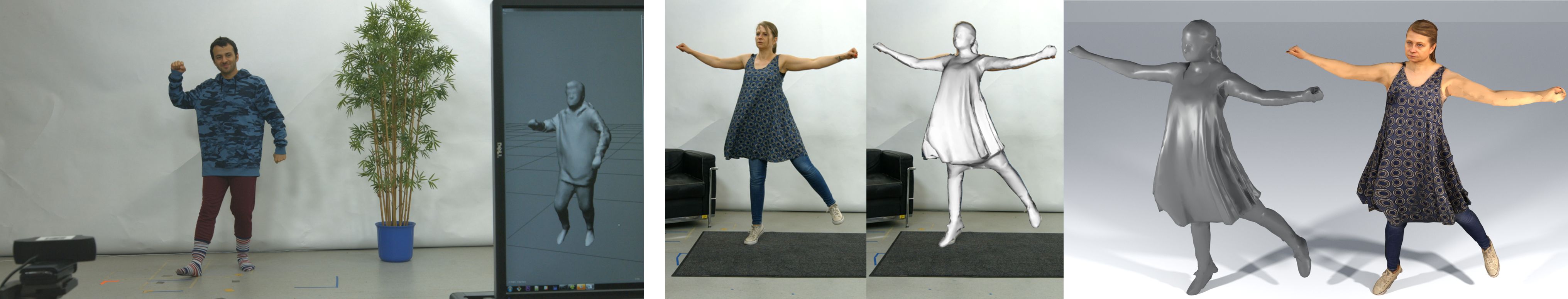}
	\caption
	{
		We propose the first real-time human performance capture approach that reconstructs dense, space-time coherent deforming geometry of people in their loose everyday clothing from just a single monocular RGB stream, e.g., captured by a webcam.
	}
	\label{fig:teaser}
\end{teaserfigure}
 
\begin{abstract}
We present the first real-time human performance capture approach that reconstructs dense, space-time coherent deforming geometry of entire humans in general everyday clothing from just a single RGB video.
We propose a novel two-stage analysis-by-synthesis optimization whose formulation and implementation are designed for high performance.
In the first stage, a skinned template model is jointly fitted to background subtracted input video, 2D and 3D skeleton joint positions found using a deep neural network, and a set of sparse facial landmark detections.
In the second stage, dense non-rigid 3D deformations of skin and even loose apparel are captured based on a novel real-time capable algorithm for non-rigid tracking using dense photometric and silhouette constraints.
Our novel energy formulation leverages automatically identified material regions on the template to model the differing non-rigid deformation behavior of skin and apparel.
The two resulting non-linear optimization problems per-frame are solved with specially-tailored data-parallel Gauss-Newton solvers.
In order to achieve real-time performance of over 25Hz, we design a pipelined parallel architecture using the CPU and two commodity GPUs.
Our method is the first real-time monocular approach for full-body performance capture.
Our method yields comparable accuracy with off-line performance capture techniques, while being orders of magnitude faster.
\end{abstract}



\thanks{This work was funded by the ERC Consolidator Grant 4DRepLy (770784).}

\maketitle

\section{Introduction}
Dynamic models of virtual human actors are key elements of modern visual effects for movies and games, and they are invaluable for believable, immersive virtual and augmented reality, telepresence, as well as 3D and free-viewpoint video.
Such virtual human characters ideally feature high-quality, space-time coherent dense models of shape, motion and deformation, as well as appearance of people, irrespective of physique or clothing style.
Creating such models at high fidelity often requires many months of work of talented artists.
To simplify the process, marker-less performance capture methods were researched to reconstruct at least parts of such models from camera recordings of real humans in motion.
\par 
Existing multi-camera methods capture human models at very good quality, but often need dense arrays of video or depth cameras and controlled studios, struggle with complex deformations, and need pre-captured templates.
Only few multi-view methods achieve real-time performance, but no real-time method for single RGB 
performance capture exists.
Many applications in interactive VR and AR, gaming, virtual try-on \cite{Hilsmann:2009:TRC,Sekine,Pons-Moll:Siggraph2017}, pre-visualization for visual effects, 3DTV or telepresence~\cite{orts2016holoportation}
critically depend on real-time performance capture.
The use of complex camera arrays and studios restricted to indoor scenes presents a practical barrier to these applications.
In daily use, systems should ideally require only one camera and work outdoors.
\par 
Under these requirements, performance capture becomes a much harder and much more underconstrained problem.
Some methods have approached this challenge by using multiple~\cite{dou2016fusion4d,collet2015high,wang2016capturing} or a single low-cost consumer-grade depth (RGB-D)~\cite{Newcombe_2015_CVPR,BodyFusion} camera for dense non-rigid deformation tracking.
While these methods are a significant step forward, RGB-D cameras are not as cheap and ubiquitous as color cameras, 
often have a limited capture range, do not work well under bright sunlight, and have limited resolution.
Real-time human performance capture with a single color camera would therefore greatly enhance and simplify performance capture and further democratize its use, in particular in the aforementioned interactive applications of ever increasing importance.
However, dense real-time reconstruction from one color view is even harder, and so today's best monocular methods only capture very coarse models, such as bone skeletons \cite{VNect_SIGGRAPH2017,sun2017compositional}.
\par 
In this paper, we propose the - to our knowledge - first real-time human performance capture method that reconstructs dense, space-time coherent deforming geometry of people in their loose everyday clothing from a single video camera.
In a pre-processing step, the method builds a rigged surface and appearance template from a short video of the person in a static pose, on which regions of skin and pieces of apparel are automatically identified using a new multi-view segmentation that leverages deep learning.
The template is fitted to the video sequence in a new coarse-to-fine two-stage optimization, whose problem formulation and implementation are rigorously designed for best accuracy at real-time performance.
In its first stage, our new real-time skeleton pose optimizer fits the skinned template to (1) 2D and 3D skeleton joint positions found with a CNN, to (2) sparse detected facial landmarks, and (3) to the foreground silhouette.
\par 
In a second stage, dense non-rigid 3D deformations of even loose apparel is captured.
To this end, we propose a novel real-time capable algorithm for non-rigid analysis-by-synthesis tracking from monocular RGB data.
It minimizes a template-to-image alignment energy jointly considering distance-field based silhouette alignment, dense photometric alignment and spatial and temporal regularizers, all designed for real-time performance.
The energy formulation leverages the shape template segmentation labels (obtained in the pre-processing stage) to account for the varying non-rigid deformation behavior of different clothing during reconstruction.
The non-linear optimization problems in both stages are solved with specially-tailored GPU accelerated Gauss-Newton solvers.
In order to achieve real-time performance of over 25~Hz, we design a pipelined solver architecture that executes the first and the second stage on two GPUs in a rolling manner.
Our approach captures high-quality models of humans and their clothing in real-time from a single monocular camera.
We demonstrate intriguing examples of live applications in 3D video and virtual try-on.
We show qualitatively and quantitatively that our method outperforms related monocular on-line methods and comes close to off-line performance capture approaches in terms of reconstruction density and accuracy.
\par 
In summary, our contributions are:
(1) We propose the first real-time system for monocular human performance capture.
In order to achieve real-time performance, we not only made specific algorithmic design choices, but also contribute several new algorithmic ideas, e.g., the adaptive material based regularization and the displacement warping to guarantee high quality results under a tight real-time constraint.
(2) We also show how to efficiently implement these design decisions by combining the compute power of two GPUs and the CPU in a pipelined architecture and how dense and sparse linear systems of equations can be efficiently optimized on the GPU.
(3) To evaluate our approach on a wide range of data, we show high quality results on an extensive new dataset of more than 20 minutes of video footage captured in 11 scenarios, which contain different types of loose apparel and challenging motions.
\section{Related Work}
Performance capture methods typically use multi-view images or depth sensors.
We focus here on approaches to capture 3D humans in motion, and leave out 
the body of work on 2D pose and shape capture.
Most monocular-based methods ignore clothing and are restricted to capturing the articulated motion and the undressed shape of the person.
Since there are almost no works that do performance capture from monocular video we focus here on multi-view and depth-based methods and approaches that capture pose and undressed shape from single images.
\paragraph{Multi-view}
Many multi-view methods use stereo and shape from silhouette cues to capture the moving actor~\cite{matusik2000image,starck2007surface,waschbusch2005scalable,collet2015high}, or reconstruct via multi-view photometric stereo~\cite{vlasic2009dynamic}.
Provided with sufficient images some methods directly non-rigidly deform a subject specific template mesh~\cite{Carranza:2003,cagniart2010free,de2008performance} or a volumetric shape representation \cite{huang2016volumetric,InriaVolumetric_2015}.
Such methods are free-form and can potentially capture arbitrary shapes~\cite{Mustafa:16} as they do not incorporate any skeletal constraints.
Such flexibility comes at the cost of robustness.
To mitigate this, some methods incorporate a skeleton in the template to constrain the motion to be nearly articulated~\cite{gall2009motion,vlasic2008articulated,liu2011markerless}.
This also enables off-line performance capture from a stereo pair of cameras~\cite{wu2013onset}.
Some systems combine reconstruction and segmentation to improve results~\cite{bray2006posecut,brox2010combined,liu2011markerless,wu2012full}.
Such methods typically require a high resolution scan of the person as input.
To side step scanning, a parametric body model can be employed.
Early models were based on simple geometric primitives~\cite{plankers2001tracking,sminchisescu2003kinematic,sigal2004tracking,metaxas1993shape}.
Recent ones are more accurate, detailed and are learned from thousands of scans~\cite{anguelov2005scape,hasler2010multilinear,park2008data,Pons-Moll:Siggraph2015,loper2015smpl,kadlecek-16-reconstructing,Meekyoung:siggraph}.
Capture approaches that use a statistical body model typically ignore clothing or treat it as noise~\cite{balan2007detailed} or explicitly estimate the shape under the apparel~\cite{bualan2008naked,ZhangCVPR2017,yang2016estimation}.
The off-line human performance capture approach of \citet{MuVS:3DV:2017} fits the SMPL body model to 2D joint detections and silhouettes in multi-view data.
Some of the recent off-line multi-view approaches jointly track facial expressions \cite{Joo2018TotalCA} and hands \cite{MANO:SIGGRAPHASIA:2017,Joo2018TotalCA}.
Even these approaches do not reconstruct dynamic hair.
To capture the geometry of the actor beyond the body shape an option is to non-rigidly deform the base model to fit a scan~\cite{ZhangCVPR2017} or a set of images~\cite{rhodin2016general}.
Recently, the approach of~\citet{Pons-Moll:Siggraph2017} can jointly capture body shape and clothing using separate meshes; very realistic results are achieved with this method, but it requires an expensive multi-view active stereo setup.
All the aforementioned approaches require multi-view setups and are not practical for consumer use.
Furthermore, none of the methods runs at real-time frame rates.
\paragraph{Depth-based}
With the availability of affordable depth camera sensors such as the Kinect, a large number of depth-based methods emerged.
Recent approaches that are based on a single depth camera, such as KinectFusion, enable the reconstruction of 3D rigid scenes~\cite{izadi2011kinectfusion,newcombe2011kinectfusion} and also appearance models~\cite{zhou2014color} by incrementally fusing geometry in a canonical frame.
The approach proposed in~\cite{Newcombe_2015_CVPR} generalized KinectFusion to capture dynamic non-rigid scenes.
The approach alternates non-rigid registration of the incoming depth frames with updates to the incomplete template, which is constructed incrementally.
Such template free methods~\cite{newcombe2011kinectfusion,slavcheva2017killingfusion,innmann2016volume,guo2017real} are flexible, but are limited to capturing slow and careful motions.
One way to make fusion and tracking more robust is by using a combination of a high frame rate/low resolution and a low frame rate/high resolution depth sensor~\cite{twinfusionguo}, improved hardware and software components~\cite{NeedForSpeedKowdle}, multiple Kinects or similar depth sensors~\cite{Ye2012,dou2016fusion4d,orts2016holoportation,Dou:2017,zhang2014depth}, or multi-view data~\cite{inria_2017,collet2015high,prada2017spatiotemporal} and registering new frames to a neighboring key frame; such methods achieve impressive reconstructions, but do not register all frames to the same canonical template and require complicated capture setups.
Another way to constrain the capture is to pre-scan the object or person to be tracked~\cite{zollhoefer2014deformable,de2008performance,Ye2012}, reducing the problem to tracking the non-rigid deformations.
Constraining the motion to be articulated is also shown to increase robustness~\cite{BodyFusion,DoubleFusion2018}.
Some works use simple human shape or statistical body models~\cite{wei2012accurate,weiss2011home,Helten:2013,ye2014real,zhang2014quality,Bogo:ICCV:2015}, some of which exploit the temporal information to infer shape.
Typically, a single shape and multiple poses are optimized to exploit the temporal information.
Such approaches are limited to capture naked human shape or at best very tight clothing.
Depth sensors are affordable and more practical than multi-view setups.
Unfortunately, they have a high power consumption, do not work well under general illumination and most of the media content is still in the format of 2D images and video.
Furthermore, depth-based methods do not directly generalize to work with monocular video.
\paragraph{Monocular 3D Pose and Shape Estimation}
Most methods to infer 3D human motion from monocular images are based on convolutional neural networks (CNNs) and leverage 2D joint detections and predict 3D joint pose in the form of stick figures, e.g., ~\cite{popa2017deep,zhou2017towards,sun2017compositional,tome2017lifting,rogez_lcr_cvpr17}.
\citet{Tekin2016} directly predict the 3D body pose from a rectified spatio-temporal volume of input frames.
The approach of \citet{TekinMSF17} learns to optimally fuse 2D and 3D image cues.
These approaches do not capture the dense deforming shape.
We also leverage a recent CNN-based 3D pose estimation method~\cite{VNect_SIGGRAPH2017}, but we only employ it to regularize the skeletal motion estimation.
Some works fit a (statistical) body surface model to images using substantial manual interaction~\cite{zhou2010parametric,jain2010movie,rogge2014garment,guan2009estimating} typically for the task of image manipulation.
Shape and clothing is recovered in~\cite{guo2012clothed,chen2013deformable}, but the user needs to click points in the image, select the clothing types from a database and dynamics are not captured.
Instead of clicked points, \citet{kraevoy2009modeling} propose to obtain the shape from contour drawings.
With the advance of 2D joint detections, the works of~\cite{bogo2016smpl,Lassner,hmrKanazawa17} fit a 3D body model~\cite{loper2015smpl} to them; since only model parameters are optimized, the results are constrained to the shape space.
More recent work~\cite{varol18_bodynet} directly regresses a coarse volumetric body shape.
Correspondences from pixels of an input image to surface points on the SMPL body model can also be directly regressed \cite{gue2018densepose}.
Capturing 3D non-rigid deformations from monocular video is very hard.
In the domain of non-rigid structure from motion, model-free methods using rigidity and temporal smoothness priors can capture coarse 3D models of simple motions and medium-scale deformations~\cite{Garg_2013_CVPR,Russell2014}.
Some methods~\cite{Salzmann2011,Bartoli2015,Yu_2015_ICCV} can non-rigidly track simple shapes and motions by off-line template fitting; but they were not shown to handle highly articulated fast body motions, including clothing, as we do.
Specifically for faces, monocular performance capture methods were presented, for example~\cite{Garrido:2016,Cao:2015:RHF}.
However, monocular full-body capture faces additional challenges due to more frequent (self-)occlusions and much more complex and diverse clothing and appearance.
To the best of our knowledge, the only approach that has shown 3D performance capture of the human body including the non-rigid deformation of clothing from monocular video is the approach of~\citet{xu17MonoPerfCap}.
Its space-time formulation can resolve difficult self-occluded poses at the expense of temporally oversmoothing the actual motion.
But at over 1 minute runtime per frame, it is impractical for many applications such as virtual try-on, gaming or virtual teleportation.
It is also challenged by starkly non-rigidly moving clothing.
Reducing the processing time without compromising accuracy introduces challenges in formulation and implementation of model-based performance capture, which we address in this work.
We present, for the first time, a real-time full-body performance capture system that just requires a monocular video as input.
We show that it comes close in accuracy to the best off-line monocular and even multi-view methods, while being orders of magnitude faster.
\begin{figure*}[t]
	\includegraphics[width=\linewidth]{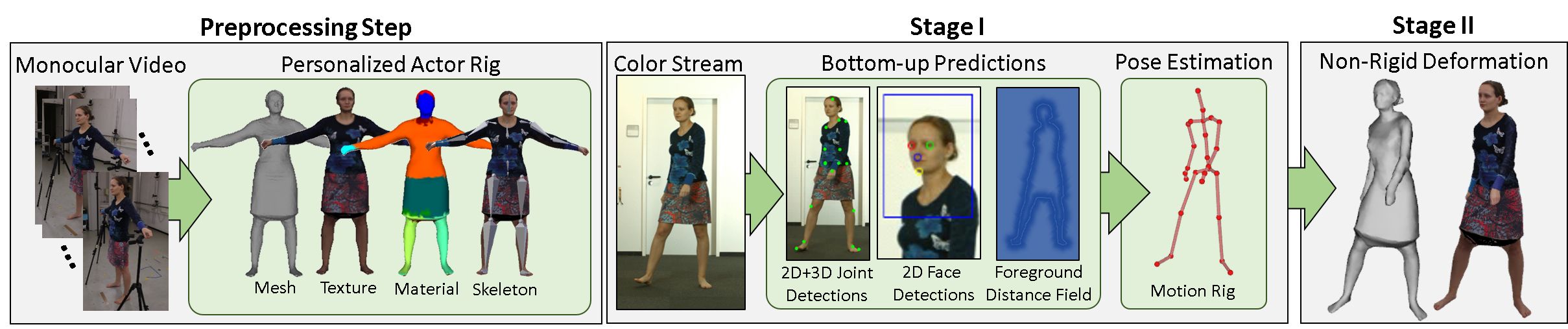}
	\caption
	{
		Our real-time performance capture approach reconstructs dense, space-time coherent deforming geometry of people in loose everyday clothing from just a single RGB stream.
		A skinned template is jointly fit to background subtracted input video, 2D and 3D joint estimates, and sparse facial detections.
		Non-rigid 3D deformations of skin and even loose apparel are captured based on a novel real-time capable dense surface tracker.
	}
	\label{fig:pipeline}
\end{figure*}
\section{Method}
The input to our method is a single color video stream.
In addition, our approach requires a textured actor model, which we acquire in a preprocessing step (Sec.~\ref{sec:model}) from a monocular video sequence.
From this input alone, our real-time human performance capture approach automatically estimates the articulated actor motion and the non-rigid deformation of skin and clothing coarse-to-fine in two subsequent stages per input frame.
In the first stage, we estimate the articulated 3D pose of the underlying kinematic skeleton.
To this end, we propose an efficient way to fit the skeletal pose of the skinned template to 2D and 3D joint positions from a state-of-the-art CNN-based regressor, to sparse detected face landmarks, and to the foreground silhouette (Sec.~\ref{sec:estimation}).
With this skeleton-deformed mesh and the warped non-rigid displacement of the previous frame as initialization, the second stage captures the surface deformation of the actor using a novel real-time template-to-image non-rigid registration approach (Sec.~\ref{sec:nonrigid}).
We express non-rigid registration as an optimization problem consisting of a silhouette alignment term, a photometric term, and several regularization terms; the formulation and combination of terms in the energy is geared towards high efficiency at high accuracy despite the monocular ambiguities.
The different components of our approach are illustrated in Fig.~\ref{fig:pipeline}.
In order to achieve real-time performance, we tackle the underlying optimization problems based on dedicated data-parallel GPU optimizers (Sec.~\ref{sec:optimization}).
In the following, we explain all components.

\subsection{Actor Model Acquisition} \label{sec:model}
Similar to many existing template-based performance capture methods, e.g., \cite{cagniart2010free,InriaVolumetric_2015,gall2009motion,vlasic2008articulated,xu17MonoPerfCap}, we reconstruct an actor model in a preprocessing step.
To this end, we take a set of $M$ images $\mathcal{I}_{\mathrm{rec}} = \{I_{\mathrm{rec}_1}, \cdots, I_{\mathrm{rec}_M}\}$ of the actor in a static neutral pose from a video captured while walking around the person, which covers the entire body.
For all our templates we used around $M=70$ images.
With these images, we generate a triangulated template mesh $\hat{\mathbf{V}} \in \mathbb{R}^{N \times 3}$ ($N$ denotes the number of vertices in the mesh) and the associated texture map of the actor using an image-based 3D reconstruction software\footnote{\url{http://www.agisoft.com}}.
We downsample the reconstructed geometry to a resolution of approximately $N=5000$ by using the Quadric Edge Collapse Decimation algorithm implemented in MeshLab\footnote{\url{http://www.meshlab.net/}}.
The vertex colors of the template mesh $\mathbf{C} \in \mathbb{R}^{N \times 3}$ are transferred from the generated texture map.
Then, skeleton joints and facial markers are manually placed on the template mesh resulting in a skeleton model.
The template mesh is rigged to this skeleton model via dual quaternion skinning~\cite{Kavan2007SDQ}, where the skinning weights are automatically computed using Blender\footnote{\url{https://www.blender.org/}} (other auto-rigging tools would be feasible).
This allows us to deform the template mesh using the estimated skeletal pose parameters (Sec.~\ref{sec:estimation}).
An important feature of our performance capture method is that we model material-dependent differences in deformation behavior, e.g. of skin and apparel during tracking (see Sec.~\ref{sec:nonrigid}).
To this end, we propose a new multi-view method to segment the template into one of seven non-rigidity classes.
We first apply the state-of-the-art human parsing method of~\citet{Gong_2017_CVPR} to each image in $\mathcal{I}_{\mathrm{rec}}$ separately to obtain the corresponding semantic label images $\mathcal{L}_{\mathrm{rec}} = \{L_{\mathrm{rec}_1}, \cdots, L_{\mathrm{rec}_M}\}$.
The semantic labels $L \in \{1,\cdots, 20\}^{N}$ for all vertices $\mathbf{V}_i$ are computed based on their back-projection into all label images, and a majority vote per vertex.
The materials are binned into 7 non-rigidity classes, each one having a different per-edge non-rigidity weight in the employed regularization term (Sec.~\ref{sec:nonrigid}).
Those weights were empirically determined by visual observation of the deformation behaviour under different weighting factors.
The different classes and the corresponding non-rigidity weights are shown in Tab.~\ref{tab:weights}.
We use a very high weight for rigid body parts, e.g., the head, medium weights for the less rigid body parts, e.g., skin and tight clothing, and a low weight for loose clothing.
We use a high rigidity weight for any kind of hair style, since we do not, similar to all other human performance capture approaches, consider and track hair dynamics.
We map the per-vertex smoothness weights to per-edge non-rigidity weights $s_{i,j}$ by averaging the weights of vertex $\mathbf{V}_i$ and $\mathbf{V}_j$.
\begin{table}
	\centering
	\caption{The employed non-rigidity weights $s_{i,j}$.}
	\begin{tabular}{ | c | c | l | }
		\hline
		Class ID & Weight & Part/Apparel Type\\ \hline \hline
		1 & 1.0 & dress, coat, jumpsuit, skirt, background \\ \hline
		2 & 2.0 & upper clothes\\ \hline
		3 & 2.5 & pants\\ \hline
		4 & 3.0 & scarf \\ \hline
		5 & 50.0 & left leg, right leg, left arm, right arm, socks\\ \hline
		6 & 100.0 & hat, glove, left shoe, right shoe,  \\ \hline
		7 & 200.0 & hair, face, sunglasses \\ \hline
	\end{tabular}
	\label{tab:weights}
\end{table}
\subsection{Input Stream Processing} \label{sec:input}
After the actor model acquisition step, our real-time performance capture approach works fully automatically and we do not rely on a careful initialization, e.g. it is sufficient to place the T-posed character model in the center of the frame.
The input to our algorithm is a single color video stream from a static camera, e.g., a webcam.
Thus, we assume camera and world space to be the same.
We calibrate the camera intrinsics using the Matlab calibration toolbox\footnote{\url{http://www.vision.caltech.edu/bouguetj/calib_doc}}.
Our skeletal pose estimation and non-rigid registration stages rely on the silhouette segmentation of the input video frames.
To this end, we leverage the background subtraction method of~\citet{Zivkovic:2006:EAD:1142319.1142328}.
We assume the background is static, that its color is sufficiently different from the foreground, and a few frames of the empty scene are recorded before performance capture commences.
We efficiently compute distance transform images $I_\mathrm{DT}$ from the foreground silhouettes, which are used in the skeletal pose estimation and non-rigid alignment step.
\subsection{Skeletal Pose Estimation} \label{sec:estimation}
We formulate skeletal pose estimation as a non-linear optimization problem in the unknown skeleton parameters $\mathcal{S}^*$:
\begin{equation}
\mathcal{S}^* = \argmin_{\mathcal{S}}{E_\mathrm{pose}(\mathcal{S})}.
\end{equation}
The set $\mathcal{S}=\{\mathbf{\theta}, \mathbf{R}, \mathbf{t}\}$ contains the joint angles $\mathbf{\theta} \in \mathbb{R}^{27}$ of the $J$ joints of the skeletal model, and the global pose $\mathbf{R} \in \mathbf{SO}(3)$ and translation $\mathbf{t} \in \mathbb{R}^{3}$ of the root.
For pose estimation, we optimize an energy of the following general form:
\begin{align}
\begin{aligned}
 E_\mathrm{pose}(\mathcal{S}) = &E_\mathrm{2D}(\mathcal{S}) +  E_\mathrm{3D}(\mathcal{S}) + E_\mathrm{silhouette}(\mathcal{S}) +  E_\mathrm{temporal}(\mathcal{S}) \\
 + &E_\mathrm{anatomic}(\mathcal{S}) \enspace{.}
 \end{aligned}
\end{align}
Here, $E_\mathrm{2D}$ and $E_\mathrm{3D}$ are alignment constraints based on regressed 2D and 3D joint positions, respectively.
In addition, $E_\mathrm{silhouette}$ is a dense alignment term that fits the silhouette of the actor model to the detected silhouette in the input color images.
At last, $E_\mathrm{temporal}$ and $E_\mathrm{anatomic}$ are temporal and anatomical regularization constraints that ensure that the speed of the motion and the joint angles stay in physically plausible ranges.
To better handle fast motion, we initialize the skeleton parameters before optimization by extrapolating the poses of the last two frames in joint angle space based on an explicit Euler step.
In the following, we explain each energy term in more detail.
\paragraph{Sparse 2D and 3D Alignment Constraint}
For each input frame $I$, we estimate the 2D and 3D joint positions $\mathbf{P}_{\mathrm{2D},i} \in \mathbb{R}^{2}$ and $\mathbf{P}_{\mathrm{3D},i} \in \mathbb{R}^{3}$ of the $J$ joints using the efficient deep skeleton joint regression network of the VNect algorithm~\cite{VNect_SIGGRAPH2017} trained with the original data of~\cite{VNect_SIGGRAPH2017}.
However, with these skeleton-only joint detections, it is not possible to determine the orientation of the head.
Therefore, we further augment the 2D joint predictions of~\cite{VNect_SIGGRAPH2017} with a subset of the facial landmark detections of~\cite{saragig_tracker}, which includes the eyes, nose and chin.
We incorporate the 2D detections $\mathbf{P}_{\mathrm{2D},i} \in \mathbb{R}^{2}$ based on the following re-projection constraint: 
\begin{align}
	\begin{aligned}
		E_\mathrm{2D}(\mathcal{S}) = \lambda_\textrm{2D}\sum_{i=1}^{J+4}
		\lambda_{i}
		\left\lVert
		\pi \left( p_{\mathbf{3D},i}(\mathbf{\mathbf{\theta}},\mathbf{R}, \mathbf{t}) \right)- \mathbf{P}_{\mathrm{2D},i}
		\right\rVert^2 \enspace{.}
	\end{aligned}
\end{align}
Here, $p_{\mathbf{3D},i}$ is the 3D position of the $i$-th joint/face marker of the used kinematic skeleton and $\pi:\mathbb{R}^3 \rightarrow \mathbb{R}^2$ is a full perspective projection that maps 3D space to the 2D image plane.
Thus, this term enforces that all projected joint positions are close to their corresponding detections.
$\lambda_{i}$ are detection-based weights.
We use $\lambda_{i}=0.326$ for the facial landmarks and $\lambda_{i}=1.0$ for all other detections to avoid that the head error dominates all other body parts.
To resolve the inherent depth ambiguities of the re-projection constraint, we also employ the following 3D-to-3D alignment term between model joints $p_{\mathbf{3D},i}(\mathbf{\mathbf{\theta}},\mathbf{R}, \mathbf{t})$ and 3D detections $\mathbf{P}_{\mathrm{3D},i}$: 
\begin{align}
	\begin{aligned}
		E_\mathrm{3D}(\mathcal{S}) = \lambda_\textrm{3D} \sum_{i =1 }^J
		\left\lVert
		p_{\mathbf{3D},i}(\mathbf{\mathbf{\theta}},\mathbf{R}, \mathbf{t}) - \left(\mathbf{P}_{\mathrm{3D},i} + \mathbf{t}' \right)
		\right\rVert^2 \enspace{.}
	\end{aligned}
\end{align}
Here, $\mathbf{t}'\in \mathbb{R}^3$ is an auxiliary variable that transforms the regressed 3D joint positions $\mathbf{P}_{\mathrm{3D},i}$ from the root centered local coordinate system to the global coordinate system.
Note that the regressed 3D joint positions $\mathbf{P}_{\mathrm{3D},i}$ are in a normalized space.
Therefore, we rescale the regressed skeleton according to the bone lengths of our parameterized skeleton model.
\paragraph{Dense Silhouette Alignment Constraint}
\begin{figure}[t]
	\includegraphics[width=\linewidth]{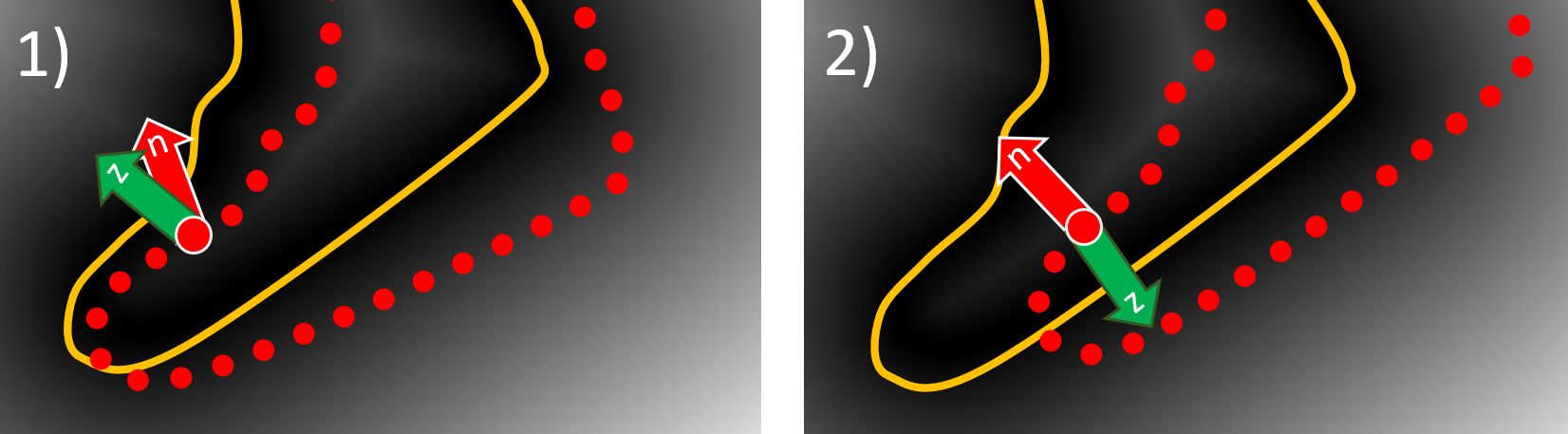}
	\caption
	{
		The two cases in the silhouette alignment constraint.
		Target silhouette (yellow), model silhouette (red), negative gradient of the distance field $\mathbf{z}$ (green arrow), and the projected 2D normal $\mathbf{n}$ of the boundary vertex (red arrow).
	}
	\label{fig:sil}
\end{figure}
We enforce a dense alignment between the boundary of the skinned actor model and the detected silhouette in the input image.
In contrast to the approach of \citet{xu17MonoPerfCap} that requires closest point computations we employ a distance transform based constraint for efficiency reasons.
Once per frame, we extract a set of contour vertices $\mathcal{B}$ from the current deformed version of the actor model.
Afterwards, we enforce that all contour vertices align well to the interface between the detected foreground and background:
\begin{equation}
E_\mathrm{silhouette}(\mathcal{S}) = \lambda_\textrm{silhouette}\sum_{i \in \mathcal{B}}{  b_i \cdot \Big[ I_\mathrm{DT}\big(\pi(\mathbf{V}_i(\mathbf{\theta},\mathbf{R}, \mathbf{t}))\big)\Big]^2} \enspace{.}
\label{Eq:sillhouette}
\end{equation}
Here, $\mathbf{V}_i$ is the $i$-th boundary vertex of the skinned actor model and the image $I_{DT}$ stores the Euclidean distance transform with respect to the detected silhouette in the input image.
The $b_i \in \{-1, +1\}$ are directional weights that guide the optimization to follow the right direction in the distance field.
In the minimization of the term in Eq.~\ref{Eq:sillhouette}, silhouette model points are pushed in the negative direction of the distance transform image gradient $\mathbf{z} = - \nabla_{xy} I_{DT} \in \mathbb{R}^2$.
By definition, $\mathbf{z}$ points in the direction of the nearest \emph{image silhouette} (IS) contour.
If model points fall outside of the IS they will be dragged towards the nearest IS contour as desired.
When model points fall inside the IS however there are two possibilities:
1) the model point normal $\mathbf{n}$ follows roughly the same direction as $\mathbf{z}$ or 2) it does not.
In case 1) the normal at the nearest IS point matches the direction of the model point normal.
This indicates that $\mathbf{z}$ is a good direction to follow.
In case 2) however, the normal at the nearest IS point follows the opposite direction, indicating that $\mathbf{z}$ is pointing towards the wrong IS contour, see Fig.~\ref{fig:sil}.
Therefore, in case 2) we follow the opposite direction $\mathbf{p}= - \mathbf{z}$ by setting $b_i = -1$.
This is preferable over just following $\mathbf{n}$, since $\mathbf{n}$ is not necessarily pointing away from the wrong IS contour.
Mathematically, we consider that we are in case 2) when $\mathbf{n}^T \mathbf{z} < 0$.
For all the other cases, we follow the direction of $\mathbf{z}$ by setting $b_i = +1$.
\paragraph{Temporal Stabilization}
To mitigate temporal noise, we use a temporal stabilization constraint, which penalizes the change in joint position between the current and previous frame:
\begin{align}
\begin{aligned}
E_\mathrm{temporal}(\mathcal{S}) = \lambda_\textrm{temporal}\sum_{i =1}^{J}
		   \lambda_{i}
		   \left\lVert
           		p_{\mathbf{3D},i}(\mathbf{\theta},\mathbf{R}, \mathbf{t})
           		-
           		p_{\mathbf{3D},i}^{t-1}(\mathbf{\theta},\mathbf{R}, \mathbf{t})
           \right\rVert^2 .
\end{aligned}
\end{align}
Here the $\lambda_{i}$ are joint-based temporal smoothness weights.
We use $\lambda_i = 2.5$ for joints on the torso and the head, $\lambda_i = 2.0$ for shoulders, $\lambda_i = 1.5$ for knees and elbows, and $\lambda_i = 1.0$ for the hands and feet.
\paragraph{Joint Angle Limits}
The joints of the human skeleton have physical limits.
We integrate this prior knowledge into our pose estimation objective based on a soft-constraint on $\mathbf{\theta} \in \mathbb{R}^{27}$.
To this end, we enforce that all degrees of freedom stay within their anatomical limits $\mathbf{\theta}_\mathrm{min} \in \mathbb{R}^{27}$ and $\mathbf{\theta}_\mathrm{max} \in \mathbb{R}^{27}$:
$$
E_\mathrm{anatomic}(\mathcal{S}) = \lambda_\mathrm{anatomic} \sum_{i=1}^{27}{ \Psi( \theta_i ) } \enspace{.}
$$
Here, $\Psi(x)$ is a quadratic barrier function that penalizes if a degree of freedom exceeds its limits:
$$
\Psi(x)
=
\begin{cases}
(x - \mathbf{\theta}_{\mathrm{max},i})^2,\text{ if } x > \mathbf{\theta}_{\mathrm{max},i}\\
(\mathbf{\theta}_{\mathrm{min},i} - x)^2 \, ,\text{ if } x < \mathbf{\theta}_{\mathrm{min},i}\\
0 \qquad \qquad \; \; \; ,  \text{ otherwise} \enspace{.}
\end{cases}
$$
This term prevents un-plausible human pose estimates.
\subsection{Non-rigid Surface Registration} \label{sec:nonrigid}
\begin{figure}[t]
	\includegraphics[width=\linewidth]{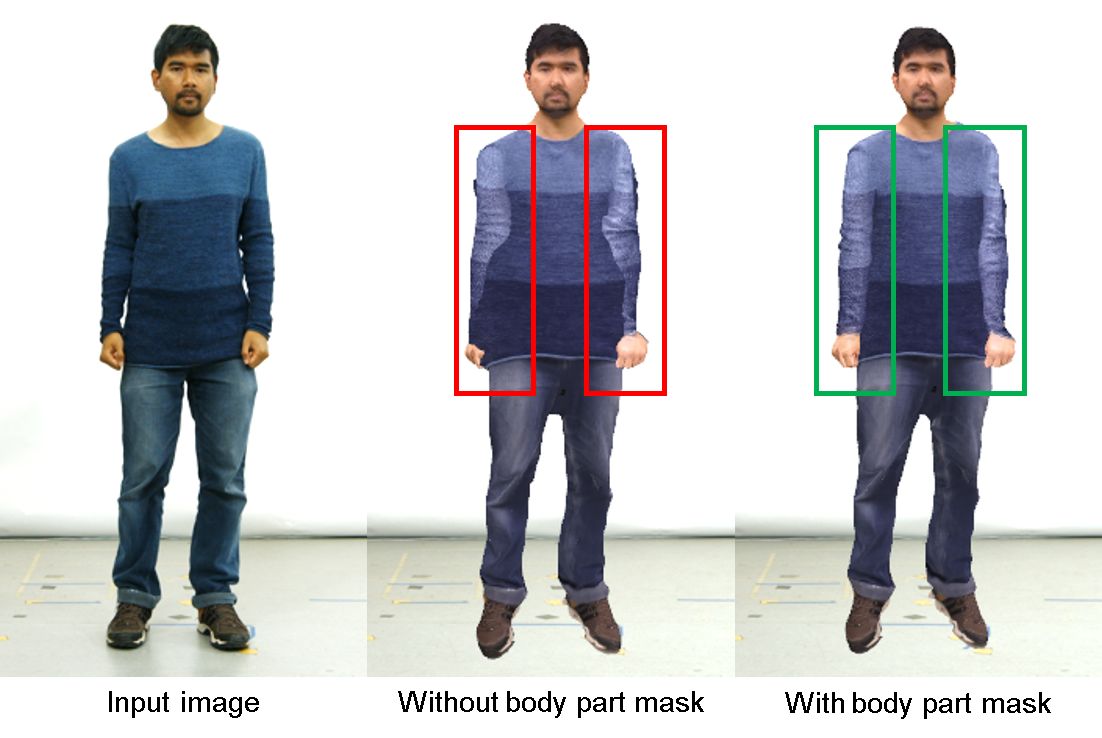}
	\caption
	{
		Left: Input image.
		Middle: Textured reconstruction without using the body part mask.
		One can clearly see the artifacts since multiple model boundaries wrongly explain the silhouette of the arms.
		Right: Using the body part mask in the distance transform image the foreground silhouette is correctly explained.
	}
	\label{fig:AblationBodySeg}
\end{figure}

The pose estimation step cannot capture realistic non-rigid deformations of skin and clothing that are not explained through skinning.
The model therefore does not yet align with the image well everywhere, in particular in cloth and some skin regions.
Hence, starting from the pose estimation result, we solve the following non-rigid surface tracking energy:
\begin{equation}
E_\mathrm{non-rigid}(\mathbf{V}) =  E_\mathrm{data}(\mathbf{V}) + E_\mathrm{reg}(\mathbf{V}) \enspace{.}
\end{equation}
The energy consists of several data terms $E_\mathrm{data}$ and regularization constraints $E_\mathrm{reg}$, which we explain in the following.
Our data terms are a combination of a dense photometric alignment term $E_\mathrm{photo}$ and a dense silhouette alignment term $E_\mathrm{silhouette}$:
\begin{equation}
E_\mathrm{data}(\mathbf{V}) = E_\mathrm{photo}(\mathbf{V}) + E_\mathrm{silhouette}(\mathbf{V}) \enspace{.}
\end{equation}
\paragraph{Dense Photometric Alignment}
The photometric alignment term measures the re-projection error densely:
\begin{equation}
E_\mathrm{photo}(\mathbf{V}) = \sum_{i \in \mathcal{V}} w_\mathrm{photo}
		  \left\lVert 
		  \sigma_c \left( I_\mathrm{Gauss} \left( \pi \left( \mathbf{V}_i \right) \right) - \mathbf{C}_{i}  \right) 
		  \right\rVert^2 ,
\end{equation}
where $\mathbf{C}_{i}$ is the color of vertex $\mathbf{V}_i$ in the template model and $\sigma_c(\cdot)$ is a robust kernel that prunes wrong correspondences according to color similarity by setting residuals that are above a certain threshold to zero.
More specifically, we project every visible vertex $\mathbf{V}_i \in \mathcal{V}$ to screen space based on the full perspective camera model $\pi$.
The visibility is obtained based on the skinned mesh after the pose estimation step using depth buffering.
In order to speed up convergence, we compute the photometric term based on a 3-level pyramid of the input image.
We perform one Gauss-Newton iteration on each level.
We use the projected positions to sample a Gaussian blurred version $I_\mathrm{Gauss}$ of the input image $I$ at the current time step, for more stable and longer range gradients.
The Gaussian kernel sizes for the 3 levels are 15, 9 and 3 respectively.
\paragraph{Dense Silhouette Alignment}
In addition to dense photometric alignment, we also enforce alignment of the projected 3D model boundary with the detected silhouette in the input image:
\begin{align}
\begin{aligned}
E_\mathrm{silhouette}(\mathbf{V}) = w_\mathrm{silhouette} \sum_{i \in \mathcal{B}}
		  b_i \cdot \Big[ I_\mathrm{DT} \big( \pi \left( \mathbf{V}_i \right) \big) \Big]^2 .
\end{aligned}
\end{align}
After Stage I, we first update the model boundary $\mathcal{B}$ and consider all vertices $\mathbf{V}_i \in \mathcal{B}$.
These boundary vertices are encouraged to match the zero iso-line of the distance transform image $I_\mathrm{DT}$, and thus be aligned with the detected input silhouette.
The $b_i$ are computed similar to the pose optimization step (see Sec.~\ref{sec:estimation}).
Due to the non-rigid deformation that cannot be recovered by our pose estimation stage, in some cases the projection of the mesh from Stage I has a gap between body parts such as arms and torso, while in the input image the gaps do not exist.
To prevent image silhouettes being wrongly explained by multiple model boundaries we project the posed model $\mathbf{V}^\mathrm{S}$ into the current frame and compute a body part mask
--- derived from the skinning weights.
We increase the extent of each body part by a dilation (maximum of 10 pixels, the torso has preference over the other parts) to obtain a conservative region boundary that closes the above mentioned gaps.
If a vertex $\mathbf{V}_i$ moves onto a region with a differing semantic label, we disable its silhouette term by setting $b_i=0$.
This drastically improves the reconstruction quality (see Fig.~\ref{fig:AblationBodySeg}).
\par
Our high-dimensional monocular non-rigid registration problem with only the data terms is ill-posed.
Therefore, we use regularization constraints:
\begin{equation}
E_\mathrm{reg}(\mathbf{V}) = E_\mathrm{smooth}(\mathbf{V}) + E_\mathrm{edge}(\mathbf{V}) + E_\mathrm{velocity}(\mathbf{V}) + E_\mathrm{acceleration}(\mathbf{V}) \enspace{.}
\end{equation}
Here, $E_\mathrm{smooth}$ and $E_\mathrm{edge}$ are spatial smoothness priors on the mesh geometry, and $E_\mathrm{velocity}$ and $E_\mathrm{acceleration}$ are temporal priors.
In the following, we give more details.
\paragraph{Spatial Smoothness}
The first prior on the mesh geometry is a spatial smoothness term with respect to the pose estimation result:
\begin{equation}
E_\mathrm{smooth} \left( \mathbf{V} \right) = w_\mathrm{smooth}
\sum_{i=1}^{N}
	 \sum_{j \in \mathcal{N}_i} \frac{s_{ij}}{|\mathcal{N}_i|} \left\lVert
		 \left( \mathbf{V}_i -  \mathbf{V}_j \right)-
		( \mathbf{V}_i^\mathrm{S} - \mathbf{V}_j^\mathrm{S} ) 
        \right\rVert^2 \enspace{.}
\end{equation}
Here, the $\mathbf{V}_i$ are the unknown optimal vertex positions and the $\mathbf{V}_i^\mathrm{S}$ are vertex positions after skinning using the current pose estimation result of Stage I.
$s_{ij}$ are the semantic label based per-edge smoothness weights (see Sec.~\ref{sec:model}) that model material dependent non-rigidity.
The energy term enforces that every edge in the deformed model is similar to the undeformed model in terms of its length and orientation.
In addition to this surface smoothness term, we also enforce locally isometric deformations:
\begin{equation}
E_\mathrm{edge} \left( \mathbf{V}  \right) = w_\mathrm{edge}
\sum_{i=1}^{N} 
	\sum_{j \in \mathcal{N}_i} \frac{s_{ij}}{|\mathcal{N}_i|}
		\left(\left\lVert \mathbf{V}_i - \mathbf{V}_j \right\rVert - 
		\left\lVert \hat{\mathbf{V}}_i - \hat{\mathbf{V}}_j \right\rVert
	\right)^2\text{,}
\end{equation}
where $\hat{\mathbf{V}}$ denotes the vertex position in the template's rest pose.
We enforce that the edge length does not change much between the rest pose $\hat{\mathbf{V}}_i$ and the optimal unknown pose $\mathbf{V}_i$.
While this is similar to the first term, it enables us to penalize stretching independently of shearing.
\paragraph{Temporal Smoothness}
We also use temporal priors that favor temporally coherent non-rigid deformations.
Similar to temporal smoothness in skeletal pose estimation, the first term
\begin{equation}
E_\mathrm{velocity} \left( \mathbf{V} \right) = w_\mathrm{velocity}
	\sum_{i=1}^{N} 
		\left\lVert \mathbf{V}_i - \mathbf{V}_i^{t-1} \right\rVert ^2 \enspace{,}
\end{equation}
encourages small velocity and the second term
\begin{equation}
E_\mathrm{acceleration} \left( \mathbf{V} \right) = w_\mathrm{acceleration}
	\sum_{i=1}^{N}
		\left\lVert 
	    \mathbf{V}_i -2\mathbf{V}_i^{t-1} + \mathbf{V}_i^{t-2} \right 
		\rVert^2 \enspace{,}
\end{equation}
encourages small acceleration between adjacent frames.
\paragraph{Displacement Warping}
The non-rigid displacements $\mathbf{d}_i^{t-1} =\mathbf{V}_i^{t-1}- \mathbf{V}_i^{{\mathrm{S},t-1}} \in \mathbb{R}^{3}$ that are added to each vertex $i$ after skinning are usually similar from frame $t-1$ to frame $t$.
We warp $\mathbf{d}_i^{t-1}$ back to the rest pose by applying Dual Quaternion skinning with the inverse rotation quaternions given by the pose at time $t-1$.
We refer to them as $\hat{\mathbf{d}}_i^{t-1}$.
For the next frame $t$, we transform $\hat{\mathbf{d}}_i^{t-1}$ according to the pose at time $t$ resulting in a skinned displacement $\mathbf{d}_i^{\mathrm{S},t}$ and initialize the non-rigid stage with $\mathbf{V}_i^{t} = \mathbf{V}_i^{{\mathrm{S},t}} + \mathbf{d}_i^{\mathrm{S},t}$.
This jump-starts the non-rigid alignment step and leads to improved tracking quality.
Similarly, we add $\mathbf{d}_i^{\mathrm{S},t}$ to the skinned actor model for more accurate dense silhouette alignment during the skeletal pose estimation stage.
\paragraph{Vertex Snapping}
After the non-rigid stage, the boundary vertices are already very close to the image silhouette.
Therefore, we can robustly snap them to the closest silhouette point by walking on the distance transform along the negative gradient direction until the zero crossing is reached.
Vertex snapping allows us to reduce the number of iteration steps, since if the solution is already close to the optimum, the updates of the solver become smaller, as is true for most optimization problems.
Therefore, if the mesh is already close to the silhouette, we `snap' it to the silhouette in a single step, instead of requiring multiple iterations of Gauss-Newton.
To obtain continuous results, non-boundary vertices are smoothly adjusted based on a Laplacian warp in a local neighborhood around the mesh contour.
\section{Data Parallel GPU Optimization} \label{sec:optimization}
The described pose estimation and non-rigid registration problems are non-linear optimizations based on an objective $E$ with respect to unknowns $\mathcal{X}$, i.e., the parameters of the kinematic model $\mathcal{S}$ for pose estimation and the vertex positions $\mathbf{V}$ for non-rigid surface deformation.
The optimal parameters $\mathcal{X}^{*}$ are found via energy minimization:
\begin{equation}
\label{eq:optimization}
\mathcal{X}^{*} = {\operatorname{arg\,min}}_{\mathcal{X}}{E}(\mathcal{X}) \enspace{.}
\end{equation}
In both capture stages, i.e. pose estimation (see Sec.~\ref{sec:estimation}) and non-rigid surface tracking (see Sec.~\ref{sec:nonrigid}), the objective $E$ can be expressed as a sum of squares:
\begin{equation}
E(\mathcal{X}) = \sum_{i}{\big[ \mathbf{F}_i(\mathcal{X}) \big]^2} = \big|\big|\mathbf{F}(\mathcal{X})\big|\big|_2^2 \enspace{.}
\end{equation}
Here, $\mathbf{F}$ is the error vector resulting from stacking all residual terms.
We tackle this optimization at real-time rates using a data-parallel iterative Gauss-Newton solver that minimizes the total error by linearizing $\mathbf{F}$ and taking local steps $\mathcal{X}_k \!=\! \mathcal{X}_{k-1} + \boldsymbol \delta_{k}^*$ obtained by the solution of a sequence of linear sub-problems (normal equations):
\begin{equation}
\mathbf{J}^T(\mathcal{X}_{\mathrm{k-1}}) \mathbf{J}(\mathcal{X}_{\mathrm{k-1}}) \cdot \boldsymbol\delta_{k}^* = -\mathbf{J}^T(\mathcal{X}_{\mathrm{k-1}}) \mathbf{F}(\mathcal{X}_{\mathrm{k-1}}) \enspace{.}
\label{Eq:gaussnewton}
\end{equation}
Here, $\mathbf{J}$ is the Jacobian of $\mathbf{F}$.
Depending on the problems (pose estimation or non-rigid registration), the linear systems have a quite different structure in terms of dimensionality and sparsity.
Thus, we use tailored parallelization strategies for each of the problems.
Since we use Gauss-Newton instead of Levenberg-Marquardt, the residual has not to be computed during the iterations, thus leading to faster runtimes and in consequence more iterations are possible within the tight real-time constraint.
\paragraph{Pose Estimation}
The normal equations of the pose optimization problem are small, but dense, i.e, the corresponding system matrix is small, rectangular and dense.
Handling each non-linear Gauss-Newton step efficiently requires a specifically tailored parallelization and optimization strategy.
First, in the beginning of each Gauss-Newton step, we compute the system matrix $\mathbf{J}^T\mathbf{J}$ and right hand side $-\mathbf{J}^T\mathbf{F}$ in global memory on the GPU.
Afterwards, we ship the small system of size $36 \times 36$ (36 = 3+3+27+3, 3 DoFs for $\mathbf{R}$, 3 for $\mathbf{t}$, 27 for $\theta$, and 3 for $\mathbf{t}'$) to the CPU and solve it based on QR decomposition.
The strategy of splitting the computation to CPU and GPU is in spirit similar to \cite{tagliasacchi2015robust}.
To compute $\mathbf{J}^T\mathbf{J}$ on the GPU, we first compute $\mathbf{J}$ fully in parallel and store it in device memory based on a kernel that launches one thread per matrix entry.
We perform a similar operation for $\mathbf{F}$.
$\mathbf{J}^T\mathbf{J}$ is then computed based on a data-parallel version of a matrix-matrix multiplication that exploits shared memory for high performance.
The same kernel also directly computes $\mathbf{J}^T\mathbf{F}$.
We launch several thread blocks per element of the output matrix/vector, which cooperate in computing the required dot products, e.g., between the i-th and j-th column of $\mathbf{J}$ or the i-th column of $\mathbf{J}$ and $\mathbf{F}$.
To this end, each thread block computes a small subpart of the dot product based on a shared memory reduction.
The per-block results are summed up based on global memory atomics.
In total, we perform 6 Gauss-Newton steps, which turned out to be a good trade-off between accuracy and speed.
\paragraph{Non-rigid Surface Registration}
The non-rigid optimization problem that results from the energy $E_\textrm{non-rigid}$ has a substantially different structure.
It leads to a large sparse system of normal equations, i.e, the corresponding system matrix is sparse and has a low number of non-zeros per row.
Similar to \cite{zollhoefer2014deformable,innmann2016volume}, during GPU-based data-parallel Preconditioned Conjugate Gradient (PCG) we parallelize over the rows (unknowns) of the system matrix $\mathbf{J}^T\mathbf{J}$ using one thread per block row (x-,y-, and z-entry of a vertex).
Each thread collects and handles all non-zeros in the corresponding row.
We use the diagonal of $\frac{1}{\mathbf{J}^T\mathbf{J}}$ as a preconditioner.
We perform 3 Gauss-Newton steps and solve the linear system based on 4 PCG iterations, which turned out to be a good trade-off between accuracy and speed.
\begin{figure*}[p]
	\includegraphics[width=0.9\textwidth]{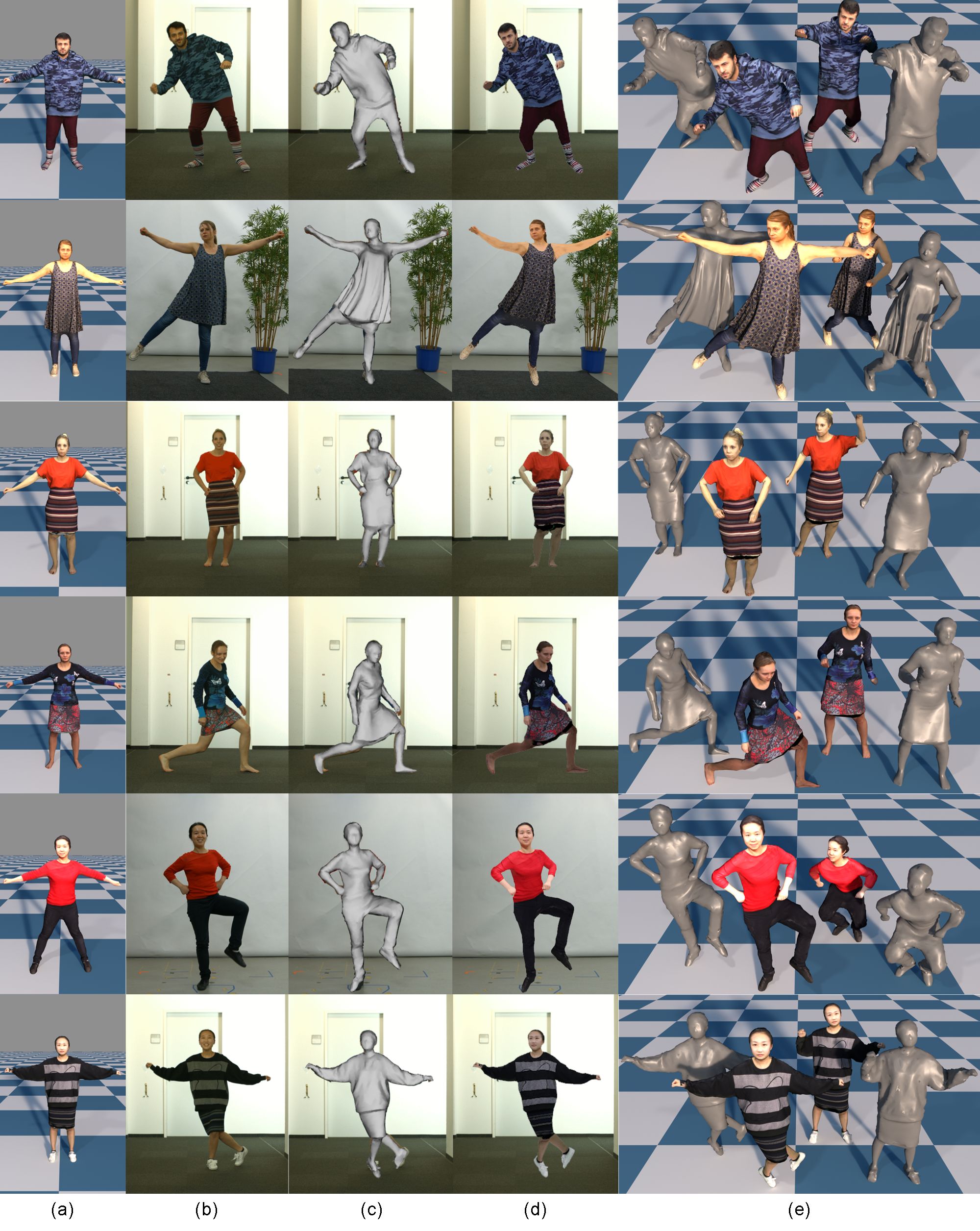}
	\caption
	{
		Qualitative results.
		We show several live monocular performance capture results of entire humans in their loose everyday clothing.
		(a) shows the template models.
		(b) shows input images to our method, while (c) shows that the corresponding results precisely overlay the person in the input images.		
		Our results can be used to render realistic images (d) or free viewpoint video (e).
	}
	\label{fig:qualitative}
\end{figure*}
\paragraph{Pipelined Implementation}
To achieve real-time performance, we use a data-parallel implementation of our entire performance capture algorithm in combination with a pipeline strategy tailored for our problem.
To this end, we run our approach in three threads on a PC with two GPUs.
Thread 1 uses only the CPU, which is responsible for data preprocessing.
Thread 2 computes the CNN-based human pose detection on the first graphics card, 
thread 3 solves the pose optimization problem and estimates the non-rigid deformation on the second graphics card.
Our distributed computation strategy induces a 2 frame delay, but for most applications it is barely noticeable.
\section{Results} \label{sec:results}
For all our tests, we employ an Intel Core i7 with two Geforce GTX 1080Ti graphics cards.
Our algorithm runs at around $25$~FPS, which fulfills the performance requirement of many real-time applications.
In all our experiments, we use the same set of parameters that are empirically determined: $\lambda_\textrm{2D}=460$, $\lambda_\textrm{3D}=28$, $\lambda_\textrm{silhouette}=200$, $\lambda_\textrm{temporal}=1.5$, $\lambda_\textrm{anatomic}=10^6$, $w_\mathrm{photo}=10000$, $w_\mathrm{silhouette}=600$, $w_\mathrm{smooth}=10.0$, $w_\mathrm{edge}=30.0$, $w_\mathrm{velocity}=0.25$ and $w_\mathrm{acceleration}=0.1$.
In the following, we first introduce our new dataset, evaluate our approach on several challenging sequences qualitatively and quantitatively, and compare to related methods.
Then, we perform an ablation evaluation to study the importance of the different components of our approach.
Finally, we demonstrate several live applications.
More results are shown in our two supplementary videos, which in total show over 20 minutes of performance capture results.
We applied smoothing with a filter of window size 3 (stencil: $[0.15,0.7,0.15]$) to the trajectories of the vertex coordinates as a post process for all video results except in the live setup.
\subsection{Dataset}\label{sec:dataset}
In order to qualitatively evaluate our method on a wide range of settings we recorded several challenging motion sequences.
These contain large variations in non-rigid clothing deformations, e.g., skirts and hooded sweaters, and fast motions like dancing and jumping jacks.
In total, we captured over 20 minutes of video footage split in 11 sequences with different sets of apparel each worn by one of seven subjects.
All sequences were recorded with a Blackmagic video camera (30fps, $540\times 960$ resolution).
We provide semantically segmented, rigged and textured templates, calibrated camera parameters and an empty background image for all sequences.
In addition, we provide the silhouettes from background subtraction, our motion estimates and the non-rigidly deformed meshes.
For eight of the sequences we captured the subject from a reference view, which we will also make available, to evaluate the tracking quality.
Fig.~\ref{fig:qualitative} shows some of the templates and example frames of the captured sequences.
All templates are shown in the supplementary video.
We will make the full dataset publicly available.
\subsection{Qualitative and Quantitative Results}
\begin{figure}[t]
	\includegraphics[width=\linewidth]{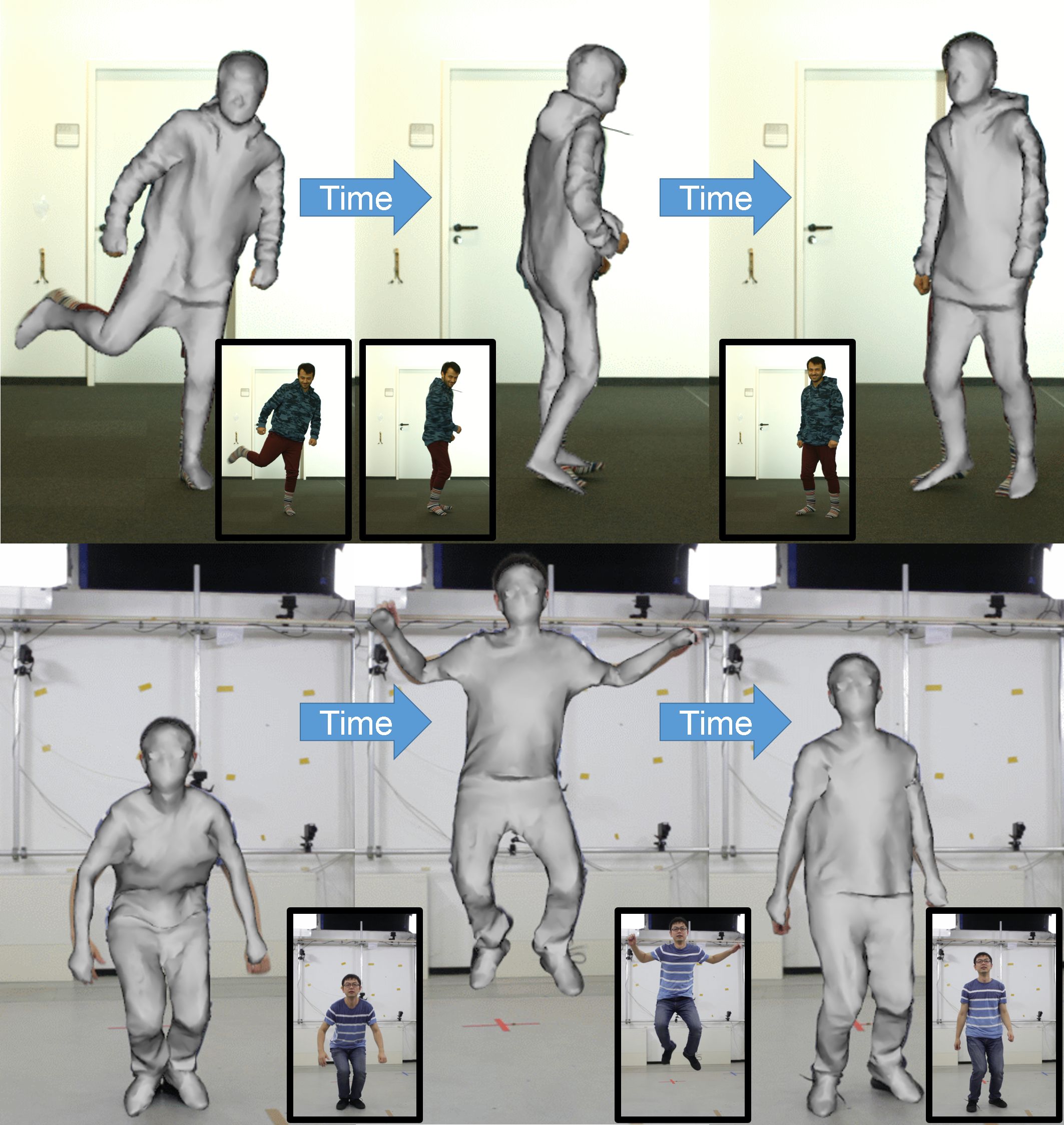}
	\caption
	{
		Our real-time approach even tracks challenging and fast motions, such as jumping and a fast $360^{\circ}$ rotation with high accuracy.
		The reconstructions overlay the input image well.
		For the complete sequence we refer to the supplemental video.
	}
	\label{fig:challenging_motion}
\end{figure}

\begin{figure}[t]
	\includegraphics[width=\linewidth]{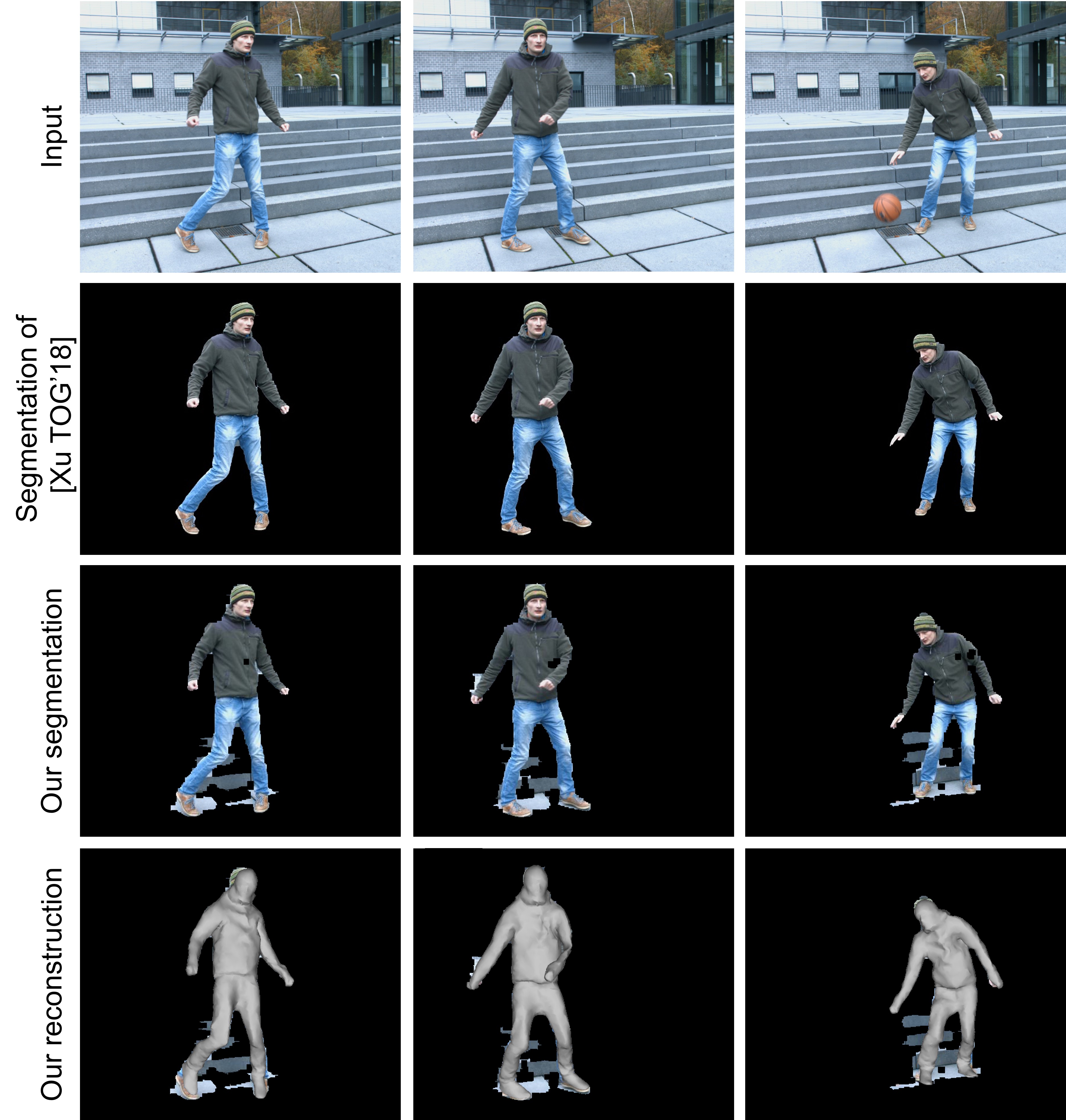}
	\caption
	{
		Comparison of the foreground segmentation of \citet{xu17MonoPerfCap} and our method.
		Note that our silhouette estimates are less accurate than the ones of \cite{xu17MonoPerfCap}.
		Nevertheless, our reconstruction results are robust to the noisy foreground estimates and look plausible.
	}
	\label{fig:silhouetteComparison}
\end{figure}

In total, we evaluated our approach on our new dataset and five existing video sequences of people in different sets of apparel.
In addition, we test our method with 4 subjects in a live setup (see Fig.~\ref{fig:teaser}) with a low cost webcam.
Our method takes frames at $540\times 960$ resolution as input.
To better evaluate our non-rigid surface registration method, we used challenging loose clothing in these sequences, including skirts, dresses, hooded sweatshirts and baggy pants.
The sequences show a wide range of difficult motions (slow to fast, self-occlusions) for monocular capture.
Additionally, we compare our approach to the state-of-the-art monocular performance capture method of~\citet{xu17MonoPerfCap} on two of their sequences and on one of our new captured sequences.
\paragraph{Qualitative Evaluation} 
In Fig.~\ref{fig:qualitative}, we show several frames from live performance capture results.
We can see that our results precisely overlay the person in the input images.
Note that body pose, head-orientation, and non-rigid deformation of loose clothing, are accurately captured.
Both the side-by-side comparison to RGB input and the accurate overlay with the reconstructed mesh show the high quality of the reconstruction.
Also note that our reconstruction results match the images captured from a laterally displaced reference view which is not used for tracking (see supplemental video).
This further evidences the fidelity of our 3D performance capture results, also in depth, which shows that our formulation effectively meets the non-trivial underconstrained monocular reconstruction challenge.
To evaluate the robustness of our method, we included many fast and challenging motions in our test set.
As shown in Fig.~\ref{fig:challenging_motion}, even the fast $360^{\circ}$ rotation (see the first row) and the jumping motion (see the second row) are successfully tracked.
This illustrates the robustness of our algorithm and its efficient and effective combined consideration of sparse and dense image cues, as well as learning-based and model-based capture, which in this combination were not used in prior work, let alone in real-time.
\begin{figure}[t]
	\includegraphics[width=\linewidth]{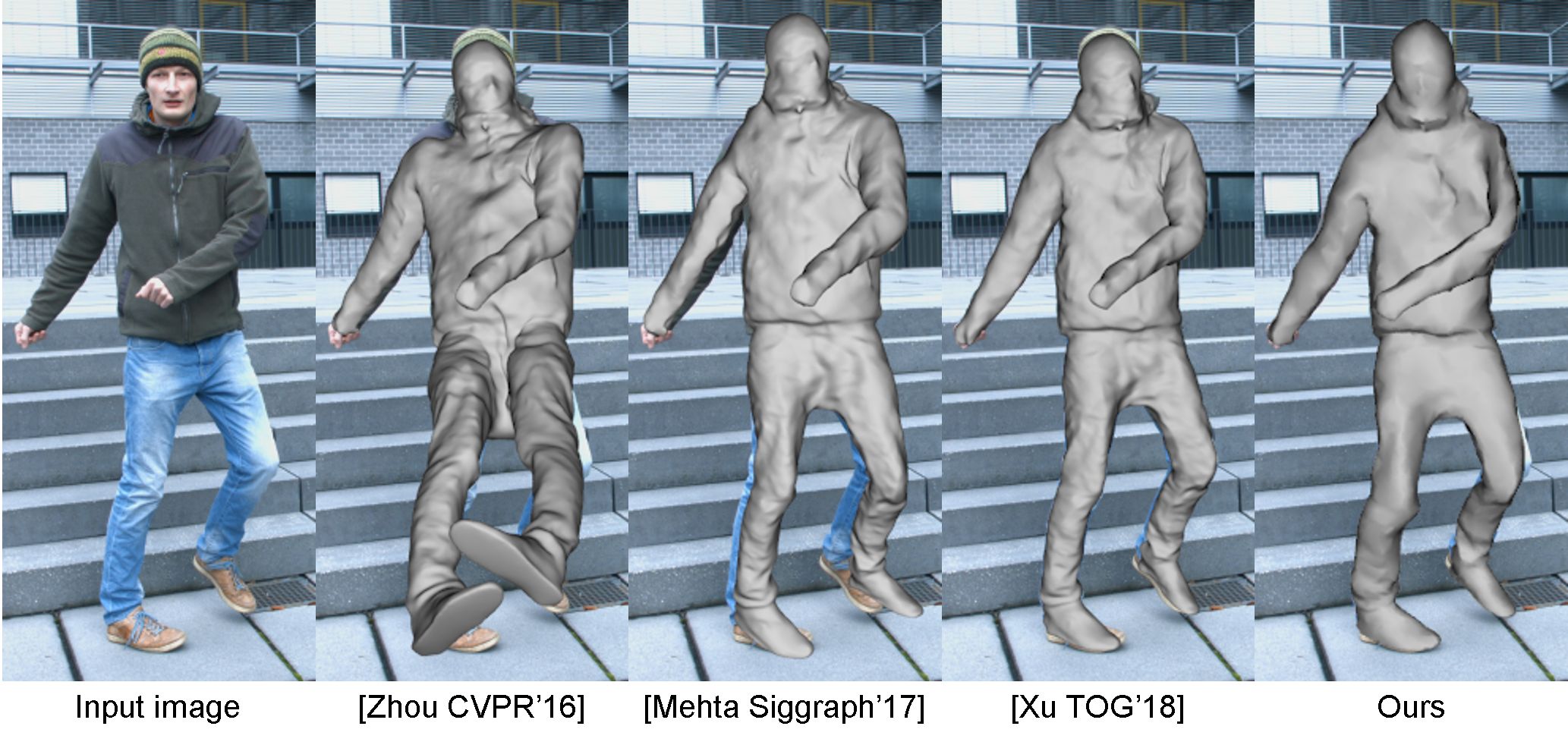}
	\caption
	{
		Qualitative comparison to related monocular methods.
		The results of our approach overlay much better with the input than the skeleton-only results of ~\citet{zhou2016sparseness} and ~\citet{VNect_SIGGRAPH2017}.
		Our results come close in quality to the off-line approach of ~\citet{xu17MonoPerfCap}.
	}
	\label{fig:compare_monocap}
\end{figure}
\begin{figure}[t]
	\includegraphics[width=\linewidth]{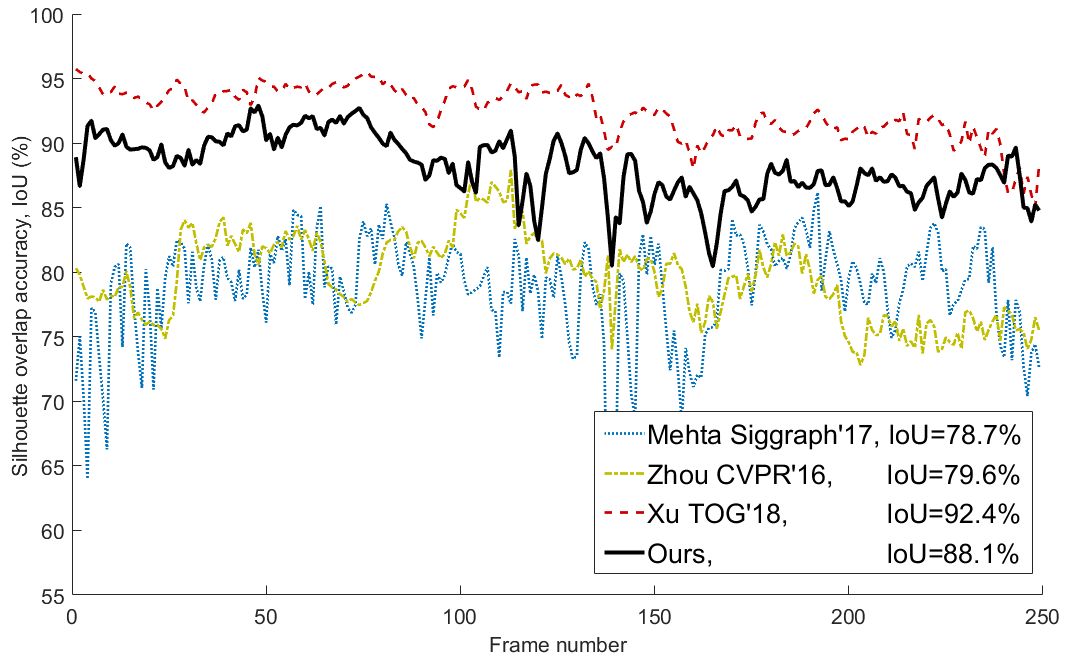}
	\caption
	{
		Quantitative comparison to related monocular methods.
		In terms of the silhouette overlap accuracy (Intersection over Union, IoU), our method achieves better results and outperforms ~\cite{zhou2016sparseness} and~\cite{VNect_SIGGRAPH2017} by $8.5\%$ and $9.4\%$, respectively.
		On average our results are only $4.3\%$ worse than the off-line approach of~\citet{xu17MonoPerfCap}, but our approach is orders of magnitude faster.
	}
	\label{fig:helge_iou}
\end{figure}
\paragraph{Comparison to Related Monocular Methods}
In Fig.~\ref{fig:compare_monocap}, we provide a comparison to 3 related state-of-the-art methods: 
The fundamentally off-line, monocular dense (surface-based) performance capture method of~\citet{xu17MonoPerfCap}, called MonoPerfCap, and two current monocular methods for 3D skeleton-only reconstruction, the 2D-to-3D lifting method of~\citet{zhou2016sparseness} and the real-time VNect algorithm~\cite{VNect_SIGGRAPH2017}.
For the latter two, we show the skinned rendering of our template using their skeleton pose.
The test sequence is provided by~\citet{xu17MonoPerfCap} with manually labeled ground truth silhouettes.
Our method's results overlay much better with the input than the skeleton-only results of ~\citet{zhou2016sparseness} and ~\citet{VNect_SIGGRAPH2017}, confirming our much better reconstructions.
Also a quantitative comparison on this sequence in terms of the silhouette overlap accuracy (Intersection over Union, IoU), Fig.~\ref{fig:helge_iou}, shows that our method achieves clearly better results and outperforms ~\cite{zhou2016sparseness} and~\cite{VNect_SIGGRAPH2017} by $8.5\%$ and $9.4\%$, respectively.
\begin{figure}[t]
	\includegraphics[width=\linewidth]{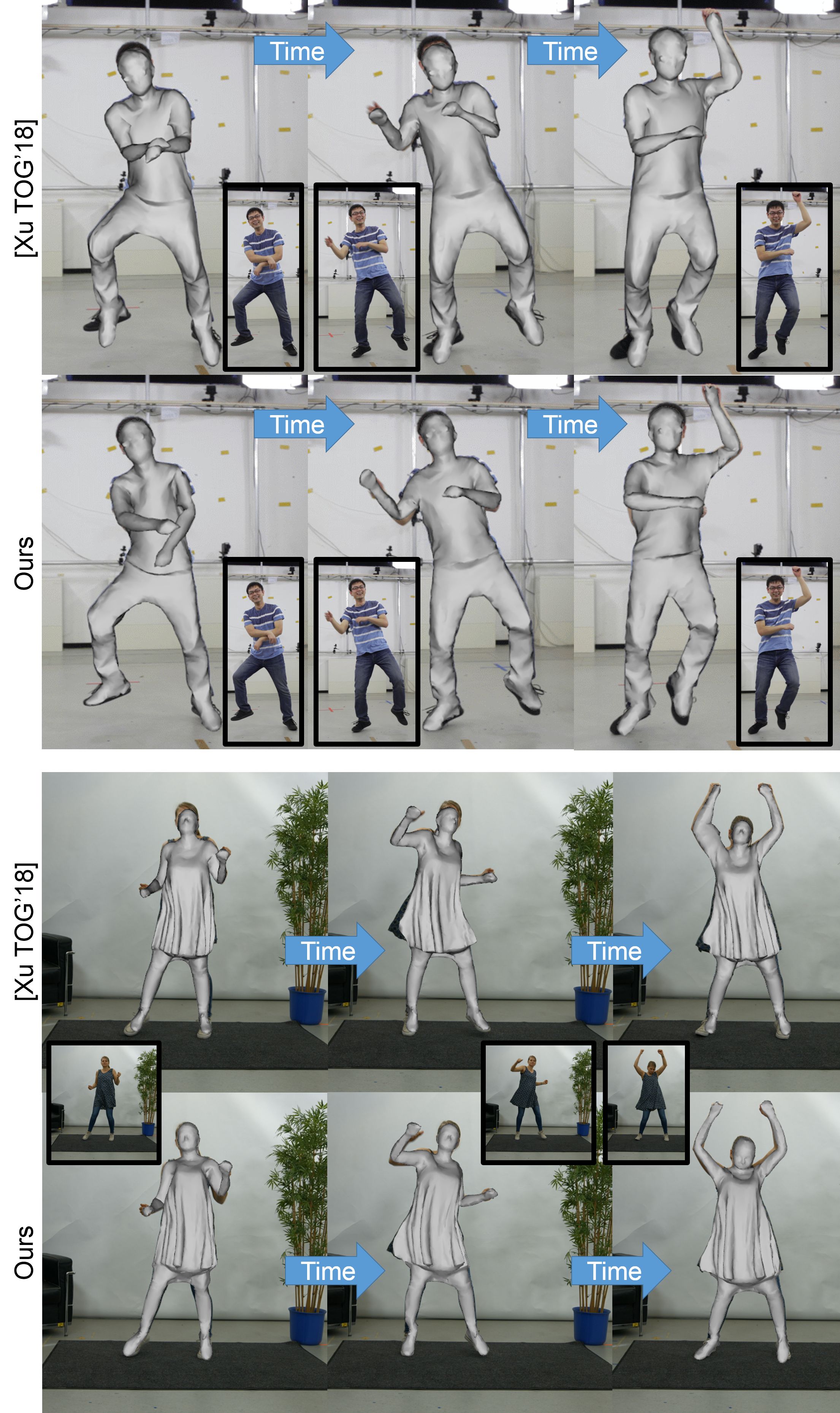}
	\caption
	{
		Qualitative comparison to MonoPerfCap \cite{xu17MonoPerfCap}.
		We achieve comparable reconstruction quality and overlay while being orders of magnitude faster.
	}
	\label{fig:compare_weipeng}
\end{figure}
Using the same metric, our IoU is only $4.3\%$ smaller than~\citet{xu17MonoPerfCap}, which is mainly caused by the fact that their foreground segmentation is more accurate than ours due to their more advanced but offline foreground segmentation strategy (see Fig.~\ref{fig:silhouetteComparison}).
But please note that our method is overall orders of magnitude faster than their algorithm which takes over $1$~minute per frame and our reconstructions are still robust to the noisy foreground segmentation.
To compare against MonoPerfCap more thoroughly, we also compare against them on one of our sequences (see Sec.~\ref{sec:dataset}), which shows more challenging non-rigid dress deformations in combination with fast motions (see bottom rows of Fig.~\ref{fig:compare_weipeng}).
On this sequence, the accuracy of the foreground estimation is roughly the same leading to the fact that our approach achieves an IoU of $86.86\%$ (averaged over 500 frames) which is almost identical to the one of~\cite{xu17MonoPerfCap} ($86.89\%$).
As shown in Fig.~\ref{fig:compare_weipeng}, we achieve comparable reconstruction quality and overlay while being orders of magnitude faster.
MonoPerfCap's window-based optimizer achieves slightly better boundary alignment and more stable tracking though some difficult, convolved, self-occluded poses, but is much slower.
Our reconstruction of head and feet is consistently better than~\cite{xu17MonoPerfCap} due to the additional facial landmark alignment term and the better pose detector that we adopted.
We provide a qualitative comparison showing highly challenging motions in the supplementary video.
\begin{figure}[t]
	\includegraphics[width=\linewidth]{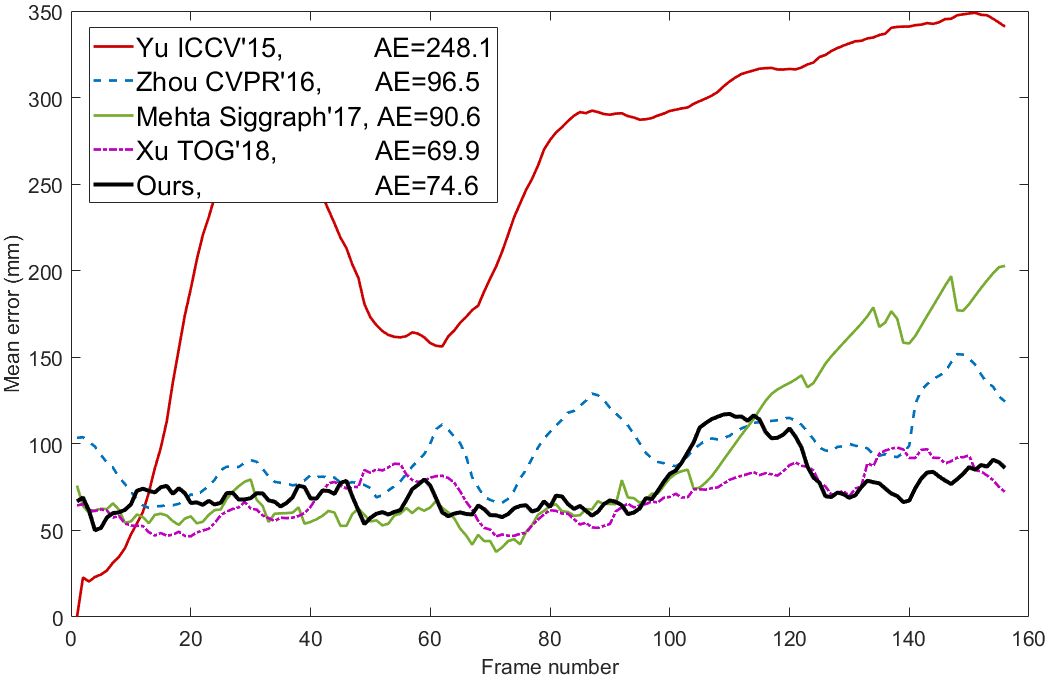}
	\caption
	{
		Quantitative comparison of the surface reconstruction accuracy on the \emph{Pablo} sequence.
		Our real-time monocular approach comes very close in quality to the results of the monocular \emph{off-line} method of~\citet{xu17MonoPerfCap}.
		It clearly outperforms the monocular non-rigid capture method of~\citet{Yu_2015_ICCV} and the rigged skeleton-only results of the 3D pose estimation methods of~\citet{zhou2016sparseness} and~\citet{VNect_SIGGRAPH2017}.
	}
	\label{fig:compare_pablo_plots}
\end{figure}
\begin{figure}[t]
	\includegraphics[width=\linewidth]{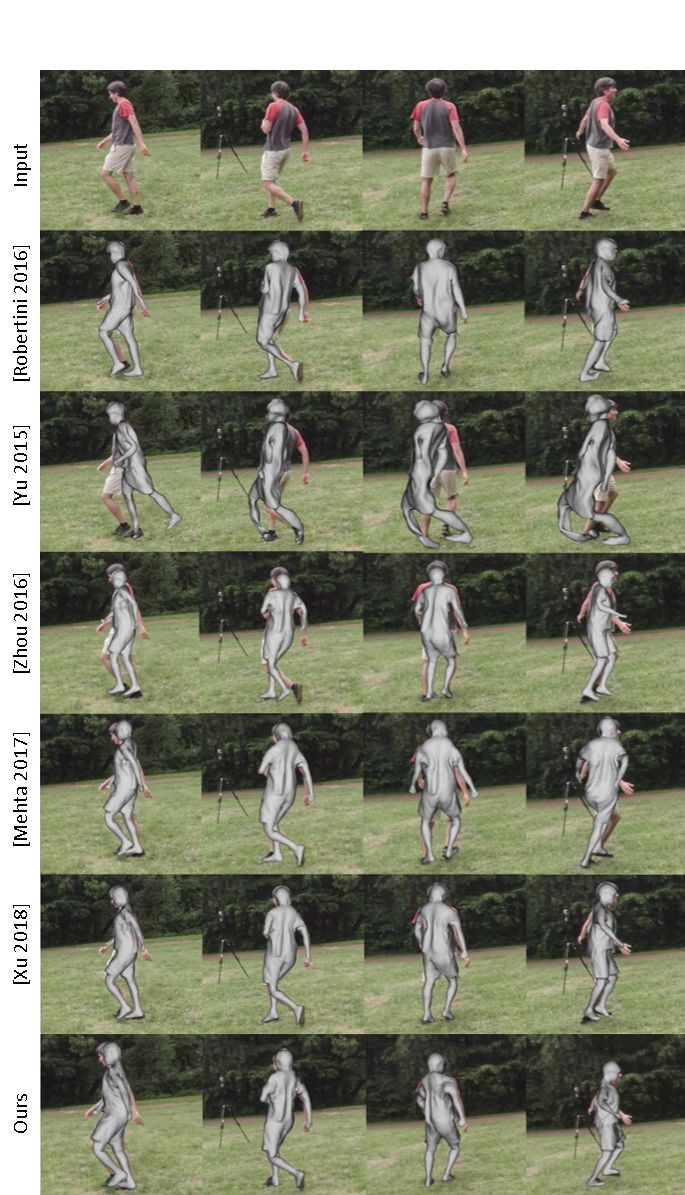}
	\caption
	{
		Qualitative comparisons of the surface reconstruction accuracy on the \emph{Pablo} sequence.
		Our real-time monocular approach comes very close in quality to the results of the fundamentally off-line \emph{multi-view} approach of~\citet{robertini2016model} and the monocular \emph{off-line} method of~\citet{xu17MonoPerfCap}.
		It clearly outperforms the monocular non-rigid capture method of~\citet{Yu_2015_ICCV} and the rigged skeleton-only results of the 3D pose estimation methods of~\citet{zhou2016sparseness} and~\citet{VNect_SIGGRAPH2017}.
	}
	\label{fig:compare_pablo}
\end{figure}
\paragraph{Surface Reconstruction Accuracy} 
To evaluate our surface reconstruction error, also relative to multi-view methods, we use the \emph{Pablo} sequence from the state-of-the-art multi-view template-based performance capture method of~\citet{robertini2016model} (they also provide the template).
As shown in Fig.~\ref{fig:compare_pablo}, our real-time monocular method comes very close in quality to the results of the fundamentally off-line \emph{multi-view} approach of~\citet{robertini2016model} and the monocular \emph{off-line} method of~\citet{xu17MonoPerfCap}.
In addition, it clearly outperforms the monocular non-rigid capture method of~\citet{Yu_2015_ICCV} and the rigged skeleton-only results of the 3D pose estimation methods of~\citet{zhou2016sparseness} and~\citet{VNect_SIGGRAPH2017} (latter two as described in the previous paragraph).
This is further evidenced by our quantitative evaluation on per-vertex position errors (see Fig.~\ref{fig:compare_pablo_plots}).
We use the reconstruction results of~\citet{robertini2016model} as reference and show the per-vertex Euclidean surface error.
Similar to~\cite{xu17MonoPerfCap}, we aligned the reconstruction of all methods to the reference meshes with a translation to eliminate the global depth offset.
The method of \citet{xu17MonoPerfCap} achieves slightly better results in terms of surface reconstruction accuracy.
Similar to our previous experiment (see Fig.~\ref{fig:silhouetteComparison}), we observed that our foreground estimates are slightly worse than the ones of \cite{xu17MonoPerfCap} which caused the lower accuracy.
\begin{figure}[t]
	\includegraphics[width=\linewidth]{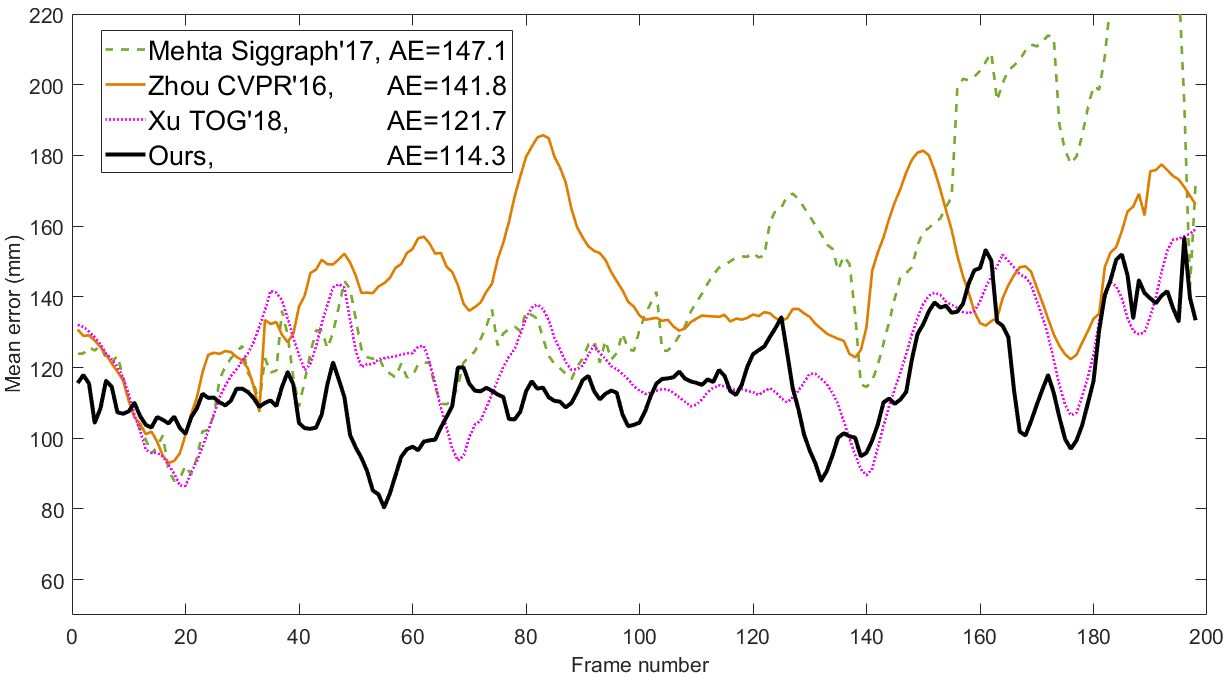}
	\caption
	{
		Comparison of the skeletal pose estimation accuracy in terms of average per-joint 3D error on the \emph{Pablo} sequence.
		Our method outperforms the three other methods, most notably the skeleton-only methods of~\citet{VNect_SIGGRAPH2017} and \citet{zhou2016sparseness}.
	}
	\label{fig:pose_plots}
\end{figure}
\paragraph{Skeletal Pose Estimation Accuracy} 
We also compare our approach in terms of joint position accuracy on the \emph{Pablo} sequence against VNect~\cite{VNect_SIGGRAPH2017},~\cite{zhou2016sparseness} and MonoPerfCap~\cite{xu17MonoPerfCap}.
As reference, we use the joint positions from the multi-view method of~\citet{robertini2016model}.
We report the average per-joint 3D error (in millimeters) after aligning the per-frame poses with a similarity transform.
As shown in Fig.~\ref{fig:pose_plots}, our method outperforms the three other methods, most notably the skeleton-only methods~\cite{VNect_SIGGRAPH2017,zhou2016sparseness}.
This shows that our combined surface and skeleton reconstruction also benefits 3D pose estimation quality in itself.
\begin{figure}[t]
	\includegraphics[width=\linewidth]{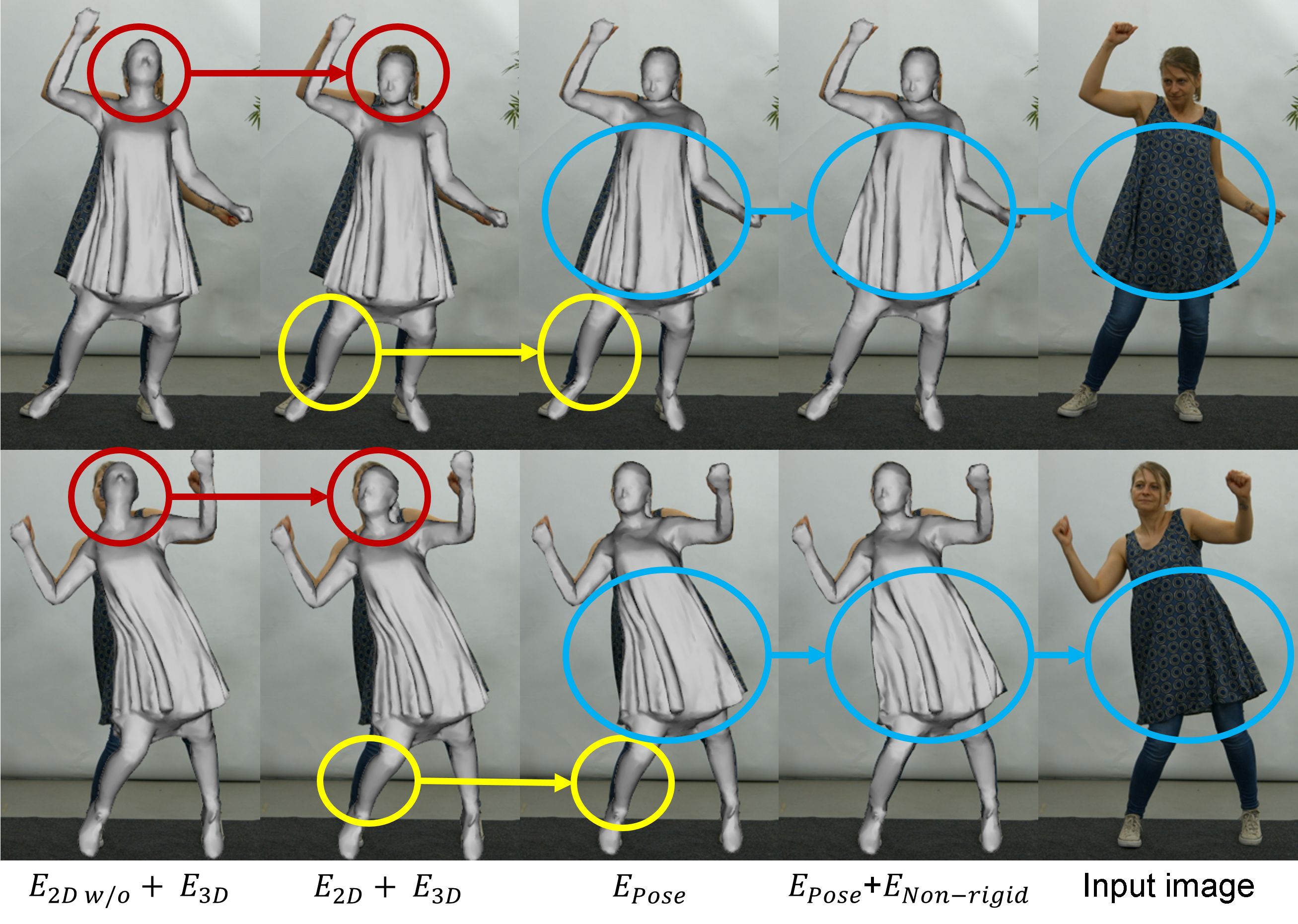}
	\caption
	{
		Ablation study.
		1) the facial landmark alignment term significantly improves the head orientation estimation (red circles), 
		2) the misalignment of $E_{2D}+E_{3D}$ is corrected by our silhouette term in $E_{pose}$ (yellow circles), 
		3) the non-rigid deformation on the surface, which cannot be modeled by skinning, is accurately captured by our non-rigid registration method $E_{non-rigid}$ (blue circles).
	}
	\label{fig:ablation}
\end{figure}
\begin{figure}[t]
	\includegraphics[width=\linewidth]{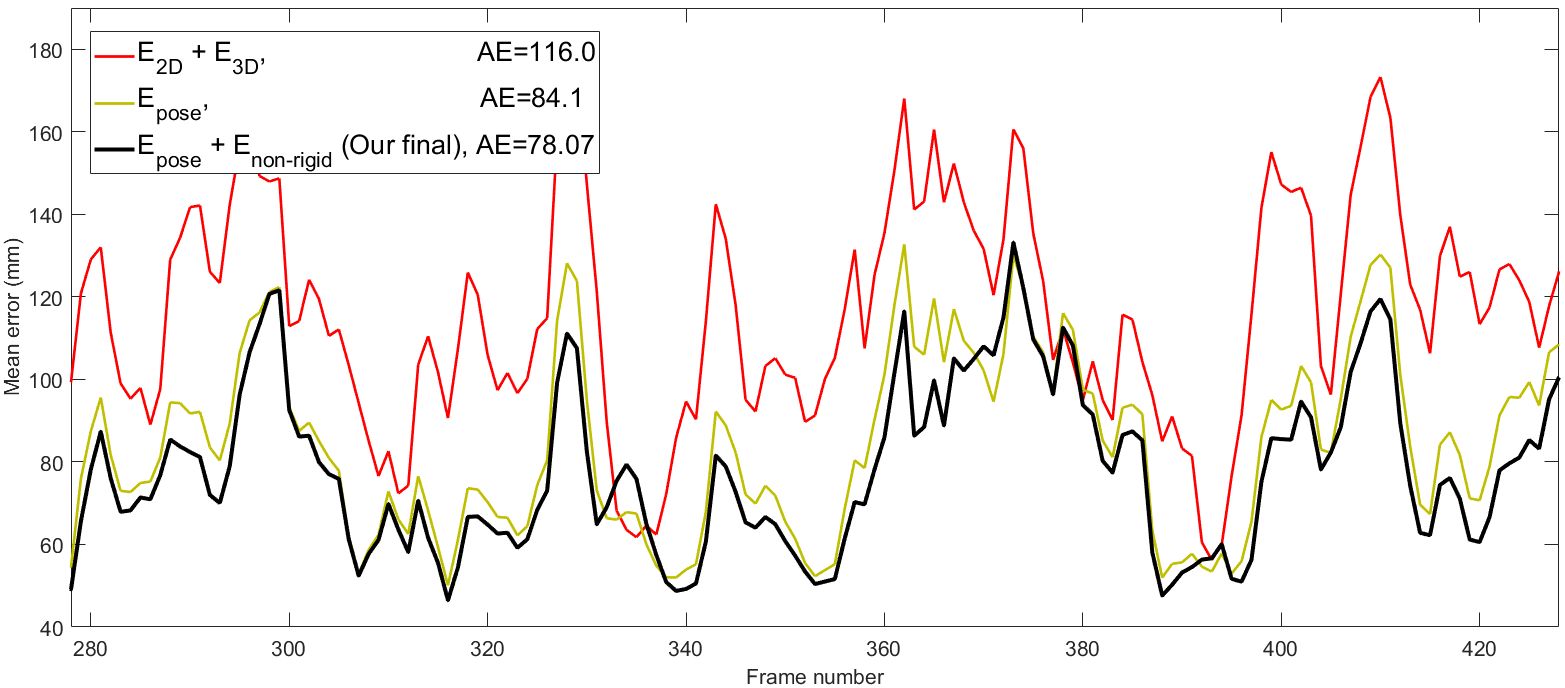}
	\caption
	{
		Ablation study.
		The mean vertex position error clearly demonstrates the consistent improvement by each of the algorithmic components of our approach.
		Our full approach consistently obtains the lowest error.
	}
	\label{fig:ablation_plots}
\end{figure}
\begin{figure}[t]
	\includegraphics[width=\linewidth]{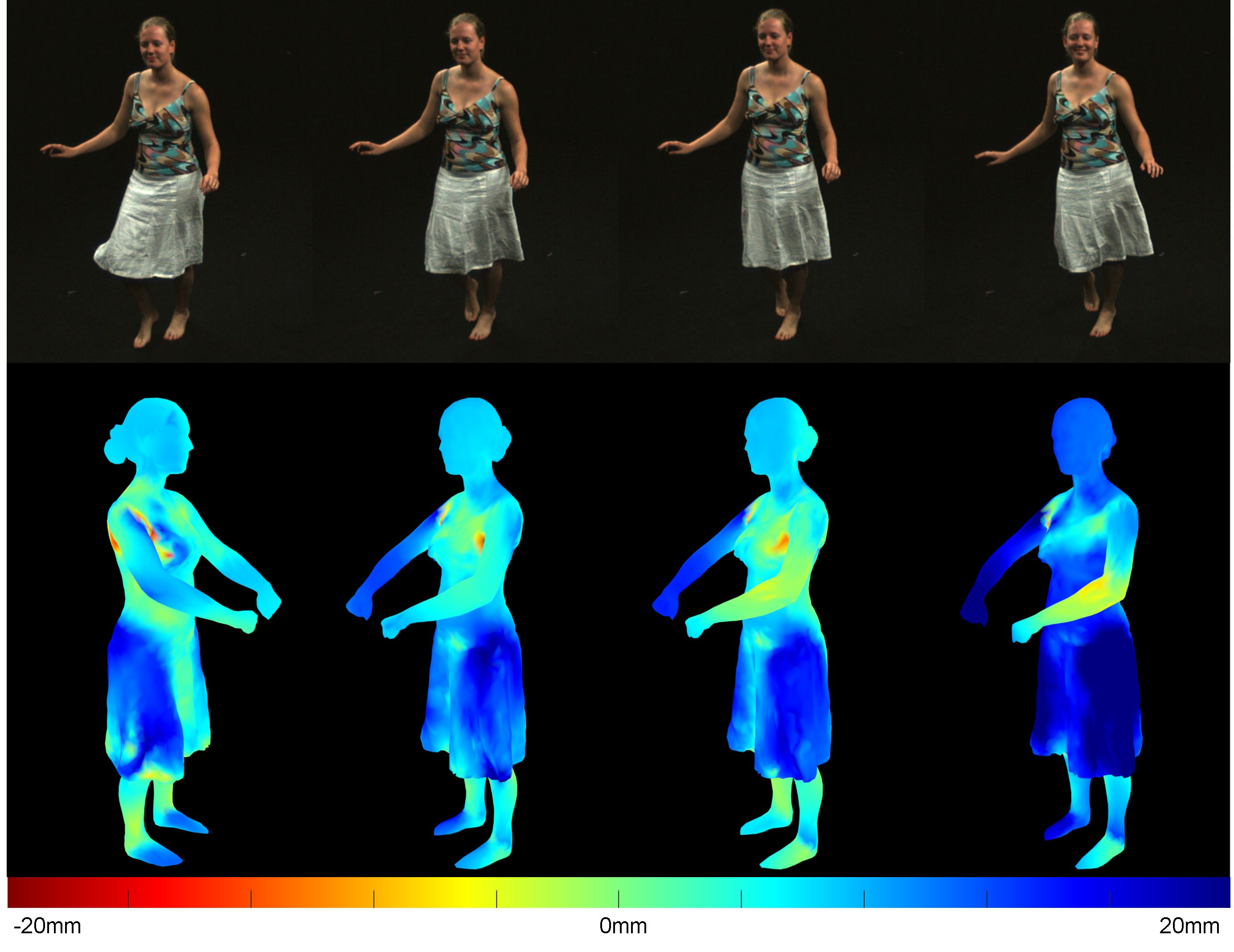}
	\caption
	{
		Improvement of the non-rigid stage ($E_{pose} + E_{non-rigid}$) over pose-only deformation ($E_{pose}$).
		Top row: Four monocular input images.
		On the bottom row, for each image, we show the per-vertex error of the pose only results minus the per-vertex error of our method.
		Consequently, negative means pose only is better and is colored in red.
		Positive means our method is better and is colored in blue.
		As expected our method achieves most improvement on the non-rigid skirt part--- which is around 20mm for the shown frames.
	}
	\label{fig:improvement}
\end{figure}
\paragraph{Ablation Study} 
We first qualitatively evaluate the importance of all algorithmic components in an ablation study on a real video sequence.
To this end, we compare the results of: 1) our pose estimation without facial landmark alignment term and the silhouette term, which we refer to as $E_{2D w/o face}+E_{3D}$, 2) our pose estimation without the silhouette term ( $E_{2D}+E_{3D}$), 3) our complete pose estimation ($E_{pose}$) and 4) our full pipeline ($E_{pose}+E_{non-rigid}$).
As shown in Fig.~\ref{fig:ablation}, 1) the facial landmark alignment term significantly improves the head orientation estimation (red circles), 2) the misalignment of $E_{2D}+E_{3D}$ is corrected by our silhouette term in $E_{pose}$ (yellow circles), 3) the non-rigid deformation on the surface, which cannot be modeled by skinning, is accurately captured by our non-rigid registration method $E_{non-rigid}$ (blue circles).
Second, we also quantitatively evaluated the importance of the terms on a sequence where high-quality reconstructions based on the multi-view performance capture results of \citet{de2008performance} are used as ground truth.
The mean vertex position error shown in Fig.~\ref{fig:ablation_plots} clearly demonstrates the consistent improvement by each of the algorithmic components of our approach.
The non-rigid alignment stage obtains on average better results than the pose-only alignment.
Since non-rigid deformations are most of the time concentrated in certain areas, e.g., a skirt, and at certain frames when articulated motion takes place, we also measure the per-frame and per-vertex improvement of the proposed non-rigid stage.
To this end, we measure the improvement of ($E_{pose} + E_{non-rigid}$) over ($E_{pose}$) by computing the per-vertex error of the pose only results minus the per-vertex error of our method.
Consequently, positive means our method is better than the pose-only deformation.
As demonstrated in Fig.~\ref{fig:improvement}, the non-rigid stage significantly improves the reconstruction of the skirt and the arm.
The improvement is especially noticeable for frames where the deformation of the skirt significantly differs from the static template model, since such motion cannot be handled by the pose-only step.
On the same dataset we also evaluated the influence of 1) the warping of the non-rigid displacement of the previous frame, 2) the proposed body part masks used in the dense silhouette alignment, and 3) the proposed vertex snapping.
Those algorithmic changes respectively lead to 2.4\%, 1.7\% and 1.7\% improvement in average 3D vertex error which sums up to a total improvement of 5.8\%.
\begin{figure}[t]
	\includegraphics[width=\linewidth]{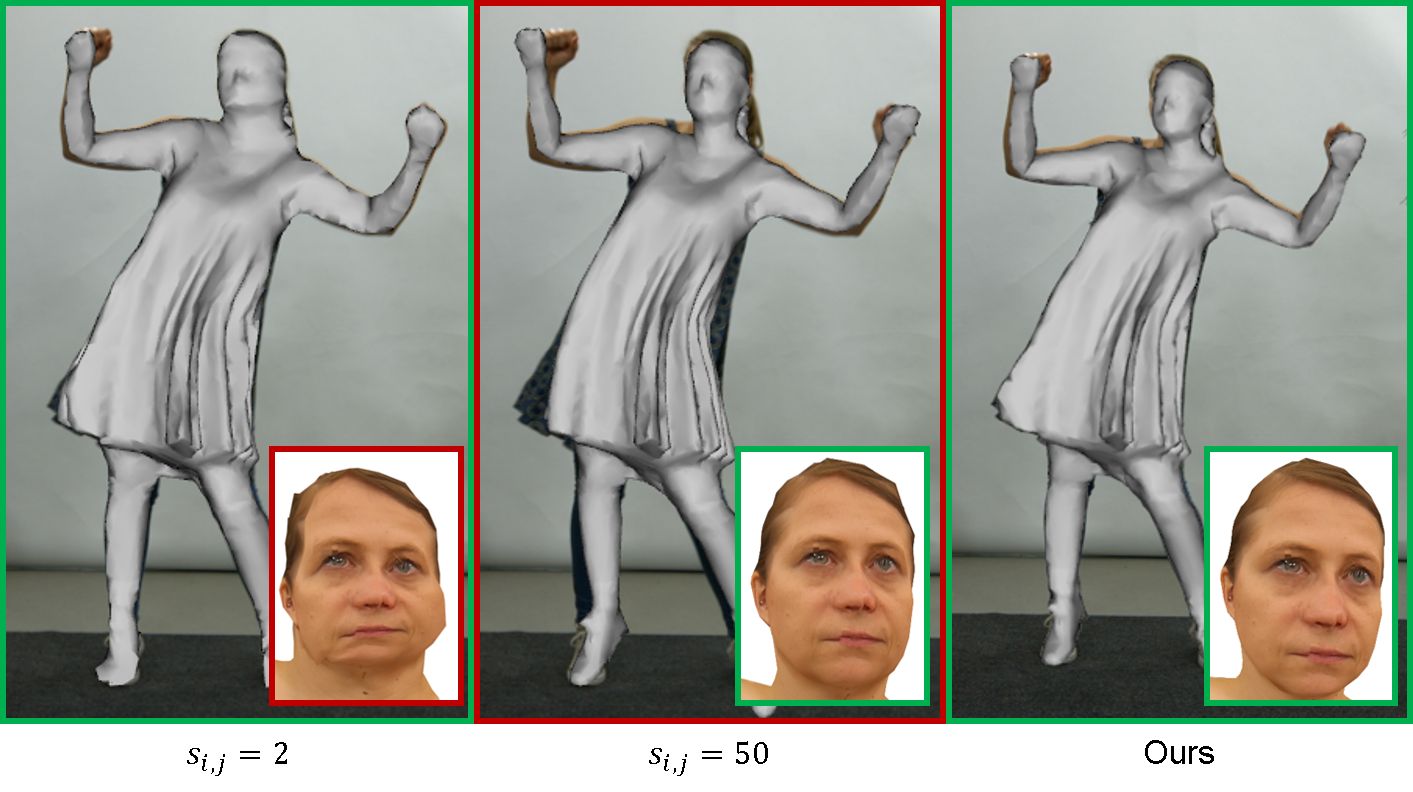}
	\caption
	{
		Importance of our material-based non-rigid deformation adaptation strategy.
		With a low global regularization weight the deformation of the skirt is well reconstructed, but the head is distorted (left).
		A high deformation weight preserves the shape of the head, but prevents tracking of the skirt motion (middle).
		Our new semantic weight adaptation strategy enables the reconstruction of both regions with high accuracy and leads to the best results (right).
	}
	\label{fig:ablation_semantic}
\end{figure}
The importance of our material-based non-rigid deformation adaptation strategy is shown in Fig.~\ref{fig:ablation_semantic}.
With constantly low non-rigidity weights ($s_{i,j}=2.0$) in all regions, the deformation of the skirt is well reconstructed, but the head is severely distorted (left).
In contrast, with high global non-rigidity weights ($s_{i,j}=50.0$), the head shape is preserved, but the skirt cannot be tracked reliably (middle).
Our new semantic weight adaptation strategy enables the reconstruction of both regions with high accuracy and leads to the best results (right).
\subsection{Applications}
\begin{figure}[t]
	\includegraphics[width=\linewidth]{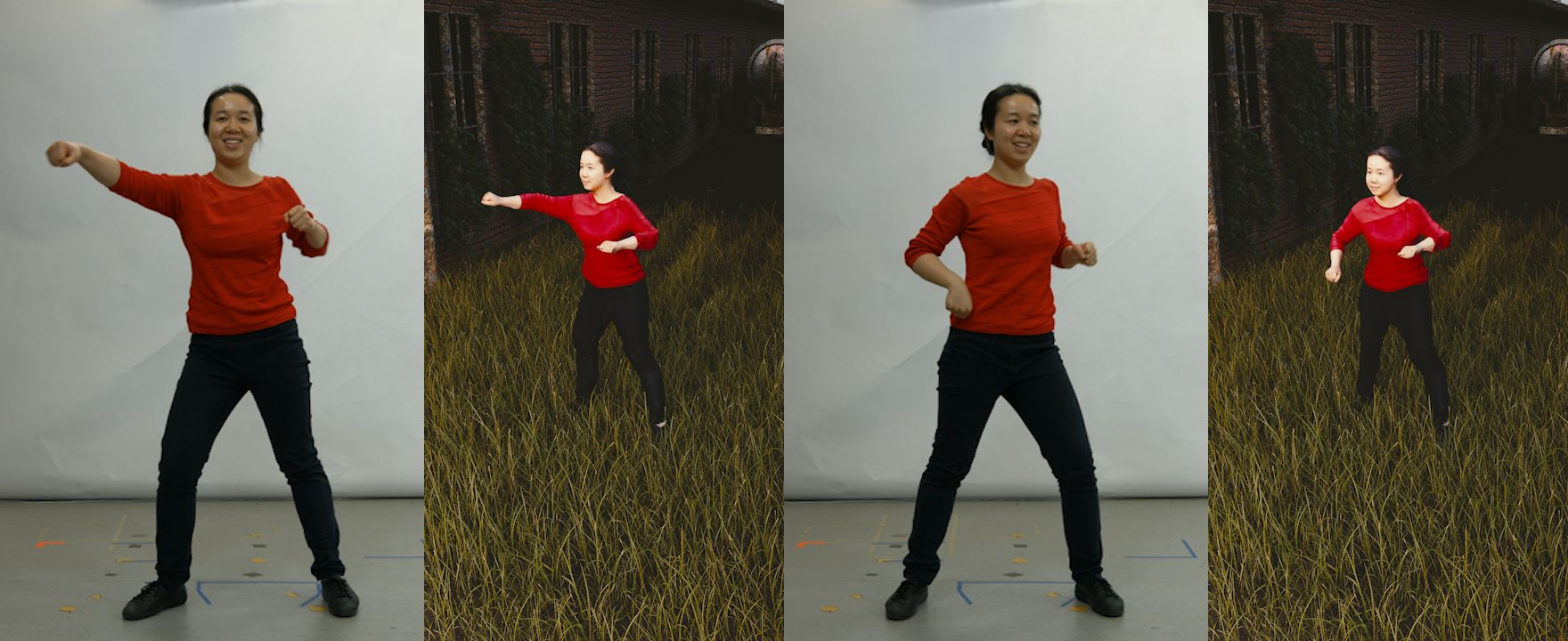}
	\caption{Free-viewpoint video rendering results using our approach.}
	\label{fig:freeview_ikhs}
\end{figure}

\begin{figure}[t]
	\includegraphics[width=\linewidth]{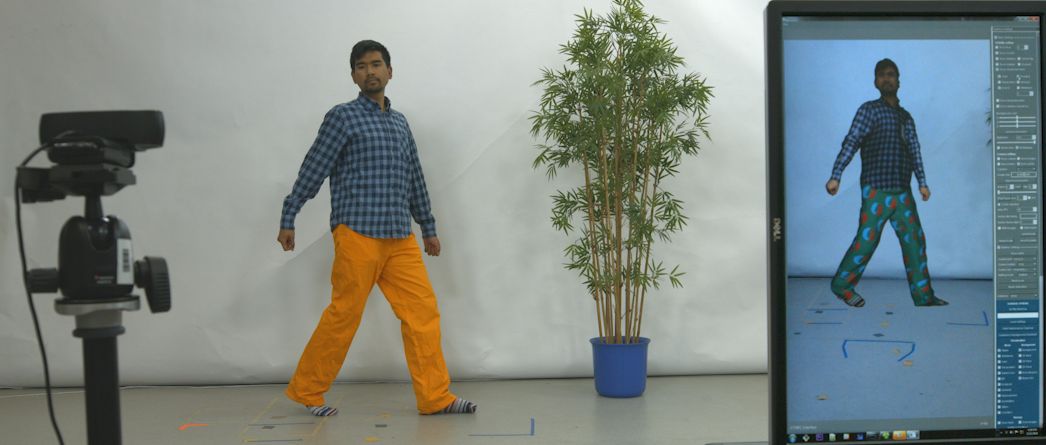}
	\caption{Live virtual try-on application based on our approach.}
	\label{fig:tryon}
\end{figure}

Our monocular real-time human performance capture method can facilitate many applications that depend on real-time capture: interactive VR and AR, human-computer interaction, pre-visualization for visual effects, 3D video or telepresence.
We exemplify this through two application demonstrators.
In Fig.~\ref{fig:freeview_ikhs}, we show that our method allows 
live free-viewpoint video rendering and computer animation of the performance captured result from just single color input.
This illustrates the potential of our method in several of the aforementioned live application domains.
In Fig.~\ref{fig:tryon}, we demonstrate a real-time virtual try-on application based on our performance capture method.
We replace the texture corresponding to the trousers on the template and visualize the tracked result in real time.
With such a system, the users can see themselves in clothing variants in real time with live feedback, which could be potentially used in VR or even AR online shopping.
\section{Discussion and Limitations} \label{sec:limitations}
We have demonstrated compelling real-time full-body human performance capture results using a single consumer-grade color camera.
Our formulation combines constraints used, individually, in different image-based reconstruction methods before.
But the specific combination we employ embedded in a hierarchical real-time approach is new and enables, for the first time, real-time monocular performance capture.
Further, our formulation geared rigorously for real-time use differs from the related, but off-line MonoPerfCap \cite{xu17MonoPerfCap} method in several ways:
In Stage I, the facial landmarks as well as the displacement warping which is also added during pose tracking improve the pose accuracy of our real-time method.
Further, we track the pose per frame instead of a batch-based formulation which reduces the computation time and allows faster motions.
Further improvement in terms of efficiency are achieved by our GPU-based pose solver.
In Stage II, our dense photometric term that adds constraints for non-boundary vertices and our adaptive material based regularization improve reconstruction quality.
Our non-rigid fitting stage is faster due to the more efficient combination of spatial regularizers that requires a much smaller number of variables than the as-rigid-as-possible regularizer.
We directly solve for the vertex displacements instead of estimating the embedded graph rotations/translations.
We found that this formulation is better suited for a parallel implementation on the GPU and it also gives a more flexible representation.
Due to our real-time constraint, we make use of an efficient distance transform-based representation, instead of the ICP-based approach that requires expensive search of correspondences between the model boundary and the image silhouettes.
Our experiments show that our method achieves a similar reconstruction quality compared to the off-line performance capture approach of \citet{xu17MonoPerfCap} while being orders of magnitude faster.
\par
\begin{figure}[t]
	\includegraphics[width=\linewidth]{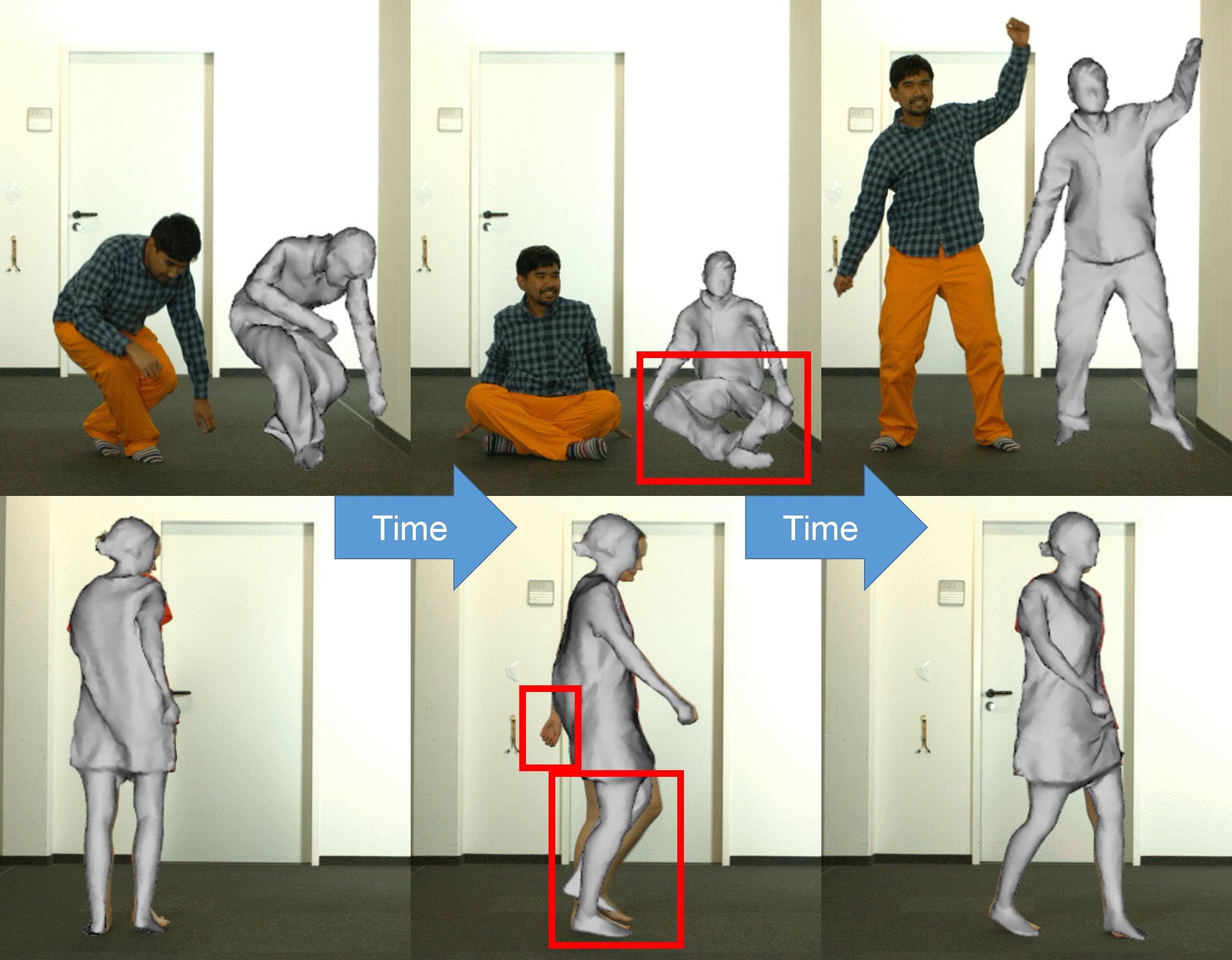}
	\caption
	{
		Failure cases.
		Top row: The underlying 3D joint regression deep network can fail for extreme poses not seen in training, which can produce glitches in the tracking results.
		Our model fitting can often but not always correct such wrong estimates.
		However, our performance capture approach robustly recovers from such situations.
		Bottom row: Our estimates for occluded parts will be less accurate than with multi-view methods due to the lack of image evidence.
		While pose and silhouette plausibly constrain the back-side of the body, fully-occluded limbs may have incorrect poses.
	}
	\label{fig:limitation}
\end{figure}

Nonetheless, our approach is subject to some limitations, see also Fig.~\ref{fig:limitation}.
Due to the ambiguities that come along with monocular performance capture, we rely on an accurate template acquisition since reconstruction errors and mislabeled part segmentations in the template itself cannot be recovered during tracking.
Further, we cannot handle topological changes that are too far from the template, e.g. removing of some clothes and deformations along the camera viewing axis can only be partially recovered by our photometric term.
The latter point could be addressed by an additional term that involves shading and illumination estimation.
As is common for learning methods, the underlying 3D joint regression deep network fails for extreme poses not seen in training.
Our model fitting can often, but not always correct such wrong estimates which produces glitches in the tracking results.
However, our performance capture approach robustly recovers from such situations, see Fig.~\ref{fig:limitation} (top).
Since our method uses foreground/background segmentation, strong shadows and shading effects, objects with similar color to the performer, and changing illumination situations can cause suboptimal segmentation; thus leading to noisy data association in the silhouette alignment term, which manifests itself as high-frequency jitter.
Our approach is robust to some degree of miss-classifications, but can get confused by big segmentation outliers.
This could be alleviated in the future by incorporating more sophisticated background segmentation strategies, e.g., based on deep neural networks.
Strong changes in shading or shadows, specular materials or non-diffuse lighting can also negatively impact the color alignment term.
A joint optimization for scene illumination and material properties could alleviate this problem.
Even though we carefully orchestrated the components of our method to achieve high accuracy and temporal stability in this challenging monocular setting, even under non-trivial occlusions, extensive (self-)occlusion are still fundamentally difficult.
Our estimates for occluded parts will be less accurate than with multi-view methods due to the lack of image evidence.
While pose and silhouette plausibly constrain the back-side of the body, fully-occluded limbs may have incorrect poses.
Additional learned motion priors could further resolve such ambiguous situations.
Fortunately, our approach recovers as soon as the difficult occlusions are gone, see Fig.~\ref{fig:limitation} (bottom).
\section{Conclusion} \label{sec:Conclusion}
We have presented the first monocular real-time human performance capture approach that reconstructs dense, space-time coherent deforming geometry of entire humans in their loose everyday clothing.
Our novel energy formulation leverages automatically identified material regions on the template to differentiate between different non-rigid deformation behaviors of skin and various types of apparel.
We tackle the underlying non-linear optimization problems at real-time based on a pipelined implementation that runs two specially-tailored data-parallel Gauss-Newton solvers, one for pose estimation and one for non-rigid tracking, at the same time.
We deem our approach as a first step towards general real-time capture of humans from just a single view, which is an invaluable tool for believable, immersive virtual and augmented reality, telepresence, virtual try-on, and many more exciting applications the future will bring to our homes.
An interesting direction for future work is the joint estimation of human motion, facial expression, hand pose, and hair dynamics from a single monocular camera.

\bibliographystyle{ACM-Reference-Format}
\bibliography{submission} 


\begin{thebibliography}{00}


\ifx \showCODEN    \undefined \def \showCODEN     #1{\unskip}     \fi
\ifx \showDOI      \undefined \def \showDOI       #1{#1}\fi
\ifx \showISBNx    \undefined \def \showISBNx     #1{\unskip}     \fi
\ifx \showISBNxiii \undefined \def \showISBNxiii  #1{\unskip}     \fi
\ifx \showISSN     \undefined \def \showISSN      #1{\unskip}     \fi
\ifx \showLCCN     \undefined \def \showLCCN      #1{\unskip}     \fi
\ifx \shownote     \undefined \def \shownote      #1{#1}          \fi
\ifx \showarticletitle \undefined \def \showarticletitle #1{#1}   \fi
\ifx \showURL      \undefined \def \showURL       {\relax}        \fi
\providecommand\bibfield[2]{#2}
\providecommand\bibinfo[2]{#2}
\providecommand\natexlab[1]{#1}
\providecommand\showeprint[2][]{arXiv:#2}

\bibitem[\protect\citeauthoryear{Allain, Franco, and Boyer}{Allain
  et~al\mbox{.}}{2015}]%
        {InriaVolumetric_2015}
\bibfield{author}{\bibinfo{person}{Benjamin Allain},
  \bibinfo{person}{Jean-S{\'e}bastien Franco}, {and} \bibinfo{person}{Edmond
  Boyer}.} \bibinfo{year}{2015}\natexlab{}.
\newblock \showarticletitle{{An Efficient Volumetric Framework for Shape
  Tracking}}. In \bibinfo{booktitle}{{\em {CVPR 2015 - IEEE International
  Conference on Computer Vision and Pattern Recognition}}}.
  \bibinfo{publisher}{{IEEE}}, \bibinfo{address}{Boston, United States},
  \bibinfo{pages}{268--276}.
\newblock
\showDOI{%
\url{https://doi.org/10.1109/CVPR.2015.7298623}}


\bibitem[\protect\citeauthoryear{Anguelov, Srinivasan, Koller, Thrun, Rodgers,
  and Davis}{Anguelov et~al\mbox{.}}{2005}]%
        {anguelov2005scape}
\bibfield{author}{\bibinfo{person}{Dragomir Anguelov}, \bibinfo{person}{Praveen
  Srinivasan}, \bibinfo{person}{Daphne Koller}, \bibinfo{person}{Sebastian
  Thrun}, \bibinfo{person}{Jim Rodgers}, {and} \bibinfo{person}{James Davis}.}
  \bibinfo{year}{2005}\natexlab{}.
\newblock \showarticletitle{{SCAPE: Shape Completion and Animation of People}}.
\newblock \bibinfo{journal}{{\em ACM Transactions on Graphics\/}}
  \bibinfo{volume}{24}, \bibinfo{number}{3} (\bibinfo{year}{2005}),
  \bibinfo{pages}{408--416}.
\newblock


\bibitem[\protect\citeauthoryear{B{\u{a}}lan and Black}{B{\u{a}}lan and
  Black}{2008}]%
        {bualan2008naked}
\bibfield{author}{\bibinfo{person}{Alexandru~O B{\u{a}}lan} {and}
  \bibinfo{person}{Michael~J Black}.} \bibinfo{year}{2008}\natexlab{}.
\newblock \showarticletitle{The naked truth: Estimating body shape under
  clothing}. In \bibinfo{booktitle}{{\em European Conference on Computer
  Vision}}. Springer, \bibinfo{pages}{15--29}.
\newblock


\bibitem[\protect\citeauthoryear{Balan, Sigal, Black, Davis, and
  Haussecker}{Balan et~al\mbox{.}}{2007}]%
        {balan2007detailed}
\bibfield{author}{\bibinfo{person}{Alexandru~O Balan}, \bibinfo{person}{Leonid
  Sigal}, \bibinfo{person}{Michael~J Black}, \bibinfo{person}{James~E Davis},
  {and} \bibinfo{person}{Horst~W Haussecker}.} \bibinfo{year}{2007}\natexlab{}.
\newblock \showarticletitle{Detailed human shape and pose from images}. In
  \bibinfo{booktitle}{{\em IEEE Conference on Computer Vision and Pattern
  Recognition (CVPR)}}. \bibinfo{pages}{1--8}.
\newblock


\bibitem[\protect\citeauthoryear{Bartoli, Gérard, Chadebecq, Collins, and
  Pizarro}{Bartoli et~al\mbox{.}}{2015}]%
        {Bartoli2015}
\bibfield{author}{\bibinfo{person}{A. Bartoli}, \bibinfo{person}{Y. Gérard},
  \bibinfo{person}{F. Chadebecq}, \bibinfo{person}{T. Collins}, {and}
  \bibinfo{person}{D. Pizarro}.} \bibinfo{year}{2015}\natexlab{}.
\newblock \showarticletitle{Shape-from-Template}.
\newblock \bibinfo{journal}{{\em IEEE Transactions on Pattern Analysis and
  Machine Intelligence\/}} \bibinfo{volume}{37}, \bibinfo{number}{10}
  (\bibinfo{date}{Oct} \bibinfo{year}{2015}), \bibinfo{pages}{2099--2118}.
\newblock
\showISSN{0162-8828}
\showDOI{%
\url{https://doi.org/10.1109/TPAMI.2015.2392759}}


\bibitem[\protect\citeauthoryear{Bogo, Black, Loper, and Romero}{Bogo
  et~al\mbox{.}}{2015}]%
        {Bogo:ICCV:2015}
\bibfield{author}{\bibinfo{person}{Federica Bogo}, \bibinfo{person}{Michael~J.
  Black}, \bibinfo{person}{Matthew Loper}, {and} \bibinfo{person}{Javier
  Romero}.} \bibinfo{year}{2015}\natexlab{}.
\newblock \showarticletitle{Detailed Full-Body Reconstructions of Moving People
  from Monocular {RGB-D} Sequences}. In \bibinfo{booktitle}{{\em International
  Conference on Computer Vision (ICCV)}}. \bibinfo{pages}{2300--2308}.
\newblock


\bibitem[\protect\citeauthoryear{Bogo, Kanazawa, Lassner, Gehler, Romero, and
  Black}{Bogo et~al\mbox{.}}{2016}]%
        {bogo2016smpl}
\bibfield{author}{\bibinfo{person}{Federica Bogo}, \bibinfo{person}{Angjoo
  Kanazawa}, \bibinfo{person}{Christoph Lassner}, \bibinfo{person}{Peter
  Gehler}, \bibinfo{person}{Javier Romero}, {and} \bibinfo{person}{Michael~J.
  Black}.} \bibinfo{year}{2016}\natexlab{}.
\newblock \showarticletitle{Keep it {SMPL}: Automatic Estimation of {3D} Human
  Pose and Shape from a Single Image}. In \bibinfo{booktitle}{{\em European
  Conference on Computer Vision (ECCV)}}.
\newblock


\bibitem[\protect\citeauthoryear{Bray, Kohli, and Torr}{Bray
  et~al\mbox{.}}{2006}]%
        {bray2006posecut}
\bibfield{author}{\bibinfo{person}{Matthieu Bray}, \bibinfo{person}{Pushmeet
  Kohli}, {and} \bibinfo{person}{Philip~HS Torr}.}
  \bibinfo{year}{2006}\natexlab{}.
\newblock \showarticletitle{Posecut: Simultaneous segmentation and 3d pose
  estimation of humans using dynamic graph-cuts}. In \bibinfo{booktitle}{{\em
  European conference on computer vision}}. Springer,
  \bibinfo{pages}{642--655}.
\newblock


\bibitem[\protect\citeauthoryear{Brox, Rosenhahn, Gall, and Cremers}{Brox
  et~al\mbox{.}}{2010}]%
        {brox2010combined}
\bibfield{author}{\bibinfo{person}{Thomas Brox}, \bibinfo{person}{Bodo
  Rosenhahn}, \bibinfo{person}{Juergen Gall}, {and} \bibinfo{person}{Daniel
  Cremers}.} \bibinfo{year}{2010}\natexlab{}.
\newblock \showarticletitle{Combined region and motion-based 3D tracking of
  rigid and articulated objects}.
\newblock \bibinfo{journal}{{\em IEEE Transactions on Pattern Analysis and
  Machine Intelligence\/}} \bibinfo{volume}{32}, \bibinfo{number}{3}
  (\bibinfo{year}{2010}), \bibinfo{pages}{402--415}.
\newblock


\bibitem[\protect\citeauthoryear{Cagniart, Boyer, and Ilic}{Cagniart
  et~al\mbox{.}}{2010}]%
        {cagniart2010free}
\bibfield{author}{\bibinfo{person}{Cedric Cagniart}, \bibinfo{person}{Edmond
  Boyer}, {and} \bibinfo{person}{Slobodan Ilic}.}
  \bibinfo{year}{2010}\natexlab{}.
\newblock \showarticletitle{Free-form mesh tracking: a patch-based approach}.
  In \bibinfo{booktitle}{{\em Computer Vision and Pattern Recognition (CVPR),
  2010 IEEE Conference on}}. IEEE, \bibinfo{pages}{1339--1346}.
\newblock


\bibitem[\protect\citeauthoryear{Cao, Bradley, Zhou, and Beeler}{Cao
  et~al\mbox{.}}{2015}]%
        {Cao:2015:RHF}
\bibfield{author}{\bibinfo{person}{Chen Cao}, \bibinfo{person}{Derek Bradley},
  \bibinfo{person}{Kun Zhou}, {and} \bibinfo{person}{Thabo Beeler}.}
  \bibinfo{year}{2015}\natexlab{}.
\newblock \showarticletitle{Real-time High-fidelity Facial Performance
  Capture}.
\newblock \bibinfo{journal}{{\em ACM Trans. Graph\/}} \bibinfo{volume}{34},
  \bibinfo{number}{4}, Article \bibinfo{articleno}{46} (\bibinfo{date}{July}
  \bibinfo{year}{2015}), \bibinfo{numpages}{9}~pages.
\newblock


\bibitem[\protect\citeauthoryear{Carranza, Theobalt, Magnor, and
  Seidel}{Carranza et~al\mbox{.}}{2003}]%
        {Carranza:2003}
\bibfield{author}{\bibinfo{person}{Joel Carranza}, \bibinfo{person}{Christian
  Theobalt}, \bibinfo{person}{Marcus~A. Magnor}, {and}
  \bibinfo{person}{Hans-Peter Seidel}.} \bibinfo{year}{2003}\natexlab{}.
\newblock \showarticletitle{Free-viewpoint Video of Human Actors}.
\newblock \bibinfo{journal}{{\em ACM Trans. Graph.\/}} \bibinfo{volume}{22},
  \bibinfo{number}{3} (\bibinfo{date}{July} \bibinfo{year}{2003}).
\newblock


\bibitem[\protect\citeauthoryear{Chen, Guo, Zhou, and Zhao}{Chen
  et~al\mbox{.}}{2013}]%
        {chen2013deformable}
\bibfield{author}{\bibinfo{person}{Xiaowu Chen}, \bibinfo{person}{Yu Guo},
  \bibinfo{person}{Bin Zhou}, {and} \bibinfo{person}{Qinping Zhao}.}
  \bibinfo{year}{2013}\natexlab{}.
\newblock \showarticletitle{Deformable model for estimating clothed and naked
  human shapes from a single image}.
\newblock \bibinfo{journal}{{\em The Visual Computer\/}} \bibinfo{volume}{29},
  \bibinfo{number}{11} (\bibinfo{year}{2013}), \bibinfo{pages}{1187--1196}.
\newblock


\bibitem[\protect\citeauthoryear{Collet, Chuang, Sweeney, Gillett, Evseev,
  Calabrese, Hoppe, Kirk, and Sullivan}{Collet et~al\mbox{.}}{2015}]%
        {collet2015high}
\bibfield{author}{\bibinfo{person}{Alvaro Collet}, \bibinfo{person}{Ming
  Chuang}, \bibinfo{person}{Pat Sweeney}, \bibinfo{person}{Don Gillett},
  \bibinfo{person}{Dennis Evseev}, \bibinfo{person}{David Calabrese},
  \bibinfo{person}{Hugues Hoppe}, \bibinfo{person}{Adam Kirk}, {and}
  \bibinfo{person}{Steve Sullivan}.} \bibinfo{year}{2015}\natexlab{}.
\newblock \showarticletitle{High-quality streamable free-viewpoint video}.
\newblock \bibinfo{journal}{{\em ACM Transactions on Graphics (TOG)\/}}
  \bibinfo{volume}{34}, \bibinfo{number}{4} (\bibinfo{year}{2015}),
  \bibinfo{pages}{69}.
\newblock


\bibitem[\protect\citeauthoryear{De~Aguiar, Stoll, Theobalt, Ahmed, Seidel, and
  Thrun}{De~Aguiar et~al\mbox{.}}{2008}]%
        {de2008performance}
\bibfield{author}{\bibinfo{person}{Edilson De~Aguiar}, \bibinfo{person}{Carsten
  Stoll}, \bibinfo{person}{Christian Theobalt}, \bibinfo{person}{Naveed Ahmed},
  \bibinfo{person}{Hans-Peter Seidel}, {and} \bibinfo{person}{Sebastian
  Thrun}.} \bibinfo{year}{2008}\natexlab{}.
\newblock \showarticletitle{Performance capture from sparse multi-view video}.
  In \bibinfo{booktitle}{{\em ACM Transactions on Graphics (TOG)}},
  Vol.~\bibinfo{volume}{27}. ACM, \bibinfo{pages}{98}.
\newblock


\bibitem[\protect\citeauthoryear{Dou, Davidson, Fanello, Khamis, Kowdle,
  Rhemann, Tankovich, and Izadi}{Dou et~al\mbox{.}}{2017}]%
        {Dou:2017}
\bibfield{author}{\bibinfo{person}{Mingsong Dou}, \bibinfo{person}{Philip
  Davidson}, \bibinfo{person}{Sean~Ryan Fanello}, \bibinfo{person}{Sameh
  Khamis}, \bibinfo{person}{Adarsh Kowdle}, \bibinfo{person}{Christoph
  Rhemann}, \bibinfo{person}{Vladimir Tankovich}, {and}
  \bibinfo{person}{Shahram Izadi}.} \bibinfo{year}{2017}\natexlab{}.
\newblock \showarticletitle{Motion2Fusion: Real-time Volumetric Performance
  Capture}.
\newblock \bibinfo{journal}{{\em ACM Trans. Graph.\/}} \bibinfo{volume}{36},
  \bibinfo{number}{6}, Article \bibinfo{articleno}{246} (\bibinfo{date}{Nov.}
  \bibinfo{year}{2017}), \bibinfo{numpages}{246:1--246:16}~pages.
\newblock


\bibitem[\protect\citeauthoryear{Dou, Khamis, Degtyarev, Davidson, Fanello,
  Kowdle, Escolano, Rhemann, Kim, Taylor, et~al\mbox{.}}{Dou
  et~al\mbox{.}}{2016}]%
        {dou2016fusion4d}
\bibfield{author}{\bibinfo{person}{Mingsong Dou}, \bibinfo{person}{Sameh
  Khamis}, \bibinfo{person}{Yury Degtyarev}, \bibinfo{person}{Philip Davidson},
  \bibinfo{person}{Sean~Ryan Fanello}, \bibinfo{person}{Adarsh Kowdle},
  \bibinfo{person}{Sergio~Orts Escolano}, \bibinfo{person}{Christoph Rhemann},
  \bibinfo{person}{David Kim}, \bibinfo{person}{Jonathan Taylor},
  {et~al\mbox{.}}} \bibinfo{year}{2016}\natexlab{}.
\newblock \showarticletitle{Fusion4d: Real-time performance capture of
  challenging scenes}.
\newblock \bibinfo{journal}{{\em ACM Transactions on Graphics (TOG)\/}}
  \bibinfo{volume}{35}, \bibinfo{number}{4} (\bibinfo{year}{2016}),
  \bibinfo{pages}{114}.
\newblock


\bibitem[\protect\citeauthoryear{Gall, Stoll, De~Aguiar, Theobalt, Rosenhahn,
  and Seidel}{Gall et~al\mbox{.}}{2009}]%
        {gall2009motion}
\bibfield{author}{\bibinfo{person}{Juergen Gall}, \bibinfo{person}{Carsten
  Stoll}, \bibinfo{person}{Edilson De~Aguiar}, \bibinfo{person}{Christian
  Theobalt}, \bibinfo{person}{Bodo Rosenhahn}, {and}
  \bibinfo{person}{Hans-Peter Seidel}.} \bibinfo{year}{2009}\natexlab{}.
\newblock \showarticletitle{Motion capture using joint skeleton tracking and
  surface estimation}. In \bibinfo{booktitle}{{\em Computer Vision and Pattern
  Recognition, 2009. CVPR 2009. IEEE Conference on}}. IEEE,
  \bibinfo{pages}{1746--1753}.
\newblock


\bibitem[\protect\citeauthoryear{Garg, Roussos, and Agapito}{Garg
  et~al\mbox{.}}{2013}]%
        {Garg_2013_CVPR}
\bibfield{author}{\bibinfo{person}{R. Garg}, \bibinfo{person}{A. Roussos},
  {and} \bibinfo{person}{L. Agapito}.} \bibinfo{year}{2013}\natexlab{}.
\newblock \showarticletitle{Dense Variational Reconstruction of Non-rigid
  Surfaces from Monocular Video}. In \bibinfo{booktitle}{{\em 2013 IEEE
  Conference on Computer Vision and Pattern Recognition}}.
  \bibinfo{pages}{1272--1279}.
\newblock
\showISSN{1063-6919}
\showDOI{%
\url{https://doi.org/10.1109/CVPR.2013.168}}


\bibitem[\protect\citeauthoryear{Garrido, Zollhoefer, Casas, Valgaerts,
  Varanasi, Perez, and Theobalt}{Garrido et~al\mbox{.}}{2016}]%
        {Garrido:2016}
\bibfield{author}{\bibinfo{person}{Pablo Garrido}, \bibinfo{person}{Michael
  Zollhoefer}, \bibinfo{person}{Dan Casas}, \bibinfo{person}{Levi Valgaerts},
  \bibinfo{person}{Kiran Varanasi}, \bibinfo{person}{Patrick Perez}, {and}
  \bibinfo{person}{Christian Theobalt}.} \bibinfo{year}{2016}\natexlab{}.
\newblock \showarticletitle{Reconstruction of Personalized 3D Face Rigs from
  Monocular Video}.
\newblock  \bibinfo{volume}{35}, \bibinfo{number}{3} (\bibinfo{year}{2016}),
  \bibinfo{pages}{28:1--28:15}.
\newblock


\bibitem[\protect\citeauthoryear{Gong, Liang, Zhang, Shen, and Lin}{Gong
  et~al\mbox{.}}{2017}]%
        {Gong_2017_CVPR}
\bibfield{author}{\bibinfo{person}{Ke Gong}, \bibinfo{person}{Xiaodan Liang},
  \bibinfo{person}{Dongyu Zhang}, \bibinfo{person}{Xiaohui Shen}, {and}
  \bibinfo{person}{Liang Lin}.} \bibinfo{year}{2017}\natexlab{}.
\newblock \showarticletitle{Look Into Person: Self-Supervised
  Structure-Sensitive Learning and a New Benchmark for Human Parsing}. In
  \bibinfo{booktitle}{{\em The IEEE Conference on Computer Vision and Pattern
  Recognition (CVPR)}}.
\newblock


\bibitem[\protect\citeauthoryear{Guan, Weiss, B{\u{a}}lan, and Black}{Guan
  et~al\mbox{.}}{2009}]%
        {guan2009estimating}
\bibfield{author}{\bibinfo{person}{Peng Guan}, \bibinfo{person}{Alexander
  Weiss}, \bibinfo{person}{Alexandru~O B{\u{a}}lan}, {and}
  \bibinfo{person}{Michael~J Black}.} \bibinfo{year}{2009}\natexlab{}.
\newblock \showarticletitle{Estimating human shape and pose from a single
  image}. In \bibinfo{booktitle}{{\em ICCV}}. \bibinfo{pages}{1381--1388}.
\newblock


\bibitem[\protect\citeauthoryear{G{\"u}ler, Neverova, and Kokkinos}{G{\"u}ler
  et~al\mbox{.}}{2018}]%
        {gue2018densepose}
\bibfield{author}{\bibinfo{person}{Riza~Alp G{\"u}ler},
  \bibinfo{person}{Natalia Neverova}, {and} \bibinfo{person}{Iasonas
  Kokkinos}.} \bibinfo{year}{2018}\natexlab{}.
\newblock \showarticletitle{DensePose: Dense Human Pose Estimation In The
  Wild}. In \bibinfo{booktitle}{{\em The IEEE Conference on Computer Vision and
  Pattern Recognition (CVPR)}}.
\newblock


\bibitem[\protect\citeauthoryear{Guo, Taylor, Fanello, Tagliasacchi, Dou,
  Davidson, Kowdle, and Izadi}{Guo et~al\mbox{.}}{2018}]%
        {twinfusionguo}
\bibfield{author}{\bibinfo{person}{Kaiwen Guo}, \bibinfo{person}{Jonathan
  Taylor}, \bibinfo{person}{Sean Fanello}, \bibinfo{person}{Andrea
  Tagliasacchi}, \bibinfo{person}{Mingsong Dou}, \bibinfo{person}{Philip
  Davidson}, \bibinfo{person}{Adarsh Kowdle}, {and} \bibinfo{person}{Shahram
  Izadi}.} \bibinfo{year}{2018}\natexlab{}.
\newblock \showarticletitle{TwinFusion: High Framerate Non-Rigid Fusion through
  Fast Correspondence Tracking}.
\newblock
\showDOI{%
\url{https://doi.org/10.1109/3DV.2018.00074}}


\bibitem[\protect\citeauthoryear{Guo, Xu, Yu, Liu, Dai, and Liu}{Guo
  et~al\mbox{.}}{2017}]%
        {guo2017real}
\bibfield{author}{\bibinfo{person}{Kaiwen Guo}, \bibinfo{person}{Feng Xu},
  \bibinfo{person}{Tao Yu}, \bibinfo{person}{Xiaoyang Liu},
  \bibinfo{person}{Qionghai Dai}, {and} \bibinfo{person}{Yebin Liu}.}
  \bibinfo{year}{2017}\natexlab{}.
\newblock \showarticletitle{Real-Time Geometry, Albedo, and Motion
  Reconstruction Using a Single RGB-D Camera}.
\newblock \bibinfo{journal}{{\em ACM Transactions on Graphics (TOG)\/}}
  \bibinfo{volume}{36}, \bibinfo{number}{3} (\bibinfo{year}{2017}),
  \bibinfo{pages}{32}.
\newblock


\bibitem[\protect\citeauthoryear{Guo, Chen, Zhou, and Zhao}{Guo
  et~al\mbox{.}}{2012}]%
        {guo2012clothed}
\bibfield{author}{\bibinfo{person}{Yu Guo}, \bibinfo{person}{Xiaowu Chen},
  \bibinfo{person}{Bin Zhou}, {and} \bibinfo{person}{Qinping Zhao}.}
  \bibinfo{year}{2012}\natexlab{}.
\newblock \showarticletitle{Clothed and naked human shapes estimation from a
  single image}.
\newblock \bibinfo{journal}{{\em Proc. of Computational Visual Media (CVM)\/}}
  (\bibinfo{year}{2012}), \bibinfo{pages}{43--50}.
\newblock


\bibitem[\protect\citeauthoryear{Hasler, Ackermann, Rosenhahn, Thorm{\"a}hlen,
  and Seidel}{Hasler et~al\mbox{.}}{2010}]%
        {hasler2010multilinear}
\bibfield{author}{\bibinfo{person}{Nils Hasler}, \bibinfo{person}{Hanno
  Ackermann}, \bibinfo{person}{Bodo Rosenhahn}, \bibinfo{person}{Thorsten
  Thorm{\"a}hlen}, {and} \bibinfo{person}{Hans-Peter Seidel}.}
  \bibinfo{year}{2010}\natexlab{}.
\newblock \showarticletitle{Multilinear pose and body shape estimation of
  dressed subjects from image sets}. In \bibinfo{booktitle}{{\em Computer
  Vision and Pattern Recognition (CVPR), 2010 IEEE Conference on}}. IEEE,
  \bibinfo{pages}{1823--1830}.
\newblock


\bibitem[\protect\citeauthoryear{Helten, Muller, Seidel, and Theobalt}{Helten
  et~al\mbox{.}}{2013}]%
        {Helten:2013}
\bibfield{author}{\bibinfo{person}{Thomas Helten}, \bibinfo{person}{Meinard
  Muller}, \bibinfo{person}{Hans-Peter Seidel}, {and}
  \bibinfo{person}{Christian Theobalt}.} \bibinfo{year}{2013}\natexlab{}.
\newblock \showarticletitle{Real-Time Body Tracking with One Depth Camera and
  Inertial Sensors}. In \bibinfo{booktitle}{{\em The IEEE International
  Conference on Computer Vision (ICCV)}}.
\newblock


\bibitem[\protect\citeauthoryear{Hilsmann and Eisert}{Hilsmann and
  Eisert}{2009}]%
        {Hilsmann:2009:TRC}
\bibfield{author}{\bibinfo{person}{Anna Hilsmann} {and} \bibinfo{person}{Peter
  Eisert}.} \bibinfo{year}{2009}\natexlab{}.
\newblock \showarticletitle{Tracking and Retexturing Cloth for Real-Time
  Virtual Clothing Applications}. In \bibinfo{booktitle}{{\em Proceedings of
  the 4th International Conference on Computer Vision/Computer Graphics
  CollaborationTechniques}} {\em (\bibinfo{series}{MIRAGE '09})}.
  \bibinfo{publisher}{Springer-Verlag}, \bibinfo{address}{Berlin, Heidelberg},
  \bibinfo{pages}{94--105}.
\newblock
\showISBNx{978-3-642-01810-7}
\showDOI{%
\url{https://doi.org/10.1007/978-3-642-01811-4_9}}


\bibitem[\protect\citeauthoryear{Huang, Allain, Franco, Navab, Ilic, and
  Boyer}{Huang et~al\mbox{.}}{2016}]%
        {huang2016volumetric}
\bibfield{author}{\bibinfo{person}{C.-H. Huang}, \bibinfo{person}{B. Allain},
  \bibinfo{person}{J.-S. Franco}, \bibinfo{person}{N. Navab},
  \bibinfo{person}{S. Ilic}, {and} \bibinfo{person}{E. Boyer}.}
  \bibinfo{year}{2016}\natexlab{}.
\newblock \showarticletitle{Volumetric 3D Tracking by Detection}. In
  \bibinfo{booktitle}{{\em Proc. CVPR}}.
\newblock


\bibitem[\protect\citeauthoryear{Huang, Bogo, Lassner, Kanazawa, Gehler,
  Romero, Akhter, and Black}{Huang et~al\mbox{.}}{2017}]%
        {MuVS:3DV:2017}
\bibfield{author}{\bibinfo{person}{Yinghao Huang}, \bibinfo{person}{Federica
  Bogo}, \bibinfo{person}{Christoph Lassner}, \bibinfo{person}{Angjoo
  Kanazawa}, \bibinfo{person}{Peter~V. Gehler}, \bibinfo{person}{Javier
  Romero}, \bibinfo{person}{Ijaz Akhter}, {and} \bibinfo{person}{Michael~J.
  Black}.} \bibinfo{year}{2017}\natexlab{}.
\newblock \showarticletitle{Towards Accurate Marker-less Human Shape and Pose
  Estimation over Time}. In \bibinfo{booktitle}{{\em International Conference
  on 3D Vision (3DV)}}.
\newblock


\bibitem[\protect\citeauthoryear{Innmann, Zollh{\"o}fer, Nie{\ss}ner, Theobalt,
  and Stamminger}{Innmann et~al\mbox{.}}{2016}]%
        {innmann2016volume}
\bibfield{author}{\bibinfo{person}{Matthias Innmann}, \bibinfo{person}{Michael
  Zollh{\"o}fer}, \bibinfo{person}{Matthias Nie{\ss}ner},
  \bibinfo{person}{Christian Theobalt}, {and} \bibinfo{person}{Marc
  Stamminger}.} \bibinfo{year}{2016}\natexlab{}.
\newblock \showarticletitle{{VolumeDeform: Real-time Volumetric Non-rigid
  Reconstruction}}.
\newblock  (\bibinfo{date}{October} \bibinfo{year}{2016}), 17.
\newblock


\bibitem[\protect\citeauthoryear{Izadi, Kim, Hilliges, Molyneaux, Newcombe,
  Kohli, Shotton, Hodges, Freeman, Davison, et~al\mbox{.}}{Izadi
  et~al\mbox{.}}{2011}]%
        {izadi2011kinectfusion}
\bibfield{author}{\bibinfo{person}{Shahram Izadi}, \bibinfo{person}{David Kim},
  \bibinfo{person}{Otmar Hilliges}, \bibinfo{person}{David Molyneaux},
  \bibinfo{person}{Richard Newcombe}, \bibinfo{person}{Pushmeet Kohli},
  \bibinfo{person}{Jamie Shotton}, \bibinfo{person}{Steve Hodges},
  \bibinfo{person}{Dustin Freeman}, \bibinfo{person}{Andrew Davison},
  {et~al\mbox{.}}} \bibinfo{year}{2011}\natexlab{}.
\newblock \showarticletitle{KinectFusion: real-time 3D reconstruction and
  interaction using a moving depth camera}. In \bibinfo{booktitle}{{\em Proc.
  UIST}}. ACM, \bibinfo{pages}{559--568}.
\newblock


\bibitem[\protect\citeauthoryear{Jain, Thorm\"{a}hlen, Seidel, and
  Theobalt}{Jain et~al\mbox{.}}{2010}]%
        {jain2010movie}
\bibfield{author}{\bibinfo{person}{Arjun Jain}, \bibinfo{person}{Thorsten
  Thorm\"{a}hlen}, \bibinfo{person}{Hans-Peter Seidel}, {and}
  \bibinfo{person}{Christian Theobalt}.} \bibinfo{year}{2010}\natexlab{}.
\newblock \showarticletitle{{MovieReshape}: Tracking and Reshaping of Humans in
  Videos}.
\newblock \bibinfo{journal}{{\em ACM Transactions on Graphics\/}}
  \bibinfo{volume}{29}, \bibinfo{number}{5} (\bibinfo{year}{2010}).
\newblock
\showDOI{%
\url{https://doi.org/10.1145/1866158.1866174}}


\bibitem[\protect\citeauthoryear{Joo, Simon, and Sheikh}{Joo
  et~al\mbox{.}}{2018}]%
        {Joo2018TotalCA}
\bibfield{author}{\bibinfo{person}{Hanbyul Joo}, \bibinfo{person}{Tomas Simon},
  {and} \bibinfo{person}{Yaser Sheikh}.} \bibinfo{year}{2018}\natexlab{}.
\newblock \showarticletitle{Total Capture: A 3D Deformation Model for Tracking
  Faces, Hands, and Bodies}.
\newblock \bibinfo{journal}{{\em CoRR\/}}  \bibinfo{volume}{abs/1801.01615}
  (\bibinfo{year}{2018}).
\newblock


\bibitem[\protect\citeauthoryear{Kadlecek, Ichim, Liu, Krivanek, and
  Kavan}{Kadlecek et~al\mbox{.}}{2016}]%
        {kadlecek-16-reconstructing}
\bibfield{author}{\bibinfo{person}{Petr Kadlecek},
  \bibinfo{person}{Alexandru-Eugen Ichim}, \bibinfo{person}{Tiantian Liu},
  \bibinfo{person}{Jaroslav Krivanek}, {and} \bibinfo{person}{Ladislav Kavan}.}
  \bibinfo{year}{2016}\natexlab{}.
\newblock \showarticletitle{Reconstructing Personalized Anatomical Models for
  Physics-based Body Animation}.
\newblock \bibinfo{journal}{{\em ACM Trans. Graph.\/}} \bibinfo{volume}{35},
  \bibinfo{number}{6} (\bibinfo{year}{2016}).
\newblock


\bibitem[\protect\citeauthoryear{Kanazawa, Black, Jacobs, and Malik}{Kanazawa
  et~al\mbox{.}}{2018}]%
        {hmrKanazawa17}
\bibfield{author}{\bibinfo{person}{Angjoo Kanazawa},
  \bibinfo{person}{Michael~J. Black}, \bibinfo{person}{David~W. Jacobs}, {and}
  \bibinfo{person}{Jitendra Malik}.} \bibinfo{year}{2018}\natexlab{}.
\newblock \showarticletitle{End-to-end Recovery of Human Shape and Pose}. In
  \bibinfo{booktitle}{{\em Computer Vision and Pattern Regognition (CVPR)}}.
\newblock


\bibitem[\protect\citeauthoryear{Kavan, Collins, {\v{Z}}{\'a}ra, and
  O'Sullivan}{Kavan et~al\mbox{.}}{2007}]%
        {Kavan2007SDQ}
\bibfield{author}{\bibinfo{person}{Ladislav Kavan}, \bibinfo{person}{Steven
  Collins}, \bibinfo{person}{Ji{\v{r}}{\'\i} {\v{Z}}{\'a}ra}, {and}
  \bibinfo{person}{Carol O'Sullivan}.} \bibinfo{year}{2007}\natexlab{}.
\newblock \showarticletitle{Skinning with dual quaternions}. In
  \bibinfo{booktitle}{{\em Proceedings of the 2007 symposium on Interactive 3D
  graphics and games}}. ACM, \bibinfo{pages}{39--46}.
\newblock


\bibitem[\protect\citeauthoryear{Kim, Pons-Moll, Pujades, Bang, Kim, Black, and
  Lee}{Kim et~al\mbox{.}}{2017}]%
        {Meekyoung:siggraph}
\bibfield{author}{\bibinfo{person}{Meekyoung Kim}, \bibinfo{person}{Gerard
  Pons-Moll}, \bibinfo{person}{Sergi Pujades}, \bibinfo{person}{Sungbae Bang},
  \bibinfo{person}{Jinwwok Kim}, \bibinfo{person}{Michael Black}, {and}
  \bibinfo{person}{Sung-Hee Lee}.} \bibinfo{year}{2017}\natexlab{}.
\newblock \showarticletitle{Data-Driven Physics for Human Soft Tissue
  Animation}.
\newblock \bibinfo{journal}{{\em ACM Transactions on Graphics, (Proc.
  SIGGRAPH)\/}} \bibinfo{volume}{36}, \bibinfo{number}{4}
  (\bibinfo{year}{2017}).
\newblock
\showURL{%
\url{http://dx.doi.org/10.1145/3072959.3073685}}


\bibitem[\protect\citeauthoryear{Kowdle, Rhemann, Fanello, Tagliasacchi,
  Taylor, Davidson, Dou, Guo, Keskin, Khamis, Kim, Tang, Tankovich, Valentin,
  and Izadi}{Kowdle et~al\mbox{.}}{2018}]%
        {NeedForSpeedKowdle}
\bibfield{author}{\bibinfo{person}{Adarsh Kowdle}, \bibinfo{person}{Christoph
  Rhemann}, \bibinfo{person}{Sean Fanello}, \bibinfo{person}{Andrea
  Tagliasacchi}, \bibinfo{person}{Jonathan Taylor}, \bibinfo{person}{Philip
  Davidson}, \bibinfo{person}{Mingsong Dou}, \bibinfo{person}{Kaiwen Guo},
  \bibinfo{person}{Cem Keskin}, \bibinfo{person}{Sameh Khamis},
  \bibinfo{person}{David Kim}, \bibinfo{person}{Danhang Tang},
  \bibinfo{person}{Vladimir Tankovich}, \bibinfo{person}{Julien Valentin},
  {and} \bibinfo{person}{Shahram Izadi}.} \bibinfo{year}{2018}\natexlab{}.
\newblock \showarticletitle{The Need 4 Speed in Real-time Dense Visual
  Tracking}. In \bibinfo{booktitle}{{\em SIGGRAPH Asia 2018 Technical Papers}}
  {\em (\bibinfo{series}{SIGGRAPH Asia '18})}. \bibinfo{publisher}{ACM},
  \bibinfo{address}{New York, NY, USA}, Article \bibinfo{articleno}{220},
  \bibinfo{numpages}{14}~pages.
\newblock
\showISBNx{978-1-4503-6008-1}
\showDOI{%
\url{https://doi.org/10.1145/3272127.3275062}}


\bibitem[\protect\citeauthoryear{Kraevoy, Sheffer, and van~de Panne}{Kraevoy
  et~al\mbox{.}}{2009}]%
        {kraevoy2009modeling}
\bibfield{author}{\bibinfo{person}{Vladislav Kraevoy}, \bibinfo{person}{Alla
  Sheffer}, {and} \bibinfo{person}{Michiel van~de Panne}.}
  \bibinfo{year}{2009}\natexlab{}.
\newblock \showarticletitle{Modeling from contour drawings}. In
  \bibinfo{booktitle}{{\em Proceedings of the 6th Eurographics Symposium on
  Sketch-Based interfaces and Modeling}}. ACM, \bibinfo{pages}{37--44}.
\newblock


\bibitem[\protect\citeauthoryear{Lassner, Romero, Kiefel, Bogo, Black, and
  V.Gehler}{Lassner et~al\mbox{.}}{2017}]%
        {Lassner}
\bibfield{author}{\bibinfo{person}{Christoph Lassner}, \bibinfo{person}{Javier
  Romero}, \bibinfo{person}{Martin Kiefel}, \bibinfo{person}{Federica Bogo},
  \bibinfo{person}{Michael~J. Black}, {and} \bibinfo{person}{Peter V.Gehler}.}
  \bibinfo{year}{2017}\natexlab{}.
\newblock \showarticletitle{Unite the People: Closing the Loop Between 3D and
  2D Human Representations}. In \bibinfo{booktitle}{{\em Proc. CVPR}}.
\newblock


\bibitem[\protect\citeauthoryear{Leroy, Franco, and Boyer}{Leroy
  et~al\mbox{.}}{2017}]%
        {inria_2017}
\bibfield{author}{\bibinfo{person}{Vincent Leroy},
  \bibinfo{person}{Jean-S{\'e}bastien Franco}, {and} \bibinfo{person}{Edmond
  Boyer}.} \bibinfo{year}{2017}\natexlab{}.
\newblock \showarticletitle{{Multi-View Dynamic Shape Refinement Using Local
  Temporal Integration}}. In \bibinfo{booktitle}{{\em {IEEE, International
  Conference on Computer Vision 2017}}}. \bibinfo{address}{Venice, Italy}.
\newblock
\showURL{%
\url{https://hal.archives-ouvertes.fr/hal-01567758}}


\bibitem[\protect\citeauthoryear{Liu, Stoll, Gall, Seidel, and Theobalt}{Liu
  et~al\mbox{.}}{2011}]%
        {liu2011markerless}
\bibfield{author}{\bibinfo{person}{Yebin Liu}, \bibinfo{person}{Carsten Stoll},
  \bibinfo{person}{Juergen Gall}, \bibinfo{person}{Hans-Peter Seidel}, {and}
  \bibinfo{person}{Christian Theobalt}.} \bibinfo{year}{2011}\natexlab{}.
\newblock \showarticletitle{Markerless motion capture of interacting characters
  using multi-view image segmentation}. In \bibinfo{booktitle}{{\em Computer
  Vision and Pattern Recognition (CVPR), 2011 IEEE Conference on}}. IEEE,
  \bibinfo{pages}{1249--1256}.
\newblock


\bibitem[\protect\citeauthoryear{Loper, Mahmood, Romero, Pons-Moll, and
  Black}{Loper et~al\mbox{.}}{2015}]%
        {loper2015smpl}
\bibfield{author}{\bibinfo{person}{Matthew Loper}, \bibinfo{person}{Naureen
  Mahmood}, \bibinfo{person}{Javier Romero}, \bibinfo{person}{Gerard
  Pons-Moll}, {and} \bibinfo{person}{Michael~J Black}.}
  \bibinfo{year}{2015}\natexlab{}.
\newblock \showarticletitle{{SMPL: A skinned multi-person linear model}}.
\newblock \bibinfo{journal}{{\em ACM Transactions on Graphics (TOG)\/}}
  \bibinfo{volume}{34}, \bibinfo{number}{6} (\bibinfo{year}{2015}).
\newblock


\bibitem[\protect\citeauthoryear{Matusik, Buehler, Raskar, Gortler, and
  McMillan}{Matusik et~al\mbox{.}}{2000}]%
        {matusik2000image}
\bibfield{author}{\bibinfo{person}{Wojciech Matusik}, \bibinfo{person}{Chris
  Buehler}, \bibinfo{person}{Ramesh Raskar}, \bibinfo{person}{Steven~J
  Gortler}, {and} \bibinfo{person}{Leonard McMillan}.}
  \bibinfo{year}{2000}\natexlab{}.
\newblock \showarticletitle{Image-based visual hulls}. In
  \bibinfo{booktitle}{{\em Proceedings of the 27th annual conference on
  Computer graphics and interactive techniques}}. ACM Press/Addison-Wesley
  Publishing Co., \bibinfo{pages}{369--374}.
\newblock


\bibitem[\protect\citeauthoryear{Mehta, Sridhar, Sotnychenko, Rhodin, Shafiei,
  Seidel, Xu, Casas, and Theobalt}{Mehta et~al\mbox{.}}{2017}]%
        {VNect_SIGGRAPH2017}
\bibfield{author}{\bibinfo{person}{Dushyant Mehta}, \bibinfo{person}{Srinath
  Sridhar}, \bibinfo{person}{Oleksandr Sotnychenko}, \bibinfo{person}{Helge
  Rhodin}, \bibinfo{person}{Mohammad Shafiei}, \bibinfo{person}{Hans-Peter
  Seidel}, \bibinfo{person}{Weipeng Xu}, \bibinfo{person}{Dan Casas}, {and}
  \bibinfo{person}{Christian Theobalt}.} \bibinfo{year}{2017}\natexlab{}.
\newblock \showarticletitle{VNect: Real-time 3D Human Pose Estimation with a
  Single RGB Camera}.
\newblock \bibinfo{journal}{{\em ACM Transactions on Graphics\/}}
  \bibinfo{volume}{36}, \bibinfo{number}{4}, 14.
\newblock
\showDOI{%
\url{https://doi.org/10.1145/3072959.3073596}}


\bibitem[\protect\citeauthoryear{Metaxas and Terzopoulos}{Metaxas and
  Terzopoulos}{1993}]%
        {metaxas1993shape}
\bibfield{author}{\bibinfo{person}{Dimitris Metaxas} {and}
  \bibinfo{person}{Demetri Terzopoulos}.} \bibinfo{year}{1993}\natexlab{}.
\newblock \showarticletitle{Shape and nonrigid motion estimation through
  physics-based synthesis}.
\newblock \bibinfo{journal}{{\em IEEE Trans. PAMI\/}} \bibinfo{volume}{15},
  \bibinfo{number}{6} (\bibinfo{year}{1993}), \bibinfo{pages}{580--591}.
\newblock


\bibitem[\protect\citeauthoryear{Mustafa, Kim, Guillemaut, and Hilton}{Mustafa
  et~al\mbox{.}}{2016}]%
        {Mustafa:16}
\bibfield{author}{\bibinfo{person}{Armin Mustafa}, \bibinfo{person}{Hansung
  Kim}, \bibinfo{person}{Jean{-}Yves Guillemaut}, {and} \bibinfo{person}{Adrian
  Hilton}.} \bibinfo{year}{2016}\natexlab{}.
\newblock \showarticletitle{Temporally Coherent 4D Reconstruction of Complex
  Dynamic Scenes}. In \bibinfo{booktitle}{{\em CVPR}}.
  \bibinfo{pages}{4660--4669}.
\newblock
\showDOI{%
\url{https://doi.org/10.1109/CVPR.2016.504}}


\bibitem[\protect\citeauthoryear{Newcombe, Fox, and Seitz}{Newcombe
  et~al\mbox{.}}{2015}]%
        {Newcombe_2015_CVPR}
\bibfield{author}{\bibinfo{person}{Richard~A. Newcombe},
  \bibinfo{person}{Dieter Fox}, {and} \bibinfo{person}{Steven~M. Seitz}.}
  \bibinfo{year}{2015}\natexlab{}.
\newblock \showarticletitle{DynamicFusion: Reconstruction and Tracking of
  Non-Rigid Scenes in Real-Time}. In \bibinfo{booktitle}{{\em The IEEE
  Conference on Computer Vision and Pattern Recognition (CVPR)}}.
\newblock


\bibitem[\protect\citeauthoryear{Newcombe, Izadi, Hilliges, Molyneaux, Kim,
  Davison, Kohi, Shotton, Hodges, and Fitzgibbon}{Newcombe
  et~al\mbox{.}}{2011}]%
        {newcombe2011kinectfusion}
\bibfield{author}{\bibinfo{person}{Richard~A Newcombe},
  \bibinfo{person}{Shahram Izadi}, \bibinfo{person}{Otmar Hilliges},
  \bibinfo{person}{David Molyneaux}, \bibinfo{person}{David Kim},
  \bibinfo{person}{Andrew~J Davison}, \bibinfo{person}{Pushmeet Kohi},
  \bibinfo{person}{Jamie Shotton}, \bibinfo{person}{Steve Hodges}, {and}
  \bibinfo{person}{Andrew Fitzgibbon}.} \bibinfo{year}{2011}\natexlab{}.
\newblock \showarticletitle{KinectFusion: Real-time dense surface mapping and
  tracking}. In \bibinfo{booktitle}{{\em Proc. ISMAR}}. IEEE,
  \bibinfo{pages}{127--136}.
\newblock


\bibitem[\protect\citeauthoryear{Orts-Escolano, Rhemann, Fanello, Chang,
  Kowdle, Degtyarev, Kim, Davidson, Khamis, Dou, et~al\mbox{.}}{Orts-Escolano
  et~al\mbox{.}}{2016}]%
        {orts2016holoportation}
\bibfield{author}{\bibinfo{person}{Sergio Orts-Escolano},
  \bibinfo{person}{Christoph Rhemann}, \bibinfo{person}{Sean Fanello},
  \bibinfo{person}{Wayne Chang}, \bibinfo{person}{Adarsh Kowdle},
  \bibinfo{person}{Yury Degtyarev}, \bibinfo{person}{David Kim},
  \bibinfo{person}{Philip~L Davidson}, \bibinfo{person}{Sameh Khamis},
  \bibinfo{person}{Mingsong Dou}, {et~al\mbox{.}}}
  \bibinfo{year}{2016}\natexlab{}.
\newblock \showarticletitle{Holoportation: Virtual 3D Teleportation in
  Real-time}. In \bibinfo{booktitle}{{\em Proceedings of the 29th Annual
  Symposium on User Interface Software and Technology}}. ACM,
  \bibinfo{pages}{741--754}.
\newblock


\bibitem[\protect\citeauthoryear{Park and Hodgins}{Park and Hodgins}{2008}]%
        {park2008data}
\bibfield{author}{\bibinfo{person}{Sang~Il Park} {and}
  \bibinfo{person}{Jessica~K Hodgins}.} \bibinfo{year}{2008}\natexlab{}.
\newblock \showarticletitle{Data-driven modeling of skin and muscle
  deformation}. In \bibinfo{booktitle}{{\em ACM Transactions on Graphics
  (TOG)}}, Vol.~\bibinfo{volume}{27}. ACM, \bibinfo{pages}{96}.
\newblock


\bibitem[\protect\citeauthoryear{Pl{\"a}nkers and Fua}{Pl{\"a}nkers and
  Fua}{2001}]%
        {plankers2001tracking}
\bibfield{author}{\bibinfo{person}{Ralf Pl{\"a}nkers} {and}
  \bibinfo{person}{Pascal Fua}.} \bibinfo{year}{2001}\natexlab{}.
\newblock \showarticletitle{Tracking and modeling people in video sequences}.
\newblock \bibinfo{journal}{{\em Computer Vision and Image Understanding\/}}
  \bibinfo{volume}{81}, \bibinfo{number}{3} (\bibinfo{year}{2001}),
  \bibinfo{pages}{285--302}.
\newblock


\bibitem[\protect\citeauthoryear{Pons-Moll, Pujades, Hu, and Black}{Pons-Moll
  et~al\mbox{.}}{2017}]%
        {Pons-Moll:Siggraph2017}
\bibfield{author}{\bibinfo{person}{Gerard Pons-Moll}, \bibinfo{person}{Sergi
  Pujades}, \bibinfo{person}{Sonny Hu}, {and} \bibinfo{person}{Michael Black}.}
  \bibinfo{year}{2017}\natexlab{}.
\newblock \showarticletitle{{ClothCap}: Seamless {4D} Clothing Capture and
  Retargeting}.
\newblock \bibinfo{journal}{{\em ACM Transactions on Graphics, (Proc.
  SIGGRAPH)\/}} \bibinfo{volume}{36}, \bibinfo{number}{4}
  (\bibinfo{year}{2017}).
\newblock
\showURL{%
\url{http://dx.doi.org/10.1145/3072959.3073711}}


\bibitem[\protect\citeauthoryear{Pons-Moll, Romero, Mahmood, and
  Black}{Pons-Moll et~al\mbox{.}}{2015}]%
        {Pons-Moll:Siggraph2015}
\bibfield{author}{\bibinfo{person}{Gerard Pons-Moll}, \bibinfo{person}{Javier
  Romero}, \bibinfo{person}{Naureen Mahmood}, {and} \bibinfo{person}{Michael~J
  Black}.} \bibinfo{year}{2015}\natexlab{}.
\newblock \showarticletitle{Dyna: a model of dynamic human shape in motion}.
\newblock \bibinfo{journal}{{\em ACM Transactions on Graphics (TOG)\/}}
  \bibinfo{volume}{34}, \bibinfo{number}{4} (\bibinfo{year}{2015}),
  \bibinfo{pages}{120}.
\newblock


\bibitem[\protect\citeauthoryear{Popa, Zanfir, and Sminchisescu}{Popa
  et~al\mbox{.}}{2017}]%
        {popa2017deep}
\bibfield{author}{\bibinfo{person}{Alin-Ionut Popa}, \bibinfo{person}{Mihai
  Zanfir}, {and} \bibinfo{person}{Cristian Sminchisescu}.}
  \bibinfo{year}{2017}\natexlab{}.
\newblock \showarticletitle{Deep Multitask Architecture for Integrated 2D and
  3D Human Sensing}.
\newblock \bibinfo{journal}{{\em IEEE Conference on Computer Vision and Pattern
  Recognition (CVPR)\/}} (\bibinfo{year}{2017}).
\newblock


\bibitem[\protect\citeauthoryear{Prada, Kazhdan, Chuang, Collet, and
  Hoppe}{Prada et~al\mbox{.}}{2017}]%
        {prada2017spatiotemporal}
\bibfield{author}{\bibinfo{person}{Fabi{\'a}n Prada}, \bibinfo{person}{Misha
  Kazhdan}, \bibinfo{person}{Ming Chuang}, \bibinfo{person}{Alvaro Collet},
  {and} \bibinfo{person}{Hugues Hoppe}.} \bibinfo{year}{2017}\natexlab{}.
\newblock \showarticletitle{Spatiotemporal atlas parameterization for evolving
  meshes}.
\newblock \bibinfo{journal}{{\em ACM Transactions on Graphics (TOG)\/}}
  \bibinfo{volume}{36}, \bibinfo{number}{4} (\bibinfo{year}{2017}),
  \bibinfo{pages}{58}.
\newblock


\bibitem[\protect\citeauthoryear{Rhodin, Robertini, Casas, Richardt, Seidel,
  and Theobalt}{Rhodin et~al\mbox{.}}{2016}]%
        {rhodin2016general}
\bibfield{author}{\bibinfo{person}{Helge Rhodin}, \bibinfo{person}{Nadia
  Robertini}, \bibinfo{person}{Dan Casas}, \bibinfo{person}{Christian
  Richardt}, \bibinfo{person}{Hans-Peter Seidel}, {and}
  \bibinfo{person}{Christian Theobalt}.} \bibinfo{year}{2016}\natexlab{}.
\newblock \showarticletitle{General Automatic Human Shape and Motion Capture
  Using Volumetric Contour Cues}. In \bibinfo{booktitle}{{\em ECCV}},
  \bibfield{editor}{\bibinfo{person}{Bastian Leibe}, \bibinfo{person}{Jiri
  Matas}, \bibinfo{person}{Nicu Sebe}, {and} \bibinfo{person}{Max Welling}}
  (Eds.). \bibinfo{publisher}{Springer International Publishing},
  \bibinfo{address}{Cham}, \bibinfo{pages}{509--526}.
\newblock


\bibitem[\protect\citeauthoryear{Robertini, Casas, Rhodin, Seidel, and
  Theobalt}{Robertini et~al\mbox{.}}{2016}]%
        {robertini2016model}
\bibfield{author}{\bibinfo{person}{Nadia Robertini}, \bibinfo{person}{Dan
  Casas}, \bibinfo{person}{Helge Rhodin}, \bibinfo{person}{Hans-Peter Seidel},
  {and} \bibinfo{person}{Christian Theobalt}.} \bibinfo{year}{2016}\natexlab{}.
\newblock \showarticletitle{{Model-based Outdoor Performance Capture}}. In
  \bibinfo{booktitle}{{\em International Conference on Computer Vision (3DV)}}.
\newblock


\bibitem[\protect\citeauthoryear{Rogez, Weinzaepfel, and Schmid}{Rogez
  et~al\mbox{.}}{2017}]%
        {rogez_lcr_cvpr17}
\bibfield{author}{\bibinfo{person}{Gregory Rogez}, \bibinfo{person}{Philippe
  Weinzaepfel}, {and} \bibinfo{person}{Cordelia Schmid}.}
  \bibinfo{year}{2017}\natexlab{}.
\newblock \showarticletitle{LCR-Net: Localization-Classification-Regression for
  Human Pose}. In \bibinfo{booktitle}{{\em CVPR 2017-IEEE Conference on
  Computer Vision \& Pattern Recognition}}.
\newblock


\bibitem[\protect\citeauthoryear{Rogge, Klose, Stengel, Eisemann, and
  Magnor}{Rogge et~al\mbox{.}}{2014}]%
        {rogge2014garment}
\bibfield{author}{\bibinfo{person}{Lorenz Rogge}, \bibinfo{person}{Felix
  Klose}, \bibinfo{person}{Michael Stengel}, \bibinfo{person}{Martin Eisemann},
  {and} \bibinfo{person}{Marcus Magnor}.} \bibinfo{year}{2014}\natexlab{}.
\newblock \showarticletitle{Garment replacement in monocular video sequences}.
\newblock \bibinfo{journal}{{\em ACM Transactions on Graphics (TOG)\/}}
  \bibinfo{volume}{34}, \bibinfo{number}{1} (\bibinfo{year}{2014}),
  \bibinfo{pages}{6}.
\newblock


\bibitem[\protect\citeauthoryear{Romero, Tzionas, and Black}{Romero
  et~al\mbox{.}}{2017}]%
        {MANO:SIGGRAPHASIA:2017}
\bibfield{author}{\bibinfo{person}{Javier Romero}, \bibinfo{person}{Dimitrios
  Tzionas}, {and} \bibinfo{person}{Michael~J. Black}.}
  \bibinfo{year}{2017}\natexlab{}.
\newblock \showarticletitle{Embodied Hands: Modeling and Capturing Hands and
  Bodies Together}.
\newblock \bibinfo{journal}{{\em ACM Transactions on Graphics, (Proc. SIGGRAPH
  Asia)\/}} \bibinfo{volume}{36}, \bibinfo{number}{6} (\bibinfo{date}{Nov.}
  \bibinfo{year}{2017}), \bibinfo{pages}{245:1--245:17}.
\newblock
\showURL{%
\url{http://doi.acm.org/10.1145/3130800.3130883}}


\bibitem[\protect\citeauthoryear{Russell, Yu, and Agapito}{Russell
  et~al\mbox{.}}{2014}]%
        {Russell2014}
\bibfield{author}{\bibinfo{person}{Chris Russell}, \bibinfo{person}{Rui Yu},
  {and} \bibinfo{person}{Lourdes Agapito}.} \bibinfo{year}{2014}\natexlab{}.
\newblock \bibinfo{booktitle}{{\em Video Pop-up: Monocular 3D Reconstruction of
  Dynamic Scenes}}.
\newblock \bibinfo{publisher}{Springer International Publishing},
  \bibinfo{address}{Cham}, \bibinfo{pages}{583--598}.
\newblock
\showISBNx{978-3-319-10584-0}
\showDOI{%
\url{https://doi.org/10.1007/978-3-319-10584-0_38}}


\bibitem[\protect\citeauthoryear{Salzmann and Fua}{Salzmann and Fua}{2011}]%
        {Salzmann2011}
\bibfield{author}{\bibinfo{person}{Mathieu Salzmann} {and}
  \bibinfo{person}{Pascal Fua}.} \bibinfo{year}{2011}\natexlab{}.
\newblock \showarticletitle{{Linear local models for monocular reconstruction
  of deformable surfaces}}.
\newblock \bibinfo{journal}{{\em IEEE Transactions on Pattern Analysis and
  Machine Intelligence\/}} \bibinfo{volume}{33}, \bibinfo{number}{5}
  (\bibinfo{year}{2011}), \bibinfo{pages}{931--944}.
\newblock
\showISBNx{978-1-4244-2242-5}
\showISSN{01628828}
\showDOI{%
\url{https://doi.org/10.1109/TPAMI.2010.158}}


\bibitem[\protect\citeauthoryear{Saragih, Lucey, and Cohn}{Saragih
  et~al\mbox{.}}{2009}]%
        {saragig_tracker}
\bibfield{author}{\bibinfo{person}{J.~M. Saragih}, \bibinfo{person}{S. Lucey},
  {and} \bibinfo{person}{J.~F. Cohn}.} \bibinfo{year}{2009}\natexlab{}.
\newblock \showarticletitle{Face alignment through subspace constrained
  mean-shifts}. In \bibinfo{booktitle}{{\em 2009 IEEE 12th International
  Conference on Computer Vision}}. \bibinfo{pages}{1034--1041}.
\newblock
\showISSN{1550-5499}
\showDOI{%
\url{https://doi.org/10.1109/ICCV.2009.5459377}}


\bibitem[\protect\citeauthoryear{Sekine, Sugita, Perbet, Stenger, and
  Nishiyama}{Sekine et~al\mbox{.}}{2014}]%
        {Sekine}
\bibfield{author}{\bibinfo{person}{M. Sekine}, \bibinfo{person}{K. Sugita},
  \bibinfo{person}{F. Perbet}, \bibinfo{person}{B. Stenger}, {and}
  \bibinfo{person}{M. Nishiyama}.} \bibinfo{year}{2014}\natexlab{}.
\newblock \showarticletitle{Virtual Fitting by Single-Shot Body Shape
  Estimation}. In \bibinfo{booktitle}{{\em Int. Conf. on 3D Body Scanning
  Technologies}}. \bibinfo{pages}{406--413}.
\newblock


\bibitem[\protect\citeauthoryear{Sigal, Bhatia, Roth, Black, and Isard}{Sigal
  et~al\mbox{.}}{2004}]%
        {sigal2004tracking}
\bibfield{author}{\bibinfo{person}{Leonid Sigal}, \bibinfo{person}{Sidharth
  Bhatia}, \bibinfo{person}{Stefan Roth}, \bibinfo{person}{Michael~J Black},
  {and} \bibinfo{person}{Michael Isard}.} \bibinfo{year}{2004}\natexlab{}.
\newblock \showarticletitle{Tracking loose-limbed people}. In
  \bibinfo{booktitle}{{\em Computer Vision and Pattern Recognition, 2004. CVPR
  2004. Proceedings of the 2004 IEEE Computer Society Conference on}},
  Vol.~\bibinfo{volume}{1}. IEEE, \bibinfo{pages}{I--421}.
\newblock


\bibitem[\protect\citeauthoryear{Slavcheva, Baust, Cremers, and Ilic}{Slavcheva
  et~al\mbox{.}}{2017}]%
        {slavcheva2017killingfusion}
\bibfield{author}{\bibinfo{person}{Miroslava Slavcheva},
  \bibinfo{person}{Maximilian Baust}, \bibinfo{person}{Daniel Cremers}, {and}
  \bibinfo{person}{Slobodan Ilic}.} \bibinfo{year}{2017}\natexlab{}.
\newblock \showarticletitle{KillingFusion: Non-rigid 3D Reconstruction without
  Correspondences}. In \bibinfo{booktitle}{{\em IEEE Conference on Computer
  Vision and Pattern Recognition (CVPR)}}, Vol.~\bibinfo{volume}{3}.
  \bibinfo{pages}{7}.
\newblock


\bibitem[\protect\citeauthoryear{Sminchisescu and Triggs}{Sminchisescu and
  Triggs}{2003}]%
        {sminchisescu2003kinematic}
\bibfield{author}{\bibinfo{person}{Cristian Sminchisescu} {and}
  \bibinfo{person}{Bill Triggs}.} \bibinfo{year}{2003}\natexlab{}.
\newblock \showarticletitle{Kinematic jump processes for monocular 3D human
  tracking}. In \bibinfo{booktitle}{{\em Computer Vision and Pattern
  Recognition, 2003. Proceedings. 2003 IEEE Computer Society Conference on}},
  Vol.~\bibinfo{volume}{1}. IEEE, \bibinfo{pages}{I--69}.
\newblock


\bibitem[\protect\citeauthoryear{Starck and Hilton}{Starck and Hilton}{2007}]%
        {starck2007surface}
\bibfield{author}{\bibinfo{person}{Jonathan Starck} {and}
  \bibinfo{person}{Adrian Hilton}.} \bibinfo{year}{2007}\natexlab{}.
\newblock \showarticletitle{Surface capture for performance-based animation}.
\newblock \bibinfo{journal}{{\em IEEE Computer Graphics and Applications\/}}
  \bibinfo{volume}{27}, \bibinfo{number}{3} (\bibinfo{year}{2007}),
  \bibinfo{pages}{21--31}.
\newblock


\bibitem[\protect\citeauthoryear{Sun, Shang, Liang, and Wei}{Sun
  et~al\mbox{.}}{2017}]%
        {sun2017compositional}
\bibfield{author}{\bibinfo{person}{Xiao Sun}, \bibinfo{person}{Jiaxiang Shang},
  \bibinfo{person}{Shuang Liang}, {and} \bibinfo{person}{Yichen Wei}.}
  \bibinfo{year}{2017}\natexlab{}.
\newblock \showarticletitle{Compositional Human Pose Regression}.
\newblock \bibinfo{journal}{{\em ICCV\/}} (\bibinfo{year}{2017}).
\newblock


\bibitem[\protect\citeauthoryear{Tagliasacchi, Schroeder, Tkach, Bouaziz,
  Botsch, and Pauly}{Tagliasacchi et~al\mbox{.}}{2015}]%
        {tagliasacchi2015robust}
\bibfield{author}{\bibinfo{person}{Andrea Tagliasacchi},
  \bibinfo{person}{Matthias Schroeder}, \bibinfo{person}{Anastasia Tkach},
  \bibinfo{person}{Sofien Bouaziz}, \bibinfo{person}{Mario Botsch}, {and}
  \bibinfo{person}{Mark Pauly}.} \bibinfo{year}{2015}\natexlab{}.
\newblock \showarticletitle{Robust Articulated-ICP for Real-Time Hand
  Tracking}.
\newblock \bibinfo{journal}{{\em Computer Graphics Forum (Symposium on Geometry
  Processing)\/}} \bibinfo{volume}{34}, \bibinfo{number}{5}
  (\bibinfo{year}{2015}).
\newblock


\bibitem[\protect\citeauthoryear{Tekin, M{\'{a}}rquez{-}Neila, Salzmann, and
  Fua}{Tekin et~al\mbox{.}}{2017}]%
        {TekinMSF17}
\bibfield{author}{\bibinfo{person}{Bugra Tekin}, \bibinfo{person}{Pablo
  M{\'{a}}rquez{-}Neila}, \bibinfo{person}{Mathieu Salzmann}, {and}
  \bibinfo{person}{Pascal Fua}.} \bibinfo{year}{2017}\natexlab{}.
\newblock \showarticletitle{Learning to Fuse 2D and 3D Image Cues for Monocular
  Body Pose Estimation}. In \bibinfo{booktitle}{{\em {ICCV}}}.
  \bibinfo{publisher}{{IEEE} Computer Society}, \bibinfo{pages}{3961--3970}.
\newblock


\bibitem[\protect\citeauthoryear{Tekin, Rozantsev, Lepetit, and Fua}{Tekin
  et~al\mbox{.}}{2016}]%
        {Tekin2016}
\bibfield{author}{\bibinfo{person}{B. Tekin}, \bibinfo{person}{A. Rozantsev},
  \bibinfo{person}{V. Lepetit}, {and} \bibinfo{person}{P. Fua}.}
  \bibinfo{year}{2016}\natexlab{}.
\newblock \showarticletitle{Direct Prediction of 3D Body Poses from Motion
  Compensated Sequences}. In \bibinfo{booktitle}{{\em 2016 IEEE Conference on
  Computer Vision and Pattern Recognition (CVPR)}}. \bibinfo{pages}{991--1000}.
\newblock


\bibitem[\protect\citeauthoryear{Tome, Russell, and Agapito}{Tome
  et~al\mbox{.}}{2017}]%
        {tome2017lifting}
\bibfield{author}{\bibinfo{person}{Denis Tome}, \bibinfo{person}{Chris
  Russell}, {and} \bibinfo{person}{Lourdes Agapito}.}
  \bibinfo{year}{2017}\natexlab{}.
\newblock \showarticletitle{Lifting from the deep: Convolutional 3d pose
  estimation from a single image}.
\newblock \bibinfo{journal}{{\em IEEE Conf. on Computer Vision and Pattern
  Recognition. Proceedings\/}} (\bibinfo{year}{2017}).
\newblock


\bibitem[\protect\citeauthoryear{Varol, Ceylan, Russell, Yang, Yumer, Laptev,
  and Schmid}{Varol et~al\mbox{.}}{2018}]%
        {varol18_bodynet}
\bibfield{author}{\bibinfo{person}{G{\"u}l Varol}, \bibinfo{person}{Duygu
  Ceylan}, \bibinfo{person}{Bryan Russell}, \bibinfo{person}{Jimei Yang},
  \bibinfo{person}{Ersin Yumer}, \bibinfo{person}{Ivan Laptev}, {and}
  \bibinfo{person}{Cordelia Schmid}.} \bibinfo{year}{2018}\natexlab{}.
\newblock \showarticletitle{{BodyNet}: Volumetric Inference of {3D} Human Body
  Shapes}. In \bibinfo{booktitle}{{\em ECCV}}.
\newblock


\bibitem[\protect\citeauthoryear{Vlasic, Baran, Matusik, and
  Popovi{\'c}}{Vlasic et~al\mbox{.}}{2008}]%
        {vlasic2008articulated}
\bibfield{author}{\bibinfo{person}{Daniel Vlasic}, \bibinfo{person}{Ilya
  Baran}, \bibinfo{person}{Wojciech Matusik}, {and} \bibinfo{person}{Jovan
  Popovi{\'c}}.} \bibinfo{year}{2008}\natexlab{}.
\newblock \showarticletitle{Articulated mesh animation from multi-view
  silhouettes}. In \bibinfo{booktitle}{{\em ACM Transactions on Graphics
  (TOG)}}, Vol.~\bibinfo{volume}{27}. ACM, \bibinfo{pages}{97}.
\newblock


\bibitem[\protect\citeauthoryear{Vlasic, Peers, Baran, Debevec, Popovi{\'c},
  Rusinkiewicz, and Matusik}{Vlasic et~al\mbox{.}}{2009}]%
        {vlasic2009dynamic}
\bibfield{author}{\bibinfo{person}{Daniel Vlasic}, \bibinfo{person}{Pieter
  Peers}, \bibinfo{person}{Ilya Baran}, \bibinfo{person}{Paul Debevec},
  \bibinfo{person}{Jovan Popovi{\'c}}, \bibinfo{person}{Szymon Rusinkiewicz},
  {and} \bibinfo{person}{Wojciech Matusik}.} \bibinfo{year}{2009}\natexlab{}.
\newblock \showarticletitle{Dynamic shape capture using multi-view photometric
  stereo}.
\newblock \bibinfo{journal}{{\em ACM Transactions on Graphics (TOG)\/}}
  \bibinfo{volume}{28}, \bibinfo{number}{5} (\bibinfo{year}{2009}),
  \bibinfo{pages}{174}.
\newblock


\bibitem[\protect\citeauthoryear{Wang, Wei, Vouga, Huang, Ceylan, Medioni, and
  Li}{Wang et~al\mbox{.}}{2016}]%
        {wang2016capturing}
\bibfield{author}{\bibinfo{person}{Ruizhe Wang}, \bibinfo{person}{Lingyu Wei},
  \bibinfo{person}{Etienne Vouga}, \bibinfo{person}{Qixing Huang},
  \bibinfo{person}{Duygu Ceylan}, \bibinfo{person}{Gerard Medioni}, {and}
  \bibinfo{person}{Hao Li}.} \bibinfo{year}{2016}\natexlab{}.
\newblock \showarticletitle{Capturing Dynamic Textured Surfaces of Moving
  Targets}. In \bibinfo{booktitle}{{\em Proceedings of the European Conference
  on Computer Vision ({ECCV})}}.
\newblock


\bibitem[\protect\citeauthoryear{Waschb{\"u}sch, W{\"u}rmlin, Cotting, Sadlo,
  and Gross}{Waschb{\"u}sch et~al\mbox{.}}{2005}]%
        {waschbusch2005scalable}
\bibfield{author}{\bibinfo{person}{Michael Waschb{\"u}sch},
  \bibinfo{person}{Stephan W{\"u}rmlin}, \bibinfo{person}{Daniel Cotting},
  \bibinfo{person}{Filip Sadlo}, {and} \bibinfo{person}{Markus Gross}.}
  \bibinfo{year}{2005}\natexlab{}.
\newblock \showarticletitle{Scalable 3D video of dynamic scenes}.
\newblock \bibinfo{journal}{{\em The Visual Computer\/}} \bibinfo{volume}{21},
  \bibinfo{number}{8-10} (\bibinfo{year}{2005}), \bibinfo{pages}{629--638}.
\newblock


\bibitem[\protect\citeauthoryear{Wei, Zhang, and Chai}{Wei
  et~al\mbox{.}}{2012}]%
        {wei2012accurate}
\bibfield{author}{\bibinfo{person}{X. Wei}, \bibinfo{person}{P. Zhang}, {and}
  \bibinfo{person}{J. Chai}.} \bibinfo{year}{2012}\natexlab{}.
\newblock \showarticletitle{Accurate Realtime Full-body Motion Capture Using a
  Single Depth Camera}.
\newblock \bibinfo{journal}{{\em ACM TOG (Proc. SIGGRAPH Asia)\/}}
  \bibinfo{volume}{31}, \bibinfo{number}{6} (\bibinfo{year}{2012}),
  \bibinfo{pages}{188:1--188:12}.
\newblock


\bibitem[\protect\citeauthoryear{Weiss, Hirshberg, and Black}{Weiss
  et~al\mbox{.}}{2011}]%
        {weiss2011home}
\bibfield{author}{\bibinfo{person}{Alexander Weiss}, \bibinfo{person}{David
  Hirshberg}, {and} \bibinfo{person}{Michael~J Black}.}
  \bibinfo{year}{2011}\natexlab{}.
\newblock \showarticletitle{Home 3D body scans from noisy image and range
  data}. In \bibinfo{booktitle}{{\em Proc. ICCV}}. IEEE,
  \bibinfo{pages}{1951--1958}.
\newblock


\bibitem[\protect\citeauthoryear{Wu, Stoll, Valgaerts, and Theobalt}{Wu
  et~al\mbox{.}}{2013}]%
        {wu2013onset}
\bibfield{author}{\bibinfo{person}{Chenglei Wu}, \bibinfo{person}{Carsten
  Stoll}, \bibinfo{person}{Levi Valgaerts}, {and} \bibinfo{person}{Christian
  Theobalt}.} \bibinfo{year}{2013}\natexlab{}.
\newblock \showarticletitle{{On-set Performance Capture of Multiple Actors With
  A Stereo Camera}}. In \bibinfo{booktitle}{{\em ACM Transactions on Graphics
  (Proceedings of SIGGRAPH Asia 2013)}}, Vol.~\bibinfo{volume}{32}.
  \bibinfo{pages}{161:1--161:11}.
\newblock
\showDOI{%
\url{https://doi.org/10.1145/2508363.2508418}}


\bibitem[\protect\citeauthoryear{Wu, Varanasi, and Theobalt}{Wu
  et~al\mbox{.}}{2012}]%
        {wu2012full}
\bibfield{author}{\bibinfo{person}{Chenglei Wu}, \bibinfo{person}{Kiran
  Varanasi}, {and} \bibinfo{person}{Christian Theobalt}.}
  \bibinfo{year}{2012}\natexlab{}.
\newblock \showarticletitle{Full body performance capture under uncontrolled
  and varying illumination: A shading-based approach}. In
  \bibinfo{booktitle}{{\em ECCV}}. \bibinfo{pages}{757--770}.
\newblock


\bibitem[\protect\citeauthoryear{Xu, Chatterjee, Zoll{\"o}fer, Rhodin, Mehta,
  Seidel, and Theobalt}{Xu et~al\mbox{.}}{2018}]%
        {xu17MonoPerfCap}
\bibfield{author}{\bibinfo{person}{Weipeng Xu}, \bibinfo{person}{Avishek
  Chatterjee}, \bibinfo{person}{Michael Zoll{\"o}fer}, \bibinfo{person}{Helge
  Rhodin}, \bibinfo{person}{Dushyant Mehta}, \bibinfo{person}{Hans-Peter
  Seidel}, {and} \bibinfo{person}{Christian Theobalt}.}
  \bibinfo{year}{2018}\natexlab{}.
\newblock \showarticletitle{{MonoPerfCap: Human Performance Capture from
  Monocular Video}}.
\newblock \bibinfo{journal}{{\em ACM Transactions on Graphics (TOG)\/}}
  (\bibinfo{year}{2018}).
\newblock


\bibitem[\protect\citeauthoryear{Yang, Franco, H{\'e}troy-Wheeler, and
  Wuhrer}{Yang et~al\mbox{.}}{2016}]%
        {yang2016estimation}
\bibfield{author}{\bibinfo{person}{Jinlong Yang},
  \bibinfo{person}{Jean-S{\'e}bastien Franco}, \bibinfo{person}{Franck
  H{\'e}troy-Wheeler}, {and} \bibinfo{person}{Stefanie Wuhrer}.}
  \bibinfo{year}{2016}\natexlab{}.
\newblock \showarticletitle{{Estimation of Human Body Shape in Motion with Wide
  Clothing}}. In \bibinfo{booktitle}{{\em {European Conference on Computer
  Vision 2016}}}. \bibinfo{address}{Amsterdam, Netherlands}.
\newblock


\bibitem[\protect\citeauthoryear{Ye, Liu, Hasler, Ji, Dai, and Theobalt}{Ye
  et~al\mbox{.}}{2012}]%
        {Ye2012}
\bibfield{author}{\bibinfo{person}{Genzhi Ye}, \bibinfo{person}{Yebin Liu},
  \bibinfo{person}{Nils Hasler}, \bibinfo{person}{Xiangyang Ji},
  \bibinfo{person}{Qionghai Dai}, {and} \bibinfo{person}{Christian Theobalt}.}
  \bibinfo{year}{2012}\natexlab{}.
\newblock \showarticletitle{{Performance capture of interacting characters with
  handheld kinects}}. In \bibinfo{booktitle}{{\em ECCV}},
  Vol.~\bibinfo{volume}{7573 LNCS}. \bibinfo{pages}{828--841}.
\newblock
\showISBNx{9783642337086}
\showISSN{03029743}
\showDOI{%
\url{https://doi.org/10.1007/978-3-642-33709-3_59}}


\bibitem[\protect\citeauthoryear{Ye and Yang}{Ye and Yang}{2014}]%
        {ye2014real}
\bibfield{author}{\bibinfo{person}{Mao Ye} {and} \bibinfo{person}{Ruigang
  Yang}.} \bibinfo{year}{2014}\natexlab{}.
\newblock \showarticletitle{Real-time simultaneous pose and shape estimation
  for articulated objects using a single depth camera}. In
  \bibinfo{booktitle}{{\em Proceedings of the IEEE Conference on Computer
  Vision and Pattern Recognition}}. \bibinfo{pages}{2345--2352}.
\newblock


\bibitem[\protect\citeauthoryear{Yu, Russell, Campbell, and Agapito}{Yu
  et~al\mbox{.}}{2015}]%
        {Yu_2015_ICCV}
\bibfield{author}{\bibinfo{person}{Rui Yu}, \bibinfo{person}{Chris Russell},
  \bibinfo{person}{Neill D.~F. Campbell}, {and} \bibinfo{person}{Lourdes
  Agapito}.} \bibinfo{year}{2015}\natexlab{}.
\newblock \showarticletitle{Direct, Dense, and Deformable: Template-Based
  Non-Rigid 3D Reconstruction From RGB Video}. In \bibinfo{booktitle}{{\em The
  IEEE International Conference on Computer Vision (ICCV)}}.
\newblock


\bibitem[\protect\citeauthoryear{Yu, Guo, Xu, Dong, Su, Zhao, Li, Dai, and
  Liu}{Yu et~al\mbox{.}}{2017}]%
        {BodyFusion}
\bibfield{author}{\bibinfo{person}{Tao Yu}, \bibinfo{person}{Kaiwen Guo},
  \bibinfo{person}{Feng Xu}, \bibinfo{person}{Yuan Dong},
  \bibinfo{person}{Zhaoqi Su}, \bibinfo{person}{Jianhui Zhao},
  \bibinfo{person}{Jianguo Li}, \bibinfo{person}{Qionghai Dai}, {and}
  \bibinfo{person}{Yebin Liu}.} \bibinfo{year}{2017}\natexlab{}.
\newblock \showarticletitle{BodyFusion: Real-time Capture of Human Motion and
  Surface Geometry Using a Single Depth Camera}. In \bibinfo{booktitle}{{\em
  The IEEE International Conference on Computer Vision (ICCV)}}.
  \bibinfo{publisher}{ACM}.
\newblock


\bibitem[\protect\citeauthoryear{Yu, Zheng, Guo, Zhao, Dai, Li, Pons-Moll, and
  Liu}{Yu et~al\mbox{.}}{2018}]%
        {DoubleFusion2018}
\bibfield{author}{\bibinfo{person}{Tao Yu}, \bibinfo{person}{Zerong Zheng},
  \bibinfo{person}{Kaiwen Guo}, \bibinfo{person}{Jianhui Zhao},
  \bibinfo{person}{Qionghai Dai}, \bibinfo{person}{Hao Li},
  \bibinfo{person}{Gerard Pons-Moll}, {and} \bibinfo{person}{Yebin Liu}.}
  \bibinfo{year}{2018}\natexlab{}.
\newblock \showarticletitle{DoubleFusion: Real-time Capture of Human
  Performances with Inner Body Shapes from a Single Depth Sensor}. In
  \bibinfo{booktitle}{{\em The IEEE International Conference on Computer Vision
  and Pattern Recognition(CVPR)}}. \bibinfo{publisher}{IEEE}.
\newblock


\bibitem[\protect\citeauthoryear{Zhang, Pujades, Black, and Pons-Moll}{Zhang
  et~al\mbox{.}}{2017}]%
        {ZhangCVPR2017}
\bibfield{author}{\bibinfo{person}{Chao Zhang}, \bibinfo{person}{Sergi
  Pujades}, \bibinfo{person}{Michael Black}, {and} \bibinfo{person}{Gerard
  Pons-Moll}.} \bibinfo{year}{2017}\natexlab{}.
\newblock \showarticletitle{Detailed, accurate, human shape estimation from
  clothed {3D} scan sequences}. In \bibinfo{booktitle}{{\em IEEE Conference on
  Computer Vision and Pattern Recognition (CVPR)}}.
\newblock
\newblock
\shownote{Spotlight.}


\bibitem[\protect\citeauthoryear{Zhang, Siu, Zhang, Liu, and Chai}{Zhang
  et~al\mbox{.}}{2014}]%
        {zhang2014depth}
\bibfield{author}{\bibinfo{person}{Peizhao Zhang}, \bibinfo{person}{Kristin
  Siu}, \bibinfo{person}{Jianjie Zhang}, \bibinfo{person}{C.~Karen Liu}, {and}
  \bibinfo{person}{Jinxiang Chai}.} \bibinfo{year}{2014}\natexlab{}.
\newblock \showarticletitle{Leveraging Depth Cameras and Wearable Pressure
  Sensors for Full-body Kinematics and Dynamics Capture}.
\newblock \bibinfo{journal}{{\em ACM Transactions on Graphics (TOG)\/}}
  \bibinfo{volume}{33}, \bibinfo{number}{6} (\bibinfo{year}{2014}),
  \bibinfo{pages}{14}.
\newblock


\bibitem[\protect\citeauthoryear{Zhang, Fu, Ye, and Yang}{Zhang
  et~al\mbox{.}}{2014}]%
        {zhang2014quality}
\bibfield{author}{\bibinfo{person}{Qing Zhang}, \bibinfo{person}{Bo Fu},
  \bibinfo{person}{Mao Ye}, {and} \bibinfo{person}{Ruigang Yang}.}
  \bibinfo{year}{2014}\natexlab{}.
\newblock \showarticletitle{Quality dynamic human body modeling using a single
  low-cost depth camera}. In \bibinfo{booktitle}{{\em 2014 IEEE Conference on
  Computer Vision and Pattern Recognition}}. IEEE, \bibinfo{pages}{676--683}.
\newblock


\bibitem[\protect\citeauthoryear{Zhou and Koltun}{Zhou and Koltun}{2014}]%
        {zhou2014color}
\bibfield{author}{\bibinfo{person}{Qian-Yi Zhou} {and} \bibinfo{person}{Vladlen
  Koltun}.} \bibinfo{year}{2014}\natexlab{}.
\newblock \showarticletitle{Color map optimization for 3D reconstruction with
  consumer depth cameras}.
\newblock \bibinfo{journal}{{\em ACM Transactions on Graphics (TOG)\/}}
  \bibinfo{volume}{33}, \bibinfo{number}{4} (\bibinfo{year}{2014}),
  \bibinfo{pages}{155}.
\newblock


\bibitem[\protect\citeauthoryear{Zhou, Fu, Liu, Cohen-Or, and Han}{Zhou
  et~al\mbox{.}}{2010}]%
        {zhou2010parametric}
\bibfield{author}{\bibinfo{person}{Shizhe Zhou}, \bibinfo{person}{Hongbo Fu},
  \bibinfo{person}{Ligang Liu}, \bibinfo{person}{Daniel Cohen-Or}, {and}
  \bibinfo{person}{Xiaoguang Han}.} \bibinfo{year}{2010}\natexlab{}.
\newblock \showarticletitle{Parametric reshaping of human bodies in images}.
\newblock \bibinfo{journal}{{\em ACM Transactions on Graphics (TOG)\/}}
  \bibinfo{volume}{29}, \bibinfo{number}{4} (\bibinfo{year}{2010}),
  \bibinfo{pages}{126}.
\newblock


\bibitem[\protect\citeauthoryear{Zhou, Huang, Sun, Xue, and Wei}{Zhou
  et~al\mbox{.}}{2017}]%
        {zhou2017towards}
\bibfield{author}{\bibinfo{person}{Xingyi Zhou}, \bibinfo{person}{Qixing
  Huang}, \bibinfo{person}{Xiao Sun}, \bibinfo{person}{Xiangyang Xue}, {and}
  \bibinfo{person}{Yichen Wei}.} \bibinfo{year}{2017}\natexlab{}.
\newblock \showarticletitle{Towards 3D Human Pose Estimation in the Wild: A
  Weakly-Supervised Approach}. In \bibinfo{booktitle}{{\em Proceedings of the
  IEEE Conference on Computer Vision and Pattern Recognition}}.
  \bibinfo{pages}{398--407}.
\newblock


\bibitem[\protect\citeauthoryear{Zhou, Zhu, Leonardos, Derpanis, and
  Daniilidis}{Zhou et~al\mbox{.}}{2016}]%
        {zhou2016sparseness}
\bibfield{author}{\bibinfo{person}{Xiaowei Zhou}, \bibinfo{person}{Menglong
  Zhu}, \bibinfo{person}{Spyridon Leonardos}, \bibinfo{person}{Konstantinos~G
  Derpanis}, {and} \bibinfo{person}{Kostas Daniilidis}.}
  \bibinfo{year}{2016}\natexlab{}.
\newblock \showarticletitle{Sparseness meets deepness: 3D human pose estimation
  from monocular video}. In \bibinfo{booktitle}{{\em Proceedings of the IEEE
  Conference on Computer Vision and Pattern Recognition}}.
  \bibinfo{pages}{4966--4975}.
\newblock


\bibitem[\protect\citeauthoryear{Zivkovic and van~der Heijden}{Zivkovic and
  van~der Heijden}{2006}]%
        {Zivkovic:2006:EAD:1142319.1142328}
\bibfield{author}{\bibinfo{person}{Zoran Zivkovic} {and}
  \bibinfo{person}{Ferdinand van~der Heijden}.}
  \bibinfo{year}{2006}\natexlab{}.
\newblock \showarticletitle{Efficient Adaptive Density Estimation Per Image
  Pixel for the Task of Background Subtraction}.
\newblock \bibinfo{journal}{{\em Pattern Recogn. Lett.\/}}
  \bibinfo{volume}{27}, \bibinfo{number}{7} (\bibinfo{date}{May}
  \bibinfo{year}{2006}), \bibinfo{pages}{773--780}.
\newblock
\showISSN{0167-8655}
\showDOI{%
\url{https://doi.org/10.1016/j.patrec.2005.11.005}}


\bibitem[\protect\citeauthoryear{Zollh{\"o}fer, Nie{\ss}ner, Izadi, Rhemann,
  Zach, Fisher, Wu, Fitzgibbon, Loop, Theobalt, and Stamminger}{Zollh{\"o}fer
  et~al\mbox{.}}{2014}]%
        {zollhoefer2014deformable}
\bibfield{author}{\bibinfo{person}{Michael Zollh{\"o}fer},
  \bibinfo{person}{Matthias Nie{\ss}ner}, \bibinfo{person}{Shahram Izadi},
  \bibinfo{person}{Christoph Rhemann}, \bibinfo{person}{Christopher Zach},
  \bibinfo{person}{Matthew Fisher}, \bibinfo{person}{Chenglei Wu},
  \bibinfo{person}{Andrew Fitzgibbon}, \bibinfo{person}{Charles Loop},
  \bibinfo{person}{Christian Theobalt}, {and} \bibinfo{person}{Marc
  Stamminger}.} \bibinfo{year}{2014}\natexlab{}.
\newblock \showarticletitle{Real-time Non-rigid Reconstruction using an RGB-D
  Camera}.
\newblock \bibinfo{journal}{{\em ACM Transactions on Graphics (TOG)\/}}
  \bibinfo{volume}{33}, \bibinfo{number}{4} (\bibinfo{year}{2014}).
\newblock


\end{thebibliography}

\end{document}